\documentclass[10pt]{article}

\usepackage[preprint]{template_neurips}

\usepackage[utf8]{inputenc}
\usepackage[T1]{fontenc}
\usepackage{microtype}
\usepackage{nicefrac}

\usepackage[dvipsnames,svgnames]{xcolor}
\definecolor{grey}{rgb}{0.9,0.9,0.9}
\definecolor{codegreen}{rgb}{0,0.6,0}
\definecolor{codegray}{rgb}{0.5,0.5,0.5}
\definecolor{codepurple}{rgb}{0.58,0,0.82}
\definecolor{backcolour}{rgb}{0.95,0.95,0.92}
\definecolor{YesGreen}{RGB}{22,130,73}
\definecolor{NoRed}{RGB}{189,44,44}
\definecolor{PartAmber}{RGB}{172,105,0}
\definecolor{OursTeal}{RGB}{12,110,103}
\definecolor{OursColBg}{RGB}{236,253,248}
\definecolor{OursBorder}{RGB}{0,150,136}
\definecolor{CatBg}{RGB}{229,236,252}
\definecolor{CatFg}{RGB}{30,60,114}
\definecolor{NaGray}{RGB}{160,160,160}

\usepackage{amsmath}
\usepackage{amssymb}
\usepackage{amsfonts}
\usepackage{mathtools}
\usepackage{amsthm}
\usepackage{bm}
\usepackage{xfrac}

\theoremstyle{plain}
\newtheorem{theorem}{Theorem}

\theoremstyle{definition}

\theoremstyle{remark}

\newtheorem*{proposition*}{Proposition}

\usepackage{graphicx}
\graphicspath{{figures/}}
\usepackage{subfigure}
\usepackage{subcaption}
\usepackage{wrapfig}
\usepackage{float}
\usepackage{adjustbox}
\usepackage{afterpage}
\usepackage{fontawesome5}

\usepackage{tikz}
\usetikzlibrary{
    fit, backgrounds, arrows.meta,
    decorations.pathreplacing, decorations.pathmorphing,
    decorations.markings, decorations.shapes,
    shapes.geometric, positioning, calc,
}
\pgfdeclarelayer{background}
\pgfsetlayers{background,main}

\usepackage{booktabs}
\usepackage{multirow}
\usepackage{colortbl}
\usepackage{tabularx}
\usepackage{ragged2e}
\usepackage{makecell}
\newcolumntype{K}[1]{>{\centering\arraybackslash}p{#1}}

\usepackage[most]{tcolorbox}
\newtcolorbox{conclusionbox}{
  colback=blue!5, colframe=blue!75!black, coltitle=black,
  fonttitle=\bfseries, boxrule=1pt, arc=1mm,
  left=2mm, right=2mm, top=1mm, bottom=1mm,
}
\newtcolorbox{insightbox}{
  colback=OursColBg, colframe=OursBorder, coltitle=black,
  fonttitle=\bfseries, boxrule=1pt, arc=1mm,
  left=3mm, right=3mm, top=2mm, bottom=2mm,
}

\usepackage{algorithm}
\usepackage{algorithmicx,algpseudocode}

\usepackage{url}
\usepackage{csquotes}
\usepackage{xspace}
\usepackage{pifont}
\usepackage{etoolbox}
\usepackage{placeins}
\usepackage[strings]{underscore}
\usepackage{paralist}
\usepackage{comment}

\usepackage{titletoc}
\usepackage[subfigure]{tocloft}

\newcommand\DoToC{%
  \startcontents
  \printcontents{}{1}[2]{%
    {\begin{center}
       \parbox{0.99\textwidth}{%
         \centering\textbf{\LARGE Render, Don't Decode: Weight-Space World Models\\
with Latent Structural Disentanglement \\[0.3cm]\textit{---Supplementary Material---}}}
     \end{center} 
     \vskip5pt\hrule\vskip1pt}\vspace*{-10pt}}%
  \vskip8pt\hrule\vskip8pt
}

\usepackage{hyperref}
\usepackage{hypcap}
\hypersetup{
    colorlinks=true,
    linkcolor=teal,
    citecolor=Purple,
    urlcolor=Maroon,
}

\usepackage[capitalize,noabbrev]{cleveref}

\setcitestyle{square}
\renewcommand{\cite}{\citep}
\renewcommand*{\bibfont}{\small}

\newcommand{\iset}[2]{\{#1,\ldots,#2\}}

\newcommand{\themethod}{\text{NOVA}\xspace}

\newcommand{\rdn}[1]{\textcolor{red}{#1}}

\newcommand{\cmark}{\textcolor{YesGreen}{\ding{51}}}
\newcommand{\xmark}{\textcolor{NoRed}{\ding{55}}}
\newcommand{\pmark}{\textcolor{PartAmber}{\raisebox{0.05ex}{\small$\approx$}}}

\title{Render, Don't Decode: Weight-Space World Models\\
with Latent Structural Disentanglement}

\author{%
  Roussel Desmond Nzoyem\thanks{Work partly done while an honorary research associate at the University of Bristol.} \\
  Department of Computer Science\\
  University of Manchester\\
  Manchester, M13 9PL \\
  \texttt{roussel.nzoyem@manchester.ac.uk} \\
  \And
  Mauro Comi \\
  Department of Computer Science\\
  University of Bristol \\
  Bristol, BS8 1QU \\
  \texttt{mauro.comi@bristol.ac.uk} \\
}

\begin{document}

\maketitle

\begin{abstract}

Training world models on vast quantities of unlabelled videos is a critical step toward fully autonomous intelligence. However, the prevailing paradigm of encoding raw pixels into opaque latent spaces and relying on heavy decoders for reconstruction leaves these models computationally expensive and uninterpretable. We address this problem by introducing \textbf{\themethod}, a world modelling framework that represents the system state as the weights and biases of an auxiliary coordinate-based implicit neural representation (INR). This structured representation is analytically rendered, which eliminates the decoder bottleneck while conferring compactness, portability, and zero-shot super-resolution. Furthermore, like most latent action models, \themethod can be distilled into a context-dependent video generator via an action-matching objective. Surprisingly, without resorting to auxiliary losses or adversarial objectives, \themethod can disentangle structural scene components such as background, foreground, and inter-frame motion, enabling users to edit either content or dynamics without compromising the other. We validate our framework on several challenging datasets, achieving strong controllable forecasting while operating on a single consumer GPU at $\sim$40M parameters. Ultimately, structured representations like INRs not only enhance our understanding of latent dynamics but also pave the way for immersive and customisable virtual experiences.

\end{abstract}

\section{Introduction}
\label{sec:introduction}


The capacity to predict plausible future states within an environment is a cornerstone of physical intelligence. World models \cite{ha2018world} achieve this by learning transition dynamics conditioned on actions, enabling embodied agents to plan and act through internal simulation (e.g. robotic manipulation and autonomous driving). Recently, the abundance of unlabelled, action-free online videos has driven the emergence of \textbf{latent action (world) models} \cite{garrido2026learning,bruce2024genie,schmidt2023learning}. Devoid of ground-truth action signals during training, they must discover underlying control mechanisms purely from observation, bringing them closer to video generation models, which learn to produce coherent sequences of frames from noise or context. 

The predominant paradigm for training world models and video generation models involves projecting and reconstructing frame sequences through a latent space, for example, using a video tokeniser \cite{bruce2024genie}.
A prominent class of autoregressive models encode individual frames into \emph{abstract} latent vectors via a variational autoencoder \cite{kingma2013auto}, then map resulting predictions back to pixels via a learned spatial decoder. This paradigm has been immensely productive \cite{hafner2020dream, micheli2022transformers, bruce2024genie}, but carries intrinsic limitations. 
First, the decoder consumes more storage, adding a number of hardly interpretable parameters that could limit usability in safety-critical scenarios \cite{rudin2019stop}. 
As a case in point, \citet[pp.~6, 23]{bruce2024genie} find that as they scale, a much larger decoder is essential, accounting for up to $87\%$ of their video tokeniser.
Second, any change in output resolution requires expensive retraining or post-hoc upsampling workarounds \cite{chambon2025naf}. Third, the latents it consumes lack portability, as they are tethered to that specific decoder. Furthermore, they have no semantic or geometric interpretations. Current attempts to disentangle semantic components such as background, identity, and motion from an abstract latent space require auxiliary losses \cite{guen2020disentangling}, adversarial objectives \cite{denton2017unsupervised}, or architectural decompositions \cite{hsieh2018learning}, all of which add training complexity without structural guarantees.

Coordinate-based \textbf{implicit neural representations} (INRs) \cite{sitzmann2020implicit,dupont2022data} represent an image signal not as a pixel array but as the weights of a neural network, thus encoding a continuous function $\mathbb{R}^2 \to \mathbb{R}^C$ that maps any spatial coordinate to its RGB ($C=3$) or greyscale ($C=1$) pixel value. They gained prominence by powering novel view synthesis frameworks such as NeRF \cite{mildenhall2021nerf}, and have recently demonstrated compression capabilities that surpass JPEG at low bitrates \cite{mostajeran2025context}. Crucially, further advantages arise from an analytical rendering process that circumvents the need for a spatial decoder. Despite inherent spectral aliasing challenges \cite{barron2021mip,lindell2022bacon}, properly constrained parameter-free rendering intrinsically supports continuous, arbitrary-resolution querying at test time, a capability that is critical in domains such as extreme weather events forecasting, where a single pixel can encapsulate a vast area of the globe (e.g., over 784 km$^2$ in state-of-the-art global AI models \cite{lam2023learning}); in such contexts, unlocking granular resolution can be the difference between life and death for massive populations. Furthermore, unlike abstract encodings, INRs are highly portable, as their analytical formulation makes them easily sharable across the scientific community. Given these fundamental strengths, \emph{structured} representations like INRs and Gaussian splats \cite{kerbl20233d} are increasingly recognised as the most promising technologies to power next-generation scientific modelling, visual perception, and computer graphics.

In this paper, we argue that INRs are a potent representation upon which world models and video generation models can be built. To realise this, we leverage the paradigm of \textbf{weight-space learning} (WSL), which views the weights and biases of neural networks such as INRs as data points for a higher-level learning system \cite{schurholt2024towards}. This paradigm was recently utilised by \citet{li2025weightflow} and \citet{nzoyem2026weightspace} to achieve remarkable performance on time series modelling benchmarks. These WSL approaches, however, remain untested on video modalities. Crucially, the latter proved unstable, requiring weight, gradient, and value clipping to prevent numerical explosion. Any viable weight-space world model must first address these acute stability issues.

\begin{figure}[t]
\centering
\includegraphics[width=0.85\linewidth]{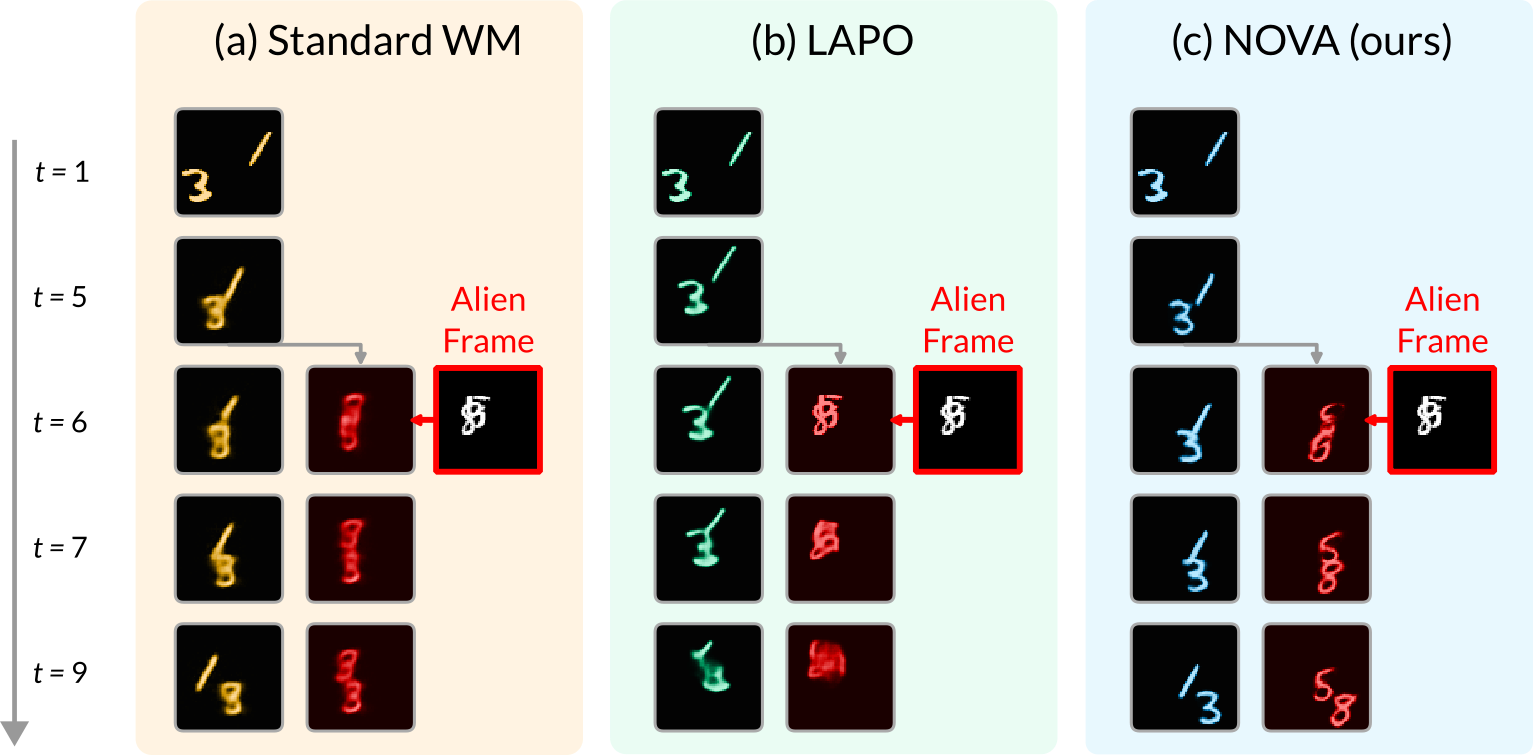}
\caption{\textbf{Video editing facilitated by object-motion disentanglement}. Conditioned on the same initial frame, all models must generate future states; at $t=6$, this state is abruptly replaced with an alien frame's encoding. \textbf{(a)} The standard world model (WM), trained in an abstract latent space with \themethod's decomposition strategy, manages to separate content from motion. \textbf{(b)} LAPO \cite{schmidt2023learning} generates clean frames, but allows the foreign frame to completely alter its dynamics; furthermore, its generation quality degrades over extended horizons (starting at $t=9$). \textbf{(c)} \themethod generates visually striking digits over long horizons and swaps their identities without altering the underlying dynamics (e.g., a $5$ seamlessly inherits the trajectory of a $1$).}
\label{fig:fork_comparison}
\vspace{-8pt}
\end{figure}

To bridge this gap, we introduce \textbf{\themethod} (\textbf{N}eural \textbf{O}ntology for \textbf{V}isual \textbf{A}bstraction), a stable and physically consistent world model framework that evolves dynamics entirely in the weight space of an implicit \emph{neural} representation. Its straightforward architecture induces a three-level semantic \emph{ontology} that mirrors human intuition about object-centric video structure: ($i$) the \textbf{background} ($\bar{\bm{z}}$) is captured by a single set of base weights shared across the entire dataset, learned to encode everything common to all videos; ($ii$) the \textbf{foreground content} ($\bm{z}_t$) is defined by per-frame weight offsets encoding what is present in the current video frame but not shared globally; and ($iii$) the \textbf{motion} ($\bm{u}_t$) is an inter-frame action signal that dictates how a latent state transitions to the next. To achieve such \emph{visual abstraction}, our framework leverages an additive formulation that cleanly isolates these properties during both forward dynamics and base composition prior to rendering.

Specifically, our key contributions are as follows:
\begin{enumerate}
    \item We build \themethod, the first world model defined by a weight-space representation. 
    We demonstrate that the inclusion of a base representation $\bar{\bm{z}}$ addresses the characteristic instability of weight-space model training. 
    Similar to recent literature \cite{schmidt2023learning}, we show that this framework can be tuned to produce actions autoregressively during inference, enabling seamless transition from world modelling to context-conditioned video generation. 
    \item We show that the $(\bar{\bm{z}}, \bm{z}_t, \bm{u}_t)$ parametrisation induces a semantically meaningful three-level decomposition, i.e. background, state, and motion, purely by construction with no auxiliary or adversarial losses. 
    This in turn empowers users to ``edit'' the generation process, whereby importing frames or actions from an arbitrary sequence cleanly maps to new sequences, a process we call \textbf{zero-shot content (or motion) retargeting} (\cref{fig:fork_comparison}).
    This hints that, contrary to expectations from recent work \cite{daniel2026latent}, global action vectors can support local dynamics disentanglement when paired with a structured base representation. 
    \item We highlight the numerous benefits of INRs as structured latent representations for video and world models (e.g., parameter-efficiency, compactness, portability, and resolution-independence).
    Remarkably, \themethod maintains \textbf{zero-shot super-resolution} capabilities even when querying coordinate grids that are significantly denser than those used during training. To facilitate this, we implement a simple but principled post-training Nyquist-Shannon frequency masking procedure that provably prevents aliasing at arbitrary scale factors.
    \item We provide a comprehensive set of benchmarks to showcase the practical usefulness of our method in sequential forecasting. We show this with experiments across three carefully selected video datasets (Moving MNIST, PhyWorld, and WeatherBench) covering continuous physics, rigid-body collisions, and geospatial dynamics, all trained on a single consumer GPU at roughly 40M parameters. Furthermore, these datasets serve as the basis for ablation studies that probe the NOVA architecture and reveal its inner workings.
\end{enumerate}


While our approach yields several practical benefits, we emphasise that our primary contribution lies in demonstrating that world models can be constructed entirely from structured weight-space representations rather than raw pixels. In doing so, we propose an effective disentanglement technique applicable to standard world models, as observed in \cref{fig:fork_comparison}. To elucidate this contribution, \cref{sec:setup} outlines the experimental setup, followed by a detailed exposition of our method in \cref{sec:method}. We present and analyse main results in \cref{sec:results}, and conclude with a discussion of the method's broader implications and limitations in \cref{sec:conclusion}.

\section{Experimental Setup}
\label{sec:setup}

\paragraph{Problem Setting.}

We consider an environment generating a sequence of 2D observations $\mathbf{o}_t \in \mathbb{R}^{H \times W \times C}$ for $t \in \iset{1}{T}$, with $T\geq2$. World modelling\footnote{To eliminate confusion from inconsistent definitions, we adopt the term ``world model'' to refer solely to the transition model, excluding adjacent components that encode states or infer actions, similar to \cite{schmidt2023learning}.} aims to discover transition dynamics $\bm{z}_{t+1} \approx \mathcal{T}(\bm{z}_t, \bm{u}_t)$, where $\bm{z}_{t} \in \mathbb{R}^{d_{\bm{z}}}$ denotes a (latent) representation of the underlying state, from which the observation $\mathbf{o}_{t}$ can be reconstructed; $\bm u_t \in \mathbb{R}^{d_{\bm{u}}}$ denotes actions typically known and part of the training data. When ground-truth actions are unavailable, the model is self-supervised and must infer $\bm{u}_t$ from $\bm o_t$ and $\bm o_{t+1}$; we refer to such models as \emph{latent action models} (LAMs).


Because both system states and actions may live in a ``latent'' space, throughout the paper, we prioritise the term \emph{weight space}, as defined in \cite{zhou2023neural}, for the space in which the dynamics evolve, i.e., $\mathbb{R}^{d_{\bm{z}}}$, and \emph{action space} for the actions themselves, i.e., $\mathbb{R}^{d_{\bm{u}}}$.



\paragraph{Datasets.} We train and evaluate on three main datasets spanning diverse dynamics:
\begin{inparaenum}[($1$)]
    \item \textbf{Moving MNIST} \cite{srivastava2015unsupervised} is a seminal multi-entity dataset showing two digits moving at constant velocity and bouncing on a grid. Velocity direction and magnitude are randomly selected, and the kinematics are deterministic; digits overlap without collision, and bouncing reflections are non-linear.
    \item \textbf{PhyWorld Collision 30K} \cite{kang2025how} comprises roughly 30 thousand simulated rigid-body collisions with varying ball radii and velocities. It tests OOD generalisation and strict physical consistency (momentum and kinetic energy conservation).
    \item \textbf{WeatherBench (2m temperature)} \cite{rasp2020weatherbench} is a processed version of the ERA5 archive \cite{hersbach2020era5}, containing global atmospheric data recorded at hourly intervals from 1979 to 2018. Specifically, we utilise the 2-meter temperature subset at a 5.625° resolution, challenging models to predict atmospheric flows that exhibit highly complex and non-linear behaviour.
    
\end{inparaenum}

\paragraph{Baselines.}
We benchmark our approach against two representative paradigms. First, a \textbf{standard WM} (see \cref{fig:naive_model} in \cref{sec:related}) is a LAM that utilises an encoder-decoder setup for fair comparison against \themethod. This baseline shares \themethod's training configurations, and differs only in its use of an abstract latent space (of comparable size) rather than the weight space to progress its dynamics. Second, we compare against \textbf{LAPO} \cite{schmidt2023learning}, a framework that removes the encoder-decoder setup by operating directly in pixel space. LAPO infers discrete action codes using an inverse dynamics network with two VQ codebooks \cite{van2017neural} optimised via EMA. Additionally, we include \textbf{WARP} \cite{nzoyem2026weightspace} as a baseline in \cref{sec:comparison_wsl}.



\paragraph{Metrics.} To quantify temporal consistency and structural preservation, we evaluate predictions against the empirical ground truth observations and their corresponding distribution of pixel intensities. We assess identity drift and distributional overlap using the \textbf{Wasserstein-1} ($W_1$) distance, \textbf{Jensen--Shannon Divergence} (JSD), and \textbf{Bhattacharyya} distance ($B$). Spatial and frequency-domain structural integrity are measured via the \textbf{Structural Similarity Index Measure} (SSIM) and \textbf{FFT magnitude distance} ($d_{\mathcal{F}}$), respectively. Furthermore, to avoid direct pixel-level comparisons, \textbf{Mean Squared Error} (MSE) is applied to evaluate the accuracy of object coordinates, masses, and velocities. 
Additional experimental datasets and formally defined metrics can be found in \cref{subsec_app:datasets}.

\section{The \themethod Framework}
\label{sec:method}

\subsection{Architecture Overview}

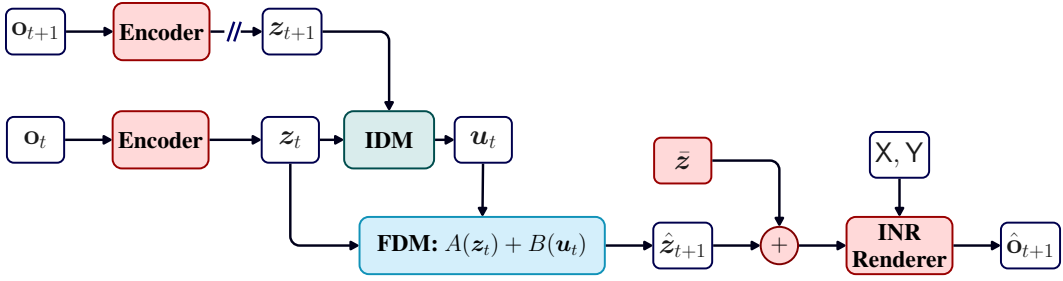
\begin{figure}[t]
\centering
\resizebox{1.0\textwidth}{!}{%
\begin{tikzpicture}[
    node distance=0.8cm and 1cm, 
    base_arrow/.style={draw=blue!10!black, line width=1.5pt, >={Triangle[length=2.3mm, width=2.3mm, round]}, rounded corners=2mm},
    arrow/.style={->, base_arrow},
    block/.style={
        rectangle, rounded corners=2mm, align=center,
        minimum height=3.5em, minimum width=5.5em, font=\Large\bfseries,
        line width=1.2pt
    },
    data/.style={
        rectangle, rounded corners=1.5mm, align=center,
        minimum height=2.8em, minimum width=3.5em,
        fill=white, draw=blue!30!black, text=black!90,
        font=\LARGE\bfseries,
        line width=1.2pt
    },
    switch/.style={
        shape=diamond, aspect=2.0, inner sep=2pt, font=\Large\bfseries,
        fill=teal!15, draw=teal!60!black, line width=1.2pt
    },
    sum_node/.style={
        circle, inner sep=3pt, font=\Large\bfseries,
        fill=red!15, draw=red!50!black, line width=1.2pt
    },
    enc_style/.style={block, fill=red!15, draw=red!60!black},
    idm_style/.style={block, fill=teal!15, draw=teal!60!black},
    fdm_style/.style={block, fill=cyan!15, draw=cyan!70!black},
    stop gradient/.style={
        postaction={decorate},
        decoration={markings, mark=at position #1 with {
            \fill[white] (-4pt, -8pt) rectangle (6pt, 8pt);
            \draw[line width=1.5pt, blue!40!black, -] (-2pt,-6pt) -- (2pt,6pt);
            \draw[line width=1.5pt, blue!40!black, -] (2pt,-6pt) -- (6pt,6pt);
        }}
    }, stop gradient/.default=0.5
]
  \node[data] (xt1_gt) {$\mathbf{o}_{t+1}$};
  \node[enc_style, right=of xt1_gt] (enc_gt) {Encoder};
  \node[data, right=1.1cm of enc_gt] (thetat1_gt) {$\bm{z}_{t+1}$};
  
  \node[data, below=of xt1_gt, yshift=-0.5cm] (xt) {$\mathbf{o}_t$};
  \node[enc_style, right=of xt] (enc) {Encoder};
  \node[data, right=1.1cm of enc] (thetat) {$\bm{z}_t$};
  
  \node[idm_style, right=0.5cm of thetat] (idm) {IDM};
  
  
  
  \node[data, right=0.4cm of idm] (at) {$\bm{u}_t$};
  
  \node[fdm_style, below=1cm of idm, xshift=2.0cm, minimum width=15em] (fdm) {FDM: $A(\bm{z}_t) + B(\bm{u}_t)$};
  \node[data, right=of fdm] (thetat1) {$\hat{\bm{z}}_{t+1}$};
  
  \node[data, above=of thetat1, fill=red!15, draw=red!60!black] (theta_base) {$\bar{\bm{z}}$};
  \node[sum_node, right=of thetat1] (sum2) {$+$};
  \node[enc_style, right=of sum2, minimum height=3.5em] (inr) {INR\\Renderer};
  \node[data, above=of inr] (coords) {$\mathsf{X},\mathsf{Y}$};
  \node[data, right=of inr] (xt1_pred) {$\hat{\mathbf{o}}_{t+1}$};
  
  \draw[arrow] (xt1_gt) -- (enc_gt);
  \draw[arrow, stop gradient=0.4] (enc_gt) -- (thetat1_gt);
  \draw[arrow] (xt) -- (enc);
  \draw[arrow] (enc) -- (thetat);
  \draw[arrow] (thetat) -- (idm);
  \draw[arrow] (thetat1_gt) -| (idm);
  
  
  
  \draw[arrow] (idm) -- (at);
  
  \draw[arrow] (thetat) |- (fdm);
  \draw[arrow] (at) -- (fdm);
  
  \draw[arrow] (fdm) -- (thetat1);
  \draw[arrow] (thetat1) -- (sum2);
  \draw[arrow] (theta_base) -| (sum2);
  \draw[arrow] (sum2) -- (inr);
  \draw[arrow] (coords) -- (inr);
  \draw[arrow] (inr) -- (xt1_pred);

  
\end{tikzpicture}}
\caption{\textbf{\themethod architecture.} The shared Encoder (\textcolor{red!80!black}{red}) maps frames to weight-space offsets $\bm{z}_t$. During training, the IDM (\textcolor{teal!80!black}{green}) infers latent action $\bm{u}_t$ from consecutive encoded states. The FDM (\textcolor{cyan!70!black}{blue}) predicts the next weight offset via the additive mapping $A(\bm{z}_t) + B(\bm{u}_t)$. Unlike conventional decoders, the Renderer (\textcolor{red!80!black}{red}) is not trained; it is an analytical function that maps a coordinate grid $(\mathsf{X},\mathsf{Y})$ to pixel values via an implicit neural representation (INR). The sign $\small \text{//}$ denotes a stop-gradient \texttt{sg} operator, included to reduce memory overheads during training.\protect\footnotemark{}}
\label{fig:nova_method}
\end{figure}

\footnotetext{We note that the omission of the \texttt{sg} operator does not lead to trivial representational collapse \cite{bardes2024vjepa}.}

\Cref{fig:nova_method} shows the \themethod architecture, comprising components we discuss in two categories.

\paragraph{Encoding and Rendering.} The Encoder is a strided convolutional backbone that maps each frame $\mathbf{o}_t \in \mathbb{R}^{H\times W\times C}$ to a compact vector $\bm{z}_t \in \mathbb{R}^{d_{\bm{z}}}$, where $d_{\bm{z}}$ is chosen to match the number of learnable parameters in the coordinate-based rendering network. We maintain a single set of base parameters $\bar{\bm{z}}$ learned across the entire training dataset, and pixels are rendered independently to reconstruct the future frame, in parallel:
\begin{align}
    \hat{\mathbf{o}}_{t+1} &= \text{MLP}_{\bm{\theta}_{t+1}}(\mathsf{x},\mathsf{y}), \,\, \forall (\mathsf{x}, \mathsf{y}) \in \mathsf{X} \times \mathsf{Y},  \quad
    \text{where} \quad \bm{\theta}_{t+1} \triangleq \mathtt{unflat}(\bar{\bm{z}}+\bm{z}_{t+1}).
    \label{eq:rendering}
\end{align}
Coordinates are normalised with $\mathsf{X}$ and $\mathsf{Y}$ regularly sampled in the range $[-1,1]$. The \texttt{unflat} operation turns a flat vector into MLP \cite{mcculloch1943logical} weight and bias tensors to be used alongside non-linear activation functions and Fourier feature encoding \cite{tancik2020fourier}.

Anchoring to $\bar{\bm{z}}$ provides two benefits: ($i$) it stabilises training by framing the prediction $\bm{z}_t$ as a weight-space residual, enabling the model to output small-magnitude offsets, analogous to residual networks \cite{he2016deep}; and ($ii$) it provides a dataset-level prior that absorbs static background, freeing $\bm{z}_t$ to encode only dynamic content, akin to meta-learning initialisations \cite{finn2017model}.

\paragraph{Inverse and Forward Dynamics.} The Inverse Dynamics Model (IDM) is an MLP that receives two consecutive encoded states $(\bm{z}_t, \bm{z}_{t+1})$, and outputs the latent action $\bm{u}_t$ explaining the transition. Rather than predicting $\bm{z}_{t+1}$ through a monolithic block, we employ an \emph{additive} Forward Dynamics Model (FDM): $\bm{z}_{t+1} = A(\bm{z}_t) + B(\bm{u}_t),$
where $A: \mathbb{R}^{d_{\bm{z}}} \to \mathbb{R}^{d_{\bm{z}}}$ and $B: \mathbb{R}^{d_{\bm{u}}} \to \mathbb{R}^{d_{\bm{z}}}$ are independent MLPs. $A$ operates on the current weight offset alone, while $B$ maps the latent action into the weight space, encoding how the action modulates the INR. 


This forward decomposition is inspired by linear state-space models \cite{gu2021efficiently, nzoyem2026weightspace} and extends them to nonlinear settings. Another key difference is that $A$ does not prescribe autonomous dynamics, but acts as a dedicated object detector \cite{terver2026lightweight}. Our $A/B$ decoupling is the mechanism through which the content/motion disentanglement emerges.

\subsection{Training Procedure}

Training generally proceeds in three phases, each targeting a different component of the model.

\paragraph{Phase 1.}
We pre-train the encoder and base latent $\bar{\bm{z}}$ by minimising the independent per-frame reconstruction loss $\mathcal{L}_1 = \frac{1}{T}\sum_{t=1}^T  \|\mathbf{o}_t - \hat{\mathbf{o}}_t\|_2^2 + \lambda \cdot \text{SSIM}(\mathbf{o}_t, \hat{\mathbf{o}}_t)$,
\text{where} $\hat{\mathbf{o}}_t$ is computed using \cref{eq:rendering}, and the $\lambda$-weighted $\text{SSIM}$ term prevents excessive smoothing \cite{wang2004image}.


While this phase is optional, it remains essential, as it exposes the encoder to not just the initial frames, but to all $T$ time steps of each sequence, thereby encouraging more robust feature extraction.

\paragraph{Phase 2.}
With the encoder and base latent $\bar{\bm{z}}$ frozen, we jointly train the FDM and IDM by minimising the latent transition objective:
\begin{equation}
    \mathcal{L}_2 = \frac{1}{T-1}\sum_{t=1}^{T-1} \|\bm{z}_{t+1} - A(\bm{z}_t) - B(\bm{u}_t)\|_2^2, \quad \text{where} \quad \bm{u}_t = \text{IDM}(\bm{z}_t, \bm{z}_{t+1}).
    \label{eq:phase2loss}
\end{equation}

\begin{wrapfigure}[11]{r}{0.29\textwidth}
\centering
\vspace{-18pt} 
\resizebox{0.85\linewidth}{!}{%
\begin{tikzpicture}[
    node distance=0.8cm and 1.2cm, 
    base_arrow/.style={draw=blue!10!black, line width=1.5pt, >={Triangle[length=2.3mm, width=2.3mm, round]}, rounded corners=2mm},
    arrow/.style={->, base_arrow},
    block/.style={
        rectangle, rounded corners=2mm, align=center,
        minimum height=3.5em, minimum width=5.5em, font=\Large\bfseries,
        line width=1.2pt
    },
    data/.style={
        rectangle, rounded corners=1.5mm, align=center,
        minimum height=2.8em, minimum width=3.5em,
        fill=white, draw=blue!30!black, text=black!90,
        font=\LARGE\bfseries,
        line width=1.2pt
    },
    sum_node/.style={
        circle, inner sep=3pt, font=\Large\bfseries,
        fill=teal!15, draw=teal!50!black, line width=1.2pt
    },
    idm_style/.style={block, fill=teal!15, draw=teal!60!black}
]

  \node[idm_style] (idm) {IDM};
  \node[idm_style, below=1.8cm of idm] (GCM) {GCM};

  \node[anchor=south west, inner sep=0pt, text=orange, font=\huge] at ([xshift=-11pt, yshift=-11pt]idm.north east) {\small\textcolor{cyan}{\faSnowflake}};
  
  \node[anchor=south west, inner sep=0pt, text=orange, font=\huge] at ([xshift=-10pt, yshift=-11pt]GCM.north east) {\small\faFire};

  \node[data, left=1.2cm of idm] (zt_1) {$\bm{z}_{t+1}$};
  \node[data, left=1.2cm of GCM] (ut_1) {$\bm{u}_{t-1}$};
  
  \path (zt_1) -- (ut_1) node[midway, data, xshift=-0.0cm] (zt) {$\bm{z}_t$};

  \node[data, right=1.2cm of idm] (u_t) {$\bm{u}_t$};
  \node[data, right=1.2cm of GCM] (u_hat) {$\hat{\bm{u}}_t$};

  \path (u_t) -- (u_hat) node[midway, sum_node, xshift=0.0cm] (loss) {$\mathcal{L}_3$};

  \draw[arrow] (zt_1) -- (idm);
  \draw[arrow] (ut_1) -- (GCM);
  \draw[arrow] (idm) -- (u_t);
  \draw[arrow] (GCM) -- (u_hat);

  \draw[arrow] (zt) -| ([xshift=-0pt]idm.south);
  \draw[arrow] (zt) -| ([xshift=-0pt]GCM.north);

  \draw[arrow, dashed, draw=teal!60!black] (u_t) -- (loss);
  \draw[arrow, dashed, draw=teal!60!black] (u_hat) -- (loss);

  \begin{scope}[on background layer]
    \draw[white!5, fill=white!3, line width=1.5pt, rounded corners=3mm] 
      ([shift={(-5pt,-5pt)}]current bounding box.south west) 
      rectangle 
      ([shift={(5pt,5pt)}]current bounding box.north east);
  \end{scope}

\end{tikzpicture}}
\caption{Action-matching training objective during phase 3. The GCM regresses onto the pseudo-action $\bm{u}_t$ generated by the frozen IDM.}
\label{fig:GCM_training}
\vspace{-10pt} 
\end{wrapfigure}
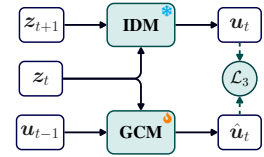

This latent-space loss is highly efficient ($d_{\bm{z}} \ll H \times W \times C$) and prevents the model from wasting capacity on irrelevant pixel-level features, similar to \cite{assran2023self}. That said, if phase 1 is skipped, pixel-space reconstruction terms can be used instead of $\mathcal{L}_2$. For instance, in our Moving MNIST experiments, jointly optimising phases 1 and 2  significantly accelerated convergence.

To regularise the dynamics models, we augment the training data distribution. First, we inject \emph{temporally inverted} trajectories, denoted by the slice notation $\mathbf{o}_{T:1:-1}$, to encourage action symmetry. Second, we crop the videos and \emph{pad the trajectory boundaries} to construct augmented $T-$length sequences $(\mathbf{o}_1, \mathbf{o}_1, \mathbf{o}_2, \dots, \mathbf{o}_P, \mathbf{o}_P)$. By introducing these static transitions where $\bm{z}_{t+1} = \bm{z}_t$, we implicitly supervise the IDM to output a ``do nothing'' action, while constraining the FDM to act as the identity mapping under this null action.


\paragraph{Phase 3.} 
In most downstream applications, future frames are unavailable, in which case we rely on a Generative Control Model (GCM), a causal sequence model, to auto-regressively predict actions $\bm{u}_t$ from $\bm{z}_{t}$ and $\bm{u}_{t-1}$ (see \cref{fig:GCM_training,alg:nova_inference}). This allows the model to autonomously generate plausible future frames, thereby transforming it into a frame-conditioned video generator or (behaviour) clone that can imitate an agent only from videos \cite{pomerleau1988alvinn}.

The GCM is trained to match the IDM's action sequence via the objective $\mathcal{L}_3 = \frac{1}{T-1}\sum_{t=1}^{T-1} \|\bm{u}_t - \text{GCM}(\bm{z}_{t}, \bm{u}_{t-1})\|_2^2$.\footnote{The required input $\bm u_0$ is computed from zero memory, with additional details in \cref{subsec_app:videogen}.} We consider both recurrent architectures like LSTM \cite{hochreiter1997long} and GRU \cite{cho2014learning} that only observe $\bm{z}_{t}$ and $\bm{u}_{t-1}$ at each time step, and Transformer-based architectures \cite{vaswani2017attention} that leverage all available $\bm{z}_{1:t}, \bm{u}_{1:t-1}$ pairs.

\subsection{Implementation} 
All models are implemented in JAX using Equinox \cite{kidger2021equinox}, optimised with Adam \cite{kingma2014adam}  with a variable learning rate optimally tuned for each dataset and phase. \themethod is trained with a single RTX 4080 GPU with 16 GB of VRAM. Across all datasets, we use the same deep and narrow INR (6 MLP layers of width 12) preceded by Fourier coordinate encoding \cite{tancik2020fourier}, which we empirically found to be expressive enough to faithfully capture a wide variety of textures. The resulting latent dimension is compact, requiring as few as $d_{\bm{z}}=961$ weights. To limit its information content, we constrain the action dimension to $d_{\bm{u}}=4$ across all experiments unless stated otherwise. By default, we prioritise recurrent architectures for the GCM to limit attention to previous action-states, since these might leak information during intervention. These choices result in a total \themethod parameter count of approximately 40M. All results are reported on the test sets, predominantly using the trained GCM to generate actions for our interventions.\footnote{The performance of the GCM is indicative of that of its IDM teacher, which supplies pseudo-actions during phase 3.}  Further architectural details, along with hyperparameters, are presented in \cref{app:architecture}. 

\section{Results and Discussion}
\label{sec:results}


Experiments are designed to demonstrate the performance of weight-space world models and video generators on readily available, yet surprisingly well-suited, unlabelled video datasets. Accordingly, we structure our results by dataset, dedicating a specific subsection to each. Additional results, including evaluations on further datasets, are provided in \cref{sec_app:results}.

\subsection{Generation, Analysis, and Editing of Moving Digits}
\label{subsec:movingmnistgen}

\paragraph{\themethod is a context-conditioned video generation model.}



\begin{figure}[h]
\vspace{-22pt}
\centering
\includegraphics[width=0.9\linewidth]{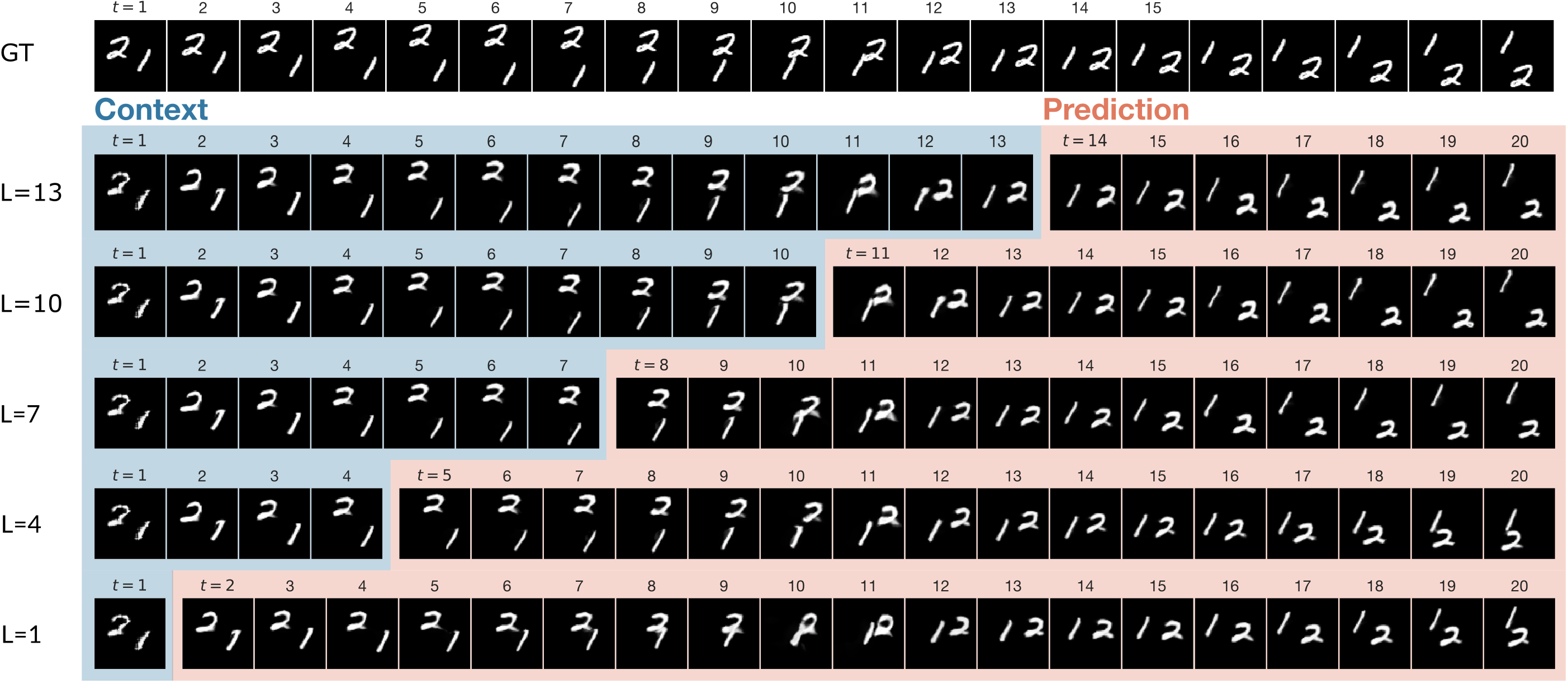}
\caption{\textbf{Robustness to minimal context.} Top row: ground truth (GT); bottom row: $L=1$ context frame. Even under this extreme constraint, \themethod produces coherent, near-perfect predictions.
}
\label{fig:video_generation_gradual}
\end{figure}

\emph{How much context does a video generation model need to capture true dynamics?} \Cref{fig:video_generation_gradual} stress-tests this dependency by progressively restricting the GCM-based generation from 13 context frames down to 1. Remarkably, \themethod mirrors the ground truth (GT) dynamics near-perfectly; we only observe a meaningful divergence once the context window drops to 4 frames. Even under the extreme constraint of a single context frame, \themethod relies on its learned dynamics priors to synthesise coherent, physically plausible trajectories without mode collapse. We observe similarly crisp, artefact-free context-conditioned generation on both PhyWorld and WeatherBench (see \cref{fig:video_generation,fig:ood_generalisation,fig:weatherbench_generation} in \cref{subsec_app:generation}).




\paragraph{\themethod is resilient over long horizons.} The true test of a world model, however, is its resilience over extended rollouts. In \cref{subsec_app:generation}, we push models trained on mere 20-step sequences to forecast 1,000 steps into the future. Under such a long horizon, the pixel-space LAPO baseline rapidly fails as it converges to a permanent dark figure, while the standard WM fails to commit to distinct digit identities with its mean-seeking average. In contrast, \themethod's formulation maintains structural integrity and object identity across the majority of its steps, suggesting that weight-space dynamics may be a necessary antidote to long-horizon mode collapse (see \cref{fig:long_horizon_app,fig:long_horizon,tab:lh_metrics}).

\paragraph{\themethod induces a hierarchy of representations.}

We aim to provide empirical evidence that \themethod learns the structural hierarchy of background, foreground, and motion (denoted by $\bar{\bm{z}}, \bm{z}_t, \bm{u}_t$, respectively) without any supervision on the semantic meaning of these components. First, in all Moving MNIST sequences \cite{srivastava2015unsupervised}, the background is uniformly dark; our study finds that this background is absorbed into $\bar{\bm{z}}$, as its direct rendering leads to black pixels (see \cref{fig:background_zbar} in \cref{sec_app:results}). It therefore remains to be shown that $\bm{z}_t$ encodes the identity of dynamic objects, and that $\bm{u}_t$ dictates their motion rather than their appearance.

\begin{figure}[htbp]
    \centering
    
    \begin{minipage}[t]{0.465\textwidth}
        \vspace{0pt}
        \centering
        \includegraphics[width=\linewidth]{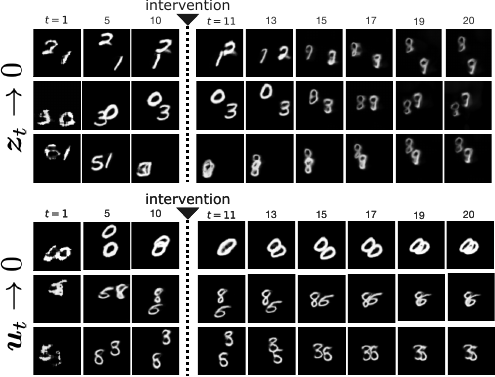}
    \end{minipage}%
    \hfill
    \begin{minipage}[t]{0.515\textwidth}
        \vspace{0pt}
        \centering
        \includegraphics[width=\linewidth, trim=0.6cm 0cm 0cm 0cm, clip]{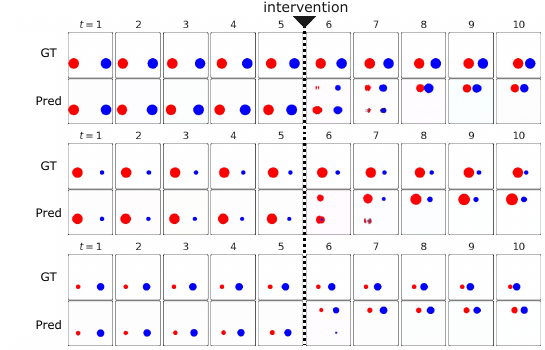}
    \end{minipage}
    
    \vspace{10pt} 

    \begin{minipage}[t]{0.465\textwidth}
        \vspace{0pt}
        \caption{\textbf{Foreground and motion editing.} The top three rows, with $\bm{z}_t \rightarrow 0$, gradually morph digits in all sequences into $8$s and $9$s. The bottom three rows, with $\bm{u}_t \rightarrow 0$, force the digits to converge to the centre of the canvas.}
        \label{fig:morphing_control_mnist}
    \end{minipage}%
    \hfill
    \begin{minipage}[t]{0.515\textwidth}
        \vspace{0pt}
        \caption{\textbf{Zero-shot motion retargeting via latent action substitution.} By replacing the native action on several PhyWorld sequences with an alien action after 5 steps, the model forces the original objects to inherit alien, ``unphysical'' trajectories.}
        \label{fig:action_swapping_phyworld}
    \end{minipage}
    
\end{figure}

To show this, we design an experiment demonstrating the possibility of editing the identity without impacting the dynamics.  Starting at $t=11$, we compute a convex combination of $\bm{z}_t$ and $\mathbf{0}$ such that $\bm{z}_t \xrightarrow{\scalebox{0.8}{$\scriptscriptstyle t \to 20$}} \mathbf{0}$, while leaving GCM-extracted actions $\bm{u}_t$ free. As we observe in the top three rows on \cref{fig:morphing_control_mnist}, digit identities gradually morph towards a resting state ($8$s and $9$s), all the while motion trajectories are preserved. We emphasise that this is achieved without any training supervision on digit identity. This finding is complemented by \cref{fig:fork_comparison}, which showed that abruptly replacing $\bm z_t$ with an alien state 
does not perturb the original dynamics. These experiments establish that the $A$ network acts as an object detector that processes foreground necessarily encoded in its only input $\bm z_t$.

Similarly, we interpolate $\bm{u}_t \xrightarrow{\scalebox{0.8}{$\scriptscriptstyle t \to 20$}} \mathbf{0}$ while leaving $\bm{z}_t$ free. As observed in the bottom three rows of \cref{fig:morphing_control_mnist}, digits decelerate and converge towards the canvas centre, while their visual identities are fully preserved throughout. This suggests that $B$ is the component that processes instructions, necessarily encoded in its input $\bm u_t$. 
Together, these results underscore the capacity to intervene at arbitrary steps of the generation process.

\subsection{Zero-Shot Motion Retargeting and OOD Evaluation on PhyWorld}
\label{subsec:motionretargeting}

To rigorously evaluate the limits of \themethod's latent representation, we test it on the PhyWorld 30K collision validation set \cite{kang2025how}, which introduces ball radii and velocities far outside the training distribution.  In this problem, the identity of a ball refers to its size, colour, and to some extent its $y$-position, which is constant throughout the video (see \cref{sec_app:limitations}). In \cref{subsec_app:phyworld_gen}, we show that our model is \textbf{robust to latent state interventions} and can recalculate physically accurate post-collision velocities using the alien quantities. We ultimately show that latent actions are the central quantity for preserving physical consistency. 


While the model maintains accurate collision dynamics across both in- and out-of-distribution settings, our primary interest lies in its capacity for action-controllable generation.

\paragraph{\themethod can abruptly retarget motion via action substitution.} We test the orthogonality of our latent spaces by performing mid-sequence motion retargeting. We allow the system to evolve normally up to step $t=5$, and then we abruptly substitute the native latent action $\bm{u}_t$ with an alien action extracted from a different sequence. While this is similar to moving digits depicted in \cref{fig:morphing_control_mnist}, we note that this abrupt intervention on PhyWorld, which we term as \textbf{zero-shot motion retargeting}, involves rigorous physics, raising important questions about the conservation of physical quantities such as momentum (Mom.) and kinetic energy (KE) \cite{kang2025how}. 

Our results in \cref{fig:action_swapping_phyworld} indicate that our intervention forces the original balls to quickly adopt the motion trajectories of the alien sequence within less than two steps, deliberately violating the momentum conservation expected of their original masses.
Crucially, the structural identity of the balls (radius and colour) remains perfectly preserved despite the unphysical motion. 
\themethod's ability to strictly obey the injected action without leaking structural features from the source sequence suggests orthogonality between weight-space identity and dynamic intent. As it allows users to seamlessly and privately\footnote{This is private since it does not leak the identity of the objects that initiated the actions.} map arbitrary trajectories onto existing objects without corrupting visual identity, motion retargeting with unphysical dynamics is desirable, especially for applications involving digital content manipulation, stylistic or editorial control.


\paragraph{Ablation: additive vs. joint forward dynamics.} In \cref{tab:phyworld_phase3_ood}, we quantify the trade-offs of our decomposed architecture by comparing our additive FDM against a standard monolithic MLP that processes the concatenated vector $[\bm{z}_t, \bm{u}_t]$ jointly. While the joint architecture achieves lower physical errors on OOD configurations, it fundamentally destroys the content/dynamics decoupling required for the motion retargeting demonstrated above. Thus, our additive formulation accepts a meaningful but acceptable trade-off in exchange for disentanglement.

\begin{table*}[t]
\centering
\caption{Comparison of additive and joint FDM performance on in-distribution (InD) and out-of-distribution (OOD) PhyWorld data. Samples are phase 3 videos generated from 3 context frames. Values are reported as mean $\pm$ standard deviation over the tested sequences.}
\label{tab:phyworld_phase3_ood}
\resizebox{\textwidth}{!}{%
\begin{tabular}{ll ccc ccc}
\toprule
\multirow{2}{*}{\textbf{Distribution}} & \multirow{2}{*}{\textbf{Model}} & \multicolumn{3}{c}{\textbf{Image Metrics}} & \multicolumn{3}{c}{\textbf{Physical Metrics}} \\
\cmidrule(lr){3-5} \cmidrule(lr){6-8}
& & \makecell{\textbf{MSE} ($\times 10^{-2}$) $\downarrow$} & \textbf{PSNR} $\uparrow$ & \textbf{SSIM} $\uparrow$ & \makecell{\textbf{Pos.} ($\times 10^{-2}$) $\downarrow$} & \textbf{Mom.} $\downarrow$ & \textbf{KE} $\downarrow$ \\
\midrule
\multirow{2}{*}{InD}
& Additive & $0.98 \pm 0.71$ & $23.12 \pm 2.69$ & $0.9404 \pm 0.0219$ & $2.61 \pm 2.19$ & $0.0548 \pm 0.0408$ & $0.0291 \pm 0.1317$ \\
& Joint    & $1.08 \pm 0.69$ & $22.38 \pm 2.04$ & $0.9358 \pm 0.0208$ & $2.67 \pm 2.53$ & $0.0582 \pm 0.0714$ & $0.0456 \pm 0.2112$ \\
\midrule
\multirow{2}{*}{OOD}
& Additive & $3.20 \pm 2.85$ & $18.83 \pm 4.81$ & $0.9015 \pm 0.0583$ & $12.50 \pm 8.99$ & $0.4025 \pm 0.5581$ & $0.5351 \pm 1.7191$ \\
& Joint    & $2.99 \pm 2.30$ & $17.83 \pm 3.61$ & $0.9015 \pm 0.0547$ & $9.67 \pm 7.90$ & $0.2093 \pm 0.2243$ & $0.0815 \pm 0.2331$ \\
\bottomrule
\end{tabular}
}
\end{table*}

\subsection{Zero-Shot Super-Resolution for Global Temperature Prediction}
\label{sec:superresolution}





High-quality visuals are crucial for immersive virtual experiences and for detailed scientific analysis. In this section, we ask: \emph{can we extract fine-grained meteorological insights from weather models simply by altering the rendering resolution?} To investigate this, we leverage \themethod's continuous spatial formulation. During training, the model evaluates the spatial field by sampling a small standard coordinate grid of $H \times W=32\times 64$ points. At inference,\footnote{To focus on upscaling quality, we disregard the GCM and rely on actions issued by the IDM to drive the video generation.} however, we introduce a scaling factor $s$ and resample the grid to a denser $(s \cdot H) \times (s \cdot W)$ resolution. When $s \gg 1$, this operation yields \textbf{zero-shot super-resolution} without requiring any specialised upsampling networks.

However, merely querying this denser grid risks aliasing if the network evaluates high-frequency features unsupported by the training resolution \cite{barron2021mip,lindell2022bacon}. To guarantee artefact-free upscaling, we implement a strict and simple \textbf{frequency-masking procedure} bounded by the Nyquist--Shannon theorem. By dynamically zeroing out any Fourier basis frequencies exceeding the Nyquist limit of the original training grid (e.g., $\xi^{y}_{\text{Nyquist}} = \sfrac{1}{2\Delta y}$), we ensure the continuous field remains securely within its learned, unaliased manifold (detailed in \cref{app:aliasing}).


\begin{figure}[h]
\centering
\includegraphics[width=0.83\linewidth]{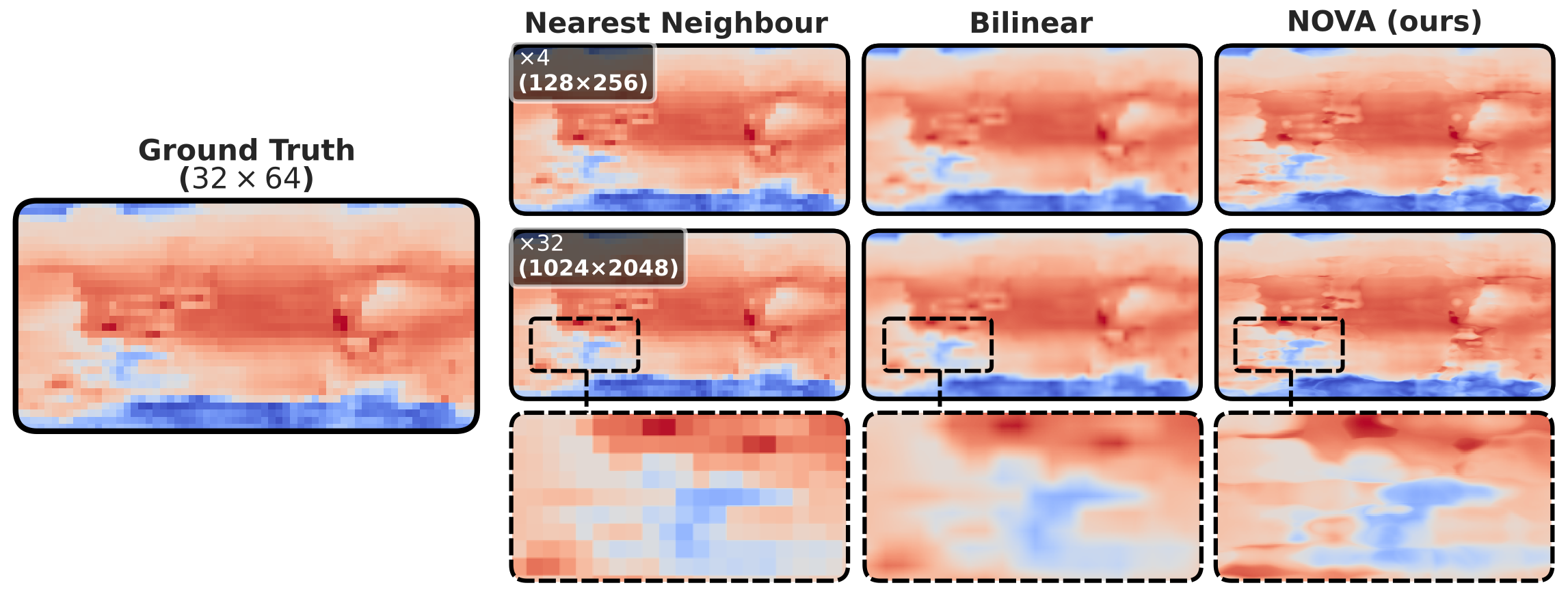}
\caption{\textbf{Zero-shot super-resolution on WeatherBench.} Ground truth at native $32\times 64$ resolution next to nearest-neighbour, bilinear interpolation, and \themethod at $\times 4$ and $\times 32$ scaling. \themethod avoids aliasing and preserves macro-structure without hallucinating high-frequency textures.}
\label{fig:superresolution}
\end{figure}

\Cref{fig:superresolution} compares \themethod against nearest-neighbour and bilinear interpolation baselines on a single WeatherBench frame \cite{rasp2020weatherbench} at three scaling factors. Our INR-based approach avoids the spurious discontinuities of nearest-neighbour and the artificial diffusion of bilinear interpolation at both downsampling and extreme upsampling scales.\footnote{While \themethod natively supports upsampling to arbitrarily high resolutions, we cap our visualisation at $\times 32$, as further magnification yields diminishing perceptual returns. We show downsampled temperature maps in \cref{fig:superresolution_unmasked}.}

\begin{wraptable}{r}{0.52\textwidth}
\vspace{-1.3em}
\small
\centering
\caption{Super-resolution ($\times32$) evaluation across the test set. We employ a stratified sampling strategy over 69 batches to assess distributional and structural faithfulness to the original resolution. \themethod consistently achieves the lowest distributional drift.}
\label{tab:distributional_metrics}
\setlength{\tabcolsep}{4pt}
\resizebox{\linewidth}{!}{
\begin{tabular}{lccc}
\toprule
\textbf{Metric} & \textbf{NN} & \textbf{Bilinear} & \textbf{NOVA} \\
\midrule
$W_1$ $\downarrow$ & $0.137 \pm 0.016$ & $0.140 \pm 0.015$ & $\mathbf{0.135 \pm 0.015}$ \\
JSD $\downarrow$ & $0.229 \pm 0.017$ & $0.221 \pm 0.022$ & $\mathbf{0.213 \pm 0.018}$ \\
$B$ $\downarrow$ & $0.056 \pm 0.009$ & $0.053 \pm 0.012$ & $\mathbf{0.048 \pm 0.009}$ \\
$d_{\mathcal{F}}$ $\downarrow$ & $\mathbf{0.427 \pm 0.065}$ & $\mathbf{0.427 \pm 0.065}$ & $\mathbf{0.427 \pm 0.067}$ \\
\bottomrule
\end{tabular}
}
\vspace{-1em}
\end{wraptable}

Quantitative results in \Cref{tab:distributional_metrics} compare distributional fidelity and state preservation metrics across all strategies. We report mean and standard deviation metrics across all 24 time steps and across the batch dimension, focusing on the $\times 32$ upsampling resolution. While we observe that these distances are inherently narrow, the key insight is consistency: despite the tight margins, \themethod systematically matches or outperforms both nearest-neighbour (NN) and bilinear interpolation across every empirical distribution metric. Taken together with qualitative insights from \cref{fig:superresolution}, it follows that, simply by re-rendering, \themethod can enhance the sharp, highly dynamic temperature distributions of evolving weather fronts without incurring artefacts. This capability has massive implications for localised extreme weather forecasting, where such precision impacts the lives of massive populations.

\section{Conclusion}
\label{sec:conclusion}

We introduced \themethod, a world model and video generation framework that operates entirely in the weight space of an INR. By leveraging an analytical INR renderer, \themethod avoids the decoder bottleneck and achieves native resolution independence. Furthermore, it decouples background and foreground from motion substitution, enabling zero-shot latent state (and action) editing. Through experiments on diverse datasets, we demonstrate strong context-conditioned controllability and forecasting, operating at around 40M parameters on a single consumer GPU.
Because the conceptual mechanism underlying \themethod's editing capabilities is applicable to standard latent action models, it could yield substantial benefits if integrated with mainstream world models.
We foresee applications in various downstream tasks for digital artists and animators, for whom this uncompromising separation of motion and form could unlock a new paradigm of limitless expressive control.




\paragraph{Limitations and Future Work.} While \themethod excels at controllable generation, its formulation leaves room for extensions: ($i$) continuous actions can include high-frequency artefacts (e.g., brightness), necessitating further regularisations for purer decoupling; ($ii$) OOD generalisation remains challenging, partly because \themethod does not feature the ability to leverage differential equations; and ($iii$) scaling to higher-dimensional INRs may demand parameter-efficient adaptations. We detail these limitations in \cref{sec_app:limitations}, highlighting them as exciting avenues for future work.

\subsection*{Broader Impact}
Weight-space world models offer a more interpretable alternative to opaque abstract latent spaces by separating geometric semantics into background, content, and motion. While this transparency is beneficial for safety-critical applications (e.g., when auditing model behaviour, identifying biases, or applying safe interventions for self-driving cars or robots), it also introduces risks. The ability to intervene and edit video content could be exploited by bad actors to generate highly realistic deepfakes, manipulate visual evidence, and spread misinformation. To mitigate these dangers and promote community-driven oversight, we fully open-source the code, datasets, and model weights for both our proposed framework and the implemented baselines: \rdn{\url{https://github.com/ddrous/nova}}.


\subsection*{Acknowledgements}
This work was supported by UK Research and Innovation grant EP/S022937/1: Interactive Artificial Intelligence. The first author thanks Bahareh Tolooshams and Enrique Crespo-Fernandez for informative early discussions on world modelling, along with Kelly Sumelong for key questions that informed the final experiments.

\bibliography{refs}

\appendix
\newpage
\DoToC

\newpage

\section{Related Work}
\label{sec:related}

Our framework shares similarities with distinct fields and frameworks within the deep learning literature. We detail these major intersections below, while providing in \cref{tab:capability_comparison} a high-level summary of how \themethod positively distinguishes itself.

\paragraph{World models and video generation.} World models \cite{li2025comprehensive,zhang2025overcoming} have seen a massive surge in popularity in recent years. 
The classic paradigm of \cite{ha2018world} uses a variational autoencoder to compress observations and an RNN to model latent transitions. DreamerV3~\cite{hafner2020dream} scales this with categorical representations and a joint-embedding world model. IRIS~\cite{micheli2022transformers} replaces the continuous latent with discrete tokens processed by a Transformer. Genie~\cite{bruce2024genie} and V-JEPA~\cite{bardes2024vjepa} learn from unlabelled video via latent-only objectives. These models all rely on a learned decoder. A complementary direction~\cite{he2025pre} fine-tunes video generation models to behave as world models. \themethod goes the other direction, as it starts with world modelling and recovers video generation capability through the GCM. LeWorldModel~\cite{maes2026leworldmodel} demonstrates the benefits of end-to-end training without freezing encoders; \themethod achieves similar benefits by allowing for the possibility to combine its first two training phases, while additionally eliminating the decoder.

\begin{figure}[htbp]
\centering
\resizebox{0.96\textwidth}{!}{%
\begin{tikzpicture}[
    node distance=0.8cm and 1cm, 
    base_arrow/.style={draw=blue!10!black, line width=1.5pt, >={Triangle[length=2.3mm, width=2.3mm, round]}, rounded corners=2mm},
    arrow/.style={->, base_arrow},
    block/.style={
        rectangle, rounded corners=2mm, align=center,
        minimum height=3.5em, minimum width=5.5em, font=\Large\bfseries,
        line width=1.2pt
    },
    data/.style={
        rectangle, rounded corners=1.5mm, align=center,
        minimum height=2.8em, minimum width=3.5em,
        fill=white, draw=blue!30!black, text=black!90,
        font=\LARGE\bfseries,
        line width=1.2pt
    },
    enc_style/.style={block, fill=red!15, draw=red!60!black},
    idm_style/.style={block, fill=teal!15, draw=teal!60!black},
    fdm_style/.style={block, fill=cyan!15, draw=cyan!70!black},
    stop gradient/.style={
        postaction={decorate},
        decoration={markings, mark=at position #1 with {
            \fill[white] (-4pt, -8pt) rectangle (6pt, 8pt);
            \draw[line width=1.5pt, blue!40!black, -] (-2pt,-6pt) -- (2pt,6pt);
            \draw[line width=1.5pt, blue!40!black, -] (2pt,-6pt) -- (6pt,6pt);
        }}
    }, stop gradient/.default=0.35
]
    \node[data] (xt1_gt) {$\mathbf{o}_{t+1}$};
    \node[enc_style, right=of xt1_gt] (encoder_gt) {Encoder};
    \node[data, right=of encoder_gt] (thetat1_gt) {$\bm{z}_{t+1}$};
    
    \node[data, below=1.2cm of xt1_gt] (xt) {$\mathbf{o}_t$};
    \node[enc_style, right=of xt] (encoder) {Encoder};
    \node[data, right=of encoder] (thetat) {$\bm{z}_t$};
    
    \node[idm_style, right=of thetat] (idm) {IDM};
    \node[data, right=of idm] (at) {$\bm{u}_t$};
    
    \node[fdm_style, below=1.2cm of at] (fdm) {FDM};
    \node[data, right=of fdm] (thetat1) {$\hat{\bm{z}}_{t+1}$};
    \node[enc_style, right=of thetat1] (decoder) {Decoder};
    \node[data, right=of decoder] (xt1) {$\hat{\mathbf{o}}_{t+1}$};

    \draw[arrow] (xt) -- (encoder);
    \draw[arrow] (encoder) -- (thetat);
    \draw[arrow] (xt1_gt) -- (encoder_gt);
    \draw[arrow, stop gradient] (encoder_gt) -- (thetat1_gt);
    
    \draw[arrow] (thetat) -- (idm);
    \draw[arrow] (thetat1_gt) -| (idm);
    \draw[arrow] (idm) -- (at);
    
    \draw[arrow] (thetat) |- (fdm);
    \draw[arrow] (at) -- (fdm);
    \draw[arrow] (fdm) -- (thetat1);
    
    \draw[arrow] (thetat1) -- (decoder);
    \draw[arrow] (decoder) -- (xt1);


\end{tikzpicture}
}
\caption{\textbf{Standard latent action model (LAM).} The IDM computes the action from consecutive latent states during training. This approach relies on a spatial decoder to project the latent state back into pixel space, creating a resolution bottleneck (e.g., \cite{schmidt2023learning}). Note that the encoder and the decoder can be identity mappings as well, to match the setting of \cite{zhang2025latent}.}
\label{fig:naive_model}
\end{figure}
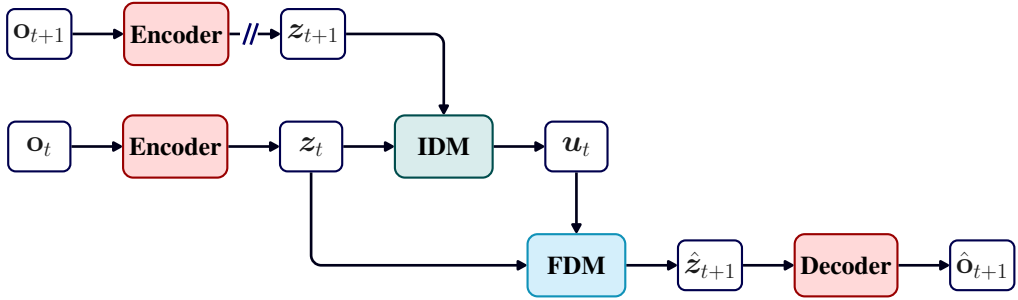

\paragraph{Latent action models (LAMs).} Similar to reinforcement learning \cite{sutton1998reinforcement}, world models typically require explicit actions to inform their dynamics. 
When ground-truth actions are unavailable, latent action models (as described in \cref{fig:naive_model})  infer these from consecutive observations. LAPO~\cite{schmidt2023learning} and LAPA~\cite{ye2024latent} learn inverse dynamics models that extract latent actions bridging consecutive states. Moto~\cite{chen2025moto} and AdaWorld~\cite{gao2025adaworld} incorporate such models into robot learning pipelines. \citet{garrido2026learning} demonstrate that continuous latent actions, although sparse and noisy, capture for in-the-wild actions learning, whereas discrete codebooks struggle. For deployment without future frames, RSSMs~\cite{hafner2020dream}, IRIS~\cite{micheli2022transformers}, and CoWVLA~\cite{yang2026chain} use autoregressive models to generate actions. Our GCM follows this principle using a causal sequence model.

\paragraph{Disentanglement in video prediction.}
Decades of work aim to separate content from motion in video. DrNET~\cite{denton2017unsupervised} uses adversarial training to decompose frames into pose and content codes. DDPAE~\cite{hsieh2018learning} employs structured probabilistic models to disentangle components with low-dimensional temporal dynamics. PhyDNet~\cite{guen2020disentangling} introduces a PDE-constrained recurrent cell that explicitly separates physical dynamics from residual uncertainty. More recent work~\cite{abdulsalam2026lamo} disentangles static from first-order dynamic changes in a latent motion model. Critically, all these approaches impose disentanglement through auxiliary objectives or explicit structural priors applied on top of abstract latent spaces, and still require a learned decoder. In \themethod, disentanglement is a consequence of the weight-space geometry itself, requiring no additional loss terms.

Our work is particularly motivated by the fact that patch-based video representations \cite{lin2020improving}, which power the majority of current video and world models, typically lack the human-style intuitiveness of object-centric decompositions \cite{daniel2026latent}. Because they struggle to capture such clean relationships between objects (which leaves them uninterpretable), they underperform, especially in out-of-distribution scenarios \cite{daniel2026latent}. Our work closes this gap by proposing weight-based representations, positioning itself as a crucial step for understanding complex scenes.

\paragraph{Implicit Neural Representations.}
INRs~\cite{sitzmann2020implicit} parameterise continuous signals via the weights of a coordinate-based MLP. Fourier features~\cite{tancik2020fourier} enable learning of high-frequency components. \citet{dupont2022data} introduce the term ``functa'', showing that meta-learned \cite{zintgraf2019fast,nzoyem2025neural} INR weight vectors can serve as data representations for downstream tasks. In the 3D domain, 3D Gaussian Splatting~\cite{kerbl20233d} and work such as GWM~\cite{lu2025gwm} exploit \emph{structured} spatial representations for world modelling. Our work extends this to 2D video dynamics, using the INR weights as the latent state that evolves under recurrent dynamics.



A fundamental advantage of INRs is their continuous formulation, encoding data as functional mappings rather than discrete spatial grids~\cite{essakine2024we}. This paradigm theoretically enables the representation of complex, high-resolution details with exceptional memory efficiency, circumventing the storage bottlenecks typical of high-dimensional video data. However, despite this compact capacity, standard INRs are notoriously susceptible to severe aliasing artefacts when evaluated at high resolutions \cite{barron2021mip}. Addressing this tension between memory efficiency and alias-free high-fidelity rendering is a central consideration in our work.

\paragraph{Weight-space sequence learning.}
\cite{li2025weightflow,nzoyem2026weightspace} demonstrate that the hidden state of a network can be parameterised as the weights of an auxiliary network, enabling (linear) weight-space sequence modelling. WeightCaster~\cite{nzoyem2026outofsupport} extends weight-space methods to out-of-support generalisation. \themethod is a world model that can be viewed as a non-linear generalisation of WARP \cite{nzoyem2026weightspace}. It mainly replaces the globally linear recurrence with a decomposed MLP dynamics model, and adds the base-weight stabilisation mechanism that makes training on complex visual signals tractable. Although designed for different applications, we compare both frameworks in further detail in \cref{sec:comparison_wsl}.


\begin{table}[t]
\centering
\caption{\textbf{Capability comparison.} \cmark = yes; \xmark = no; \pmark = partial or config-dependent. $\dagger$ = requires known action labels at training time. Entries reflect standard published configurations. \themethod is uniquely positioned among representative baselines by simultaneously achieving decoder-free and resolution-independent rendering, latent content-motion disentanglement, zero-shot super-resolution, and action-label-free training directly from video data, all while remaining efficient enough to be trained on a single consumer GPU.}
\label{tab:capability_comparison}
\vspace{4pt}
\setlength{\tabcolsep}{3pt}
\resizebox{\textwidth}{!}{%
\begin{tabular}{lcccccccccc}
\toprule
 & \rotatebox{72}{\themethod (ours) [\cref{fig:nova_method}]} & \rotatebox{72}{Standard WM [\cref{fig:naive_model}]} & \rotatebox{72}{WARP~\cite{nzoyem2026weightspace}} & \rotatebox{72}{DreamerV3$^\dagger$~\cite{hafner2020dream}} & \rotatebox{72}{GENIE~\cite{bruce2024genie}} & \rotatebox{72}{V-JEPA~\cite{bardes2024vjepa}} & \rotatebox{72}{IRIS$^\dagger$~\cite{micheli2022transformers}} & \rotatebox{72}{PredRNN~\cite{wang2022predrnn}} & \rotatebox{72}{PhyDNet~\cite{guen2020disentangling}} & \rotatebox{72}{LeWorldModel$^\dagger$~\cite{maes2026leworldmodel}} \\
\midrule
\multicolumn{11}{l}{\textit{Architecture}} \\
Decoder-free rendering & \cmark & \xmark & \cmark & \xmark & \xmark & \cmark & \xmark & \xmark & \xmark & \cmark \\
Resolution-independent rendering & \cmark & \xmark & \cmark & \xmark & \xmark & \xmark & \xmark & \xmark & \xmark & \xmark \\
Single-GPU training & \cmark & \cmark & \cmark & \xmark & \xmark & \xmark & \cmark & \cmark & \cmark & \xmark \\
\midrule
\multicolumn{11}{l}{\textit{Disentanglement \& Control}} \\
Latent content--motion disentanglement & \cmark & \pmark & \xmark & \xmark & \xmark & \xmark & \xmark & \xmark & \xmark & \xmark \\
Structural $A/B$ dynamics decomposition & \cmark & \pmark & \xmark & \xmark & \xmark & \xmark & \xmark & \xmark & \pmark & \xmark \\
Zero-shot content/motion retargeting & \cmark & \pmark & \pmark & \xmark & \xmark & \xmark & \xmark & \xmark & \xmark & \xmark \\
Latent action model (actions from video) & \cmark & \cmark & \pmark & \xmark & \cmark & \xmark & \xmark & \xmark & \xmark & \xmark \\
Continuous action space & \cmark & \cmark & \cmark & \cmark & \xmark & \xmark & \xmark & \xmark & \xmark & \cmark \\
\midrule
\multicolumn{11}{l}{\textit{Generation}} \\
Zero-shot super-resolution & \cmark & \xmark & \cmark & \xmark & \xmark & \xmark & \xmark & \xmark & \xmark & \xmark \\
Context-conditioned forecasting & \cmark & \cmark & \cmark & \cmark & \cmark & \cmark & \cmark & \cmark & \cmark & \cmark \\
Autonomous generation (no test-time actions) & \cmark & \cmark & \pmark & \cmark & \cmark & \cmark & \cmark & \cmark & \cmark & \cmark \\
Artefact-free rendering & \cmark & \pmark & \cmark & \pmark & \pmark & \cmark & \pmark & \pmark & \pmark & \cmark \\
\midrule
\multicolumn{11}{l}{\textit{Domain coverage}} \\
Continuous geospatial physics & \cmark & \cmark & \cmark & \xmark & \xmark & \pmark & \xmark & \cmark & \cmark & \xmark \\
Goal-directed discrete control & \cmark & \cmark & \pmark & \cmark & \cmark & \xmark & \cmark & \xmark & \xmark & \cmark \\
\bottomrule
\end{tabular}}
\end{table}

\section{Methodological Details}

\subsection{Motivation: INRs and The Decoder Bottleneck}



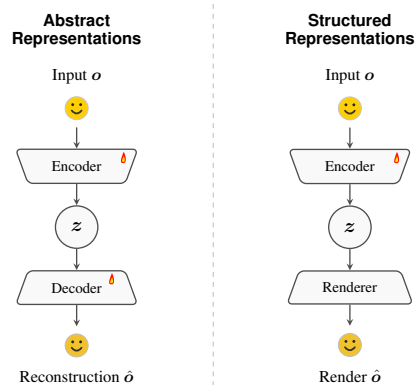
\begin{wrapfigure}[19]{R}{0.49\textwidth}
  \vspace{-10pt}
  \centering
  \resizebox{0.80\linewidth}{!}{%
  \begin{tikzpicture}[
    >=stealth,
    node distance=0.4cm and 3.2cm,
    box/.style={draw=black!70, thick, rounded corners=3pt, minimum height=0.7cm, align=center, fill=gray!5, font=\small},
    trap-c/.style={box, trapezium, minimum width=2.6cm, trapezium left angle=110, trapezium right angle=110},
    trap-e/.style={box, trapezium, minimum width=2.6cm, trapezium left angle=70, trapezium right angle=70},
    circ/.style={draw=black!70, thick, circle, minimum size=0.9cm, fill=gray!5, font=\large\bfseries},
    arr/.style={->, thick, draw=black!70}
  ]
  \newcommand{\SimpleSmiley}[2]{
    \begin{scope}[shift={(#1)},scale=#2]
      \fill[yellow!65!orange] (0,0) circle(0.61);
      \draw[gray!45, line width=0.55pt] (0,0) circle(0.61);
      \fill[black!85] (-0.21, 0.19) circle(0.077);
      \fill[black!85] ( 0.21, 0.19) circle(0.077);
      \draw[black!80, line width=1.1pt, line cap=round]
        (-0.26,-0.12) ..controls(-0.09,-0.33) and (0.09,-0.33).. (0.26,-0.12);
    \end{scope}
  }
  \newcommand{\DegradedSmiley}[2]{
    \begin{scope}[shift={(#1)},scale=#2]
      \fill[yellow!55!orange!90!gray] (0,0) circle(0.61);
      \draw[gray!50, line width=0.55pt] (0,0) circle(0.61);
      \fill[black!85] (-0.21, 0.19) circle(0.077);
      \fill[black!85] ( 0.21, 0.19) circle(0.077);
      \draw[black!80, line width=1.1pt, line cap=round]
        (-0.26,-0.12) ..controls(-0.09,-0.33) and (0.09,-0.33).. (0.26,-0.12);
    \end{scope}
  }
  \newcommand{\FireIcon}{%
    \tikz[baseline=-0.5ex, scale=0.12]{
      \fill[red!85!orange] (0.5,0) .. controls (1.0,0) and (1.0,1) .. (0.5,2)
        .. controls (0.2,1.3) and (0,0.8) .. (0,0.3)
        .. controls (0,0) and (0.2,0) .. (0.5,0) -- cycle;
      \fill[yellow!90!orange] (0.5,0.2) .. controls (0.75,0.2) and (0.75,0.8) .. (0.5,1.2)
        .. controls (0.3,0.8) and (0.2,0.5) .. (0.2,0.4)
        .. controls (0.2,0.2) and (0.3,0.2) .. (0.5,0.2) -- cycle;
    }%
  }
  \node[align=center, font=\bfseries\sffamily] (titL) {Abstract\\Representations};
  \node[below=0.2cm of titL] (inL-lbl) {Input $\bm{o}$};
  \node[minimum size=0.8cm, below=0.0cm of inL-lbl] (inL-box) {};
  \SimpleSmiley{inL-box.center}{0.4}
  \node[trap-c, below=0.45cm of inL-box] (enc) {Encoder};
  \node at ($(enc.center) + (1.0, 0.15)$) {\FireIcon};
  \node[circ, below=0.45cm of enc] (z) {$\bm{z}$};
  \node[trap-e, below=0.45cm of z] (dec) {Decoder};
  \node at ($(dec.center) + (0.75, 0.15)$) {\FireIcon};
  \node[minimum size=0.8cm, below=0.45cm of dec] (outL-box) {};
  \DegradedSmiley{outL-box.center}{0.4}
  \node[below=0.0cm of outL-box] (outL-lbl) {Reconstruction $\hat{\bm{o}}$};
  \draw[arr] (inL-box.south) -- (enc.north);
  \draw[arr] (enc.south) -- (z.north);
  \draw[arr] (z.south) -- (dec.north);
  \draw[arr] (dec.south) -- (outL-box.north);
  \node[align=center, font=\bfseries\sffamily, right=2.8cm of titL] (titR) {Structured\\Representations};
  \node[below=0.2cm of titR] (inR-lbl) {Input $\bm{o}$};
  \node[minimum size=0.8cm, below=0.0cm of inR-lbl] (inR-box) {};
  \SimpleSmiley{inR-box.center}{0.4}
  \node[trap-c, below=0.45cm of inR-box] (hyp) {Encoder};
  \node at ($(hyp.center) + (1.0, 0.15)$) {\FireIcon};
  \node[circ, below=0.45cm of hyp] (thet) {$\bm{z}$};
  \node[trap-e, below=0.45cm of thet] (rnd) {Renderer};
  \node[minimum size=0.8cm, below=0.45cm of rnd] (outR-box) {};
  \DegradedSmiley{outR-box.center}{0.4}
  \node[below=0.0cm of outR-box] (outR-lbl) {Render $\hat{\bm{o}}$};
  \draw[arr] (inR-box.south) -- (hyp.north);
  \draw[arr] (hyp.south) -- (thet.north);
  \draw[arr] (thet.south) -- (rnd.north);
  \draw[arr] (rnd.south) -- (outR-box.north);
  \coordinate (midL) at ($(titL.center)!0.5!(titR.center)$);
  \coordinate (divT) at (midL |- titL.north);
  \coordinate (divB) at (midL |- outL-lbl.south);
  \draw[dashed, draw=gray!40, thick] ($(divT)+(0,0.1)$) -- ($(divB)-(0,0.1)$);
  \end{tikzpicture}}
  \caption{Two paradigms for signal encoding. \textbf{Abstract} encoders map input $\bm{o}$ to an opaque latent $\bm{z}$ requiring a trained decoder. \textbf{Structured} encoders map $\bm{o}$ to INR weights $\bm{z}$, which are subsequently rendered analytically.}
  \label{fig:repr_paradigms}
\end{wrapfigure}

\Cref{fig:repr_paradigms} contrasts abstract and structured representations. The abstract paradigm has three weaknesses as a world model latent space. First, the decoder typically consumes more parameters than the encoder \cite{bruce2024genie}, making it a computational bottleneck. Second, the abstract latent space has no guaranteed semantic structure since disentangling background from foreground, or content from motion, requires explicit post-training alterations. Third, the renderer is resolution-locked: changing the output resolution requires retraining or expensive upsampling.

The structured paradigm eliminates these problems. The INR renderer is a fixed mathematical function, since querying any spatial coordinates requires no parameters. The weight vector $\bm{z}$ has a direct interpretation as it parameterises a continuous spatial function. And the continuous domain means we can query at any resolution by simply changing the coordinate grid.

By leveraging Fourier features \cite{tancik2020fourier}, these representations provide a differentiable, 3D-consistent bridge between latent space and pixels. Fourier mapping allows the network to overcome spectral bias, capturing the high-frequency details and sharp gradients (e.g., fine textures, localised weather fronts) that standard MLPs often blur. This continuous, analytical framework enabled zero-shot super-resolution. Such scalability is critical for applications like global weather forecasting, where a single pixel must represent vastly different scales without loss of structural integrity.

Finally, structured representations can be significantly more compact than raw pixel arrays \cite{mostajeran2025context}. By encoding complex data into 1D weight vectors, they can be processed with generic MLPs rather than specialised architectures like CNNs or RNNs, thus simplifying the pipeline.





\subsection{Architecture \& Hyperparameters}
\label{app:architecture}

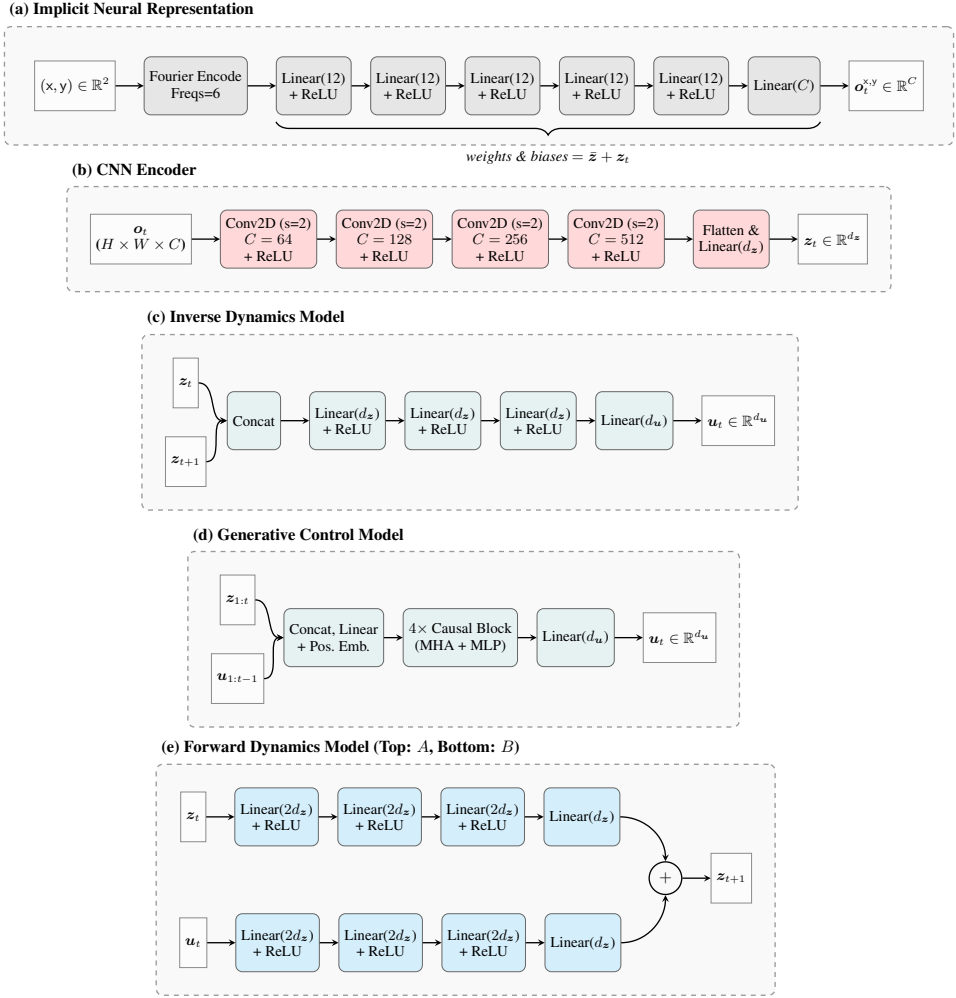
\begin{figure}[h]
\centering
\resizebox{0.9\textwidth}{!}{%
\begin{tikzpicture}[
    node distance=0.8cm and 0.5cm,
    layer/.style={rectangle, draw=black!60, fill=blue!5, rounded corners, align=center, minimum height=3.5em, font=\small},
    tensor/.style={rectangle, draw=grey!60!black, fill=grey!10, align=center, minimum height=3em, font=\small\bfseries},
    arrow/.style={->, thick, >=stealth}
]

  \node[tensor] (root_in) at (-0.95,+3.5) {$(\mathsf{x},\mathsf{y}) \in \mathbb{R}^2$};
  \node[layer, right=0.6cm of root_in, fill=black!10] (fourier) {Fourier Encode\\Freqs=6};
  \node[layer, right=0.6cm of fourier, fill=black!10] (r1) {Linear(12)\\+ ReLU};
  \node[layer, right=0.4cm of r1, fill=black!10] (r2) {Linear(12)\\+ ReLU};
  \node[layer, right=0.4cm of r2, fill=black!10] (r3) {Linear(12)\\+ ReLU};
  \node[layer, right=0.4cm of r3, fill=black!10] (r4) {Linear(12)\\+ ReLU};
  \node[layer, right=0.4cm of r4, fill=black!10] (r5) {Linear(12)\\+ ReLU};
  \node[layer, right=0.4cm of r5, fill=black!10] (r6) {Linear($C$)};
  \node[tensor, right=0.6cm of r6] (root_out) {$\bm{o}_t^{\mathsf{x},\mathsf{y}} \in \mathbb{R}^C$};
  \path[arrow] (root_in) edge (fourier) (fourier) edge (r1) (r1) edge (r2) (r2) edge (r3) (r3) edge (r4) (r4) edge (r5) (r5) edge (r6) (r6) edge (root_out);
  \draw[decorate, decoration={brace, amplitude=8pt, mirror}, thick]
      ([yshift=-4pt]r1.south west) -- ([yshift=-4pt]r6.south east)
      node[midway, below=14pt, font=\small\itshape]
          {weights \& biases $= \bar{\bm{z}} + \bm{z}_t$};

  \node[tensor] (cnn_in) at (0.45,0.25) {$\bm{o}_t$\\($H \times W \times C$)};
  \node[layer, right=0.6cm of cnn_in, fill=red!15] (c1) {Conv2D (s=2)\\$C=64$\\+ ReLU};
  \node[layer, right=0.4cm of c1, fill=red!15] (c2) {Conv2D (s=2)\\$C=128$\\+ ReLU};
  \node[layer, right=0.4cm of c2, fill=red!15] (c3) {Conv2D (s=2)\\$C=256$\\+ ReLU};
  \node[layer, right=0.4cm of c3, fill=red!15] (c4) {Conv2D (s=2)\\$C=512$\\+ ReLU};
  \node[layer, right=0.6cm of c4, fill=red!15] (flat) {Flatten \&\\Linear($d_{\bm{z}}$)};
  \node[tensor, right=0.6cm of flat] (cnn_out) {$\bm{z}_t \in \mathbb{R}^{d_{\bm{z}}}$};
  \path[arrow] (cnn_in) edge (c1) (c1) edge (c2) (c2) edge (c3) (c3) edge (c4) (c4) edge (flat) (flat) edge (cnn_out);

  \node[tensor] (idm_in1) at (1.4,-2.8) {$\bm{z}_t$};
  \node[tensor] (idm_in2) [below=0.6cm of idm_in1] {$\bm{z}_{t+1}$};
  \node[layer, right=0.6cm of idm_in1, yshift=-0.8cm, fill=teal!10] (idm_cat) {Concat};
  \node[layer, right=0.6cm of idm_cat, fill=teal!10] (idm_l1) {Linear($d_{\bm{z}}$)\\+ ReLU};
  \node[layer, right=0.4cm of idm_l1, fill=teal!10] (idm_l2) {Linear($d_{\bm{z}}$)\\+ ReLU};
  \node[layer, right=0.4cm of idm_l2, fill=teal!10] (idm_l3) {Linear($d_{\bm{z}}$)\\+ ReLU};
  \node[layer, right=0.4cm of idm_l3, fill=teal!10] (idm_l4) {Linear($d_{\bm{u}}$)};
  \node[tensor, right=0.6cm of idm_l4] (idm_out) {$\bm{u}_t \in \mathbb{R}^{d_{\bm{u}}}$};
  \path[arrow] (idm_in1) edge[out=0,in=180] (idm_cat) (idm_in2) edge[out=0,in=180] (idm_cat);
  \path[arrow] (idm_cat) edge (idm_l1) (idm_l1) edge (idm_l2) (idm_l2) edge (idm_l3) (idm_l3) edge (idm_l4) (idm_l4) edge (idm_out);


  \node[tensor] (dmm_in1) at (2.5,-7.4) {$\bm{z}_{1:t}$};
  \node[tensor] (dmm_in2) [below=0.6cm of dmm_in1] {$\bm{u}_{1:t-1}$};
  \node[layer, right=0.6cm of dmm_in1, yshift=-0.8cm, fill=teal!10] (dmm_cat) {Concat, Linear\\+ Pos.\ Emb.};
  \node[layer, right=0.4cm of dmm_cat, fill=teal!10] (dmm_tf) {$4\times$ Causal Block\\(MHA + MLP)};
  \node[layer, right=0.4cm of dmm_tf, fill=teal!10] (dmm_proj) {Linear($d_{\bm{u}}$)};
  \node[tensor, right=0.6cm of dmm_proj] (dmm_out) {$\bm{u}_t \in \mathbb{R}^{d_{\bm{u}}}$};
  \path[arrow] (dmm_in1) edge[out=0,in=180] (dmm_cat) (dmm_in2) edge[out=0,in=180] (dmm_cat);
  \path[arrow] (dmm_cat) edge (dmm_tf) (dmm_tf) edge (dmm_proj) (dmm_proj) edge (dmm_out);

  \node[tensor] (fdm_in1) at (1.56,-12.0) {$\bm{z}_t$};
  \node[tensor] (fdm_in2) [below=1.6cm of fdm_in1] {$\bm{u}_t$};
  \node[layer, right=0.6cm of fdm_in1, fill=cyan!15] (a1) {Linear($2d_{\bm{z}}$)\\+ ReLU};
  \node[layer, right=0.4cm of a1, fill=cyan!15] (a2) {Linear($2d_{\bm{z}}$)\\+ ReLU};
  \node[layer, right=0.4cm of a2, fill=cyan!15] (a3) {Linear($2d_{\bm{z}}$)\\+ ReLU};
  \node[layer, right=0.4cm of a3, fill=cyan!15] (a4) {Linear($d_{\bm{z}}$)};
  \node[layer, right=0.6cm of fdm_in2, fill=cyan!15] (b1) {Linear($2d_{\bm{z}}$)\\+ ReLU};
  \node[layer, right=0.4cm of b1, fill=cyan!15] (b2) {Linear($2d_{\bm{z}}$)\\+ ReLU};
  \node[layer, right=0.4cm of b2, fill=cyan!15] (b3) {Linear($2d_{\bm{z}}$)\\+ ReLU};
  \node[layer, right=0.4cm of b3, fill=cyan!15] (b4) {Linear($d_{\bm{z}}$)};
  \node[circle, draw, thick, right=0.6cm of a4, yshift=-1.3cm] (plus) {$+$};
  \node[tensor, right=0.6cm of plus] (fdm_out) {$\bm{z}_{t+1}$};
  \path[arrow] (fdm_in1) edge (a1) (a1) edge (a2) (a2) edge (a3) (a3) edge (a4) (a4) edge[out=0,in=90] (plus);
  \path[arrow] (fdm_in2) edge (b1) (b1) edge (b2) (b2) edge (b3) (b3) edge (b4) (b4) edge[out=0,in=-90] (plus);
  \path[arrow] (plus) edge (fdm_out);

  \begin{pgfonlayer}{background}
    \node[draw=gray!80, dashed, thick, rounded corners, fill=gray!5,
          fit=(root_in)(root_out)(r2), inner sep=18pt] (bg_root) {};
    \node[draw=gray!80, dashed, thick, rounded corners, fill=gray!5,
          fit=(cnn_in)(cnn_out)(c1)(c2), inner sep=14pt] (bg_cnn) {};
    \node[draw=gray!80, dashed, thick, rounded corners, fill=gray!5,
          fit=(idm_in1)(idm_in2)(idm_out)(idm_l3), inner sep=14pt] (bg_idm) {};
    \node[draw=gray!80, dashed, thick, rounded corners, fill=gray!5,
          fit=(dmm_in1)(dmm_in2)(dmm_out)(dmm_tf), inner sep=14pt] (bg_dmm) {};
    \node[draw=gray!80, dashed, thick, rounded corners, fill=gray!5,
          fit=(fdm_in1)(fdm_in2)(fdm_out)(b1)(a2), inner sep=14pt] (bg_fdm) {};
  \end{pgfonlayer}
  \node[anchor=south west, font=\bfseries\normalsize] at ([yshift=0.05cm]bg_root.north west) {(a) Implicit Neural Representation};
  \node[anchor=south west, font=\bfseries\normalsize] at ([yshift=0.05cm]bg_cnn.north west)  {(b) CNN Encoder};
  \node[anchor=south west, font=\bfseries\normalsize] at ([yshift=0.05cm]bg_idm.north west)  {(c) Inverse Dynamics Model};
  \node[anchor=south west, font=\bfseries\normalsize] at ([yshift=0.05cm]bg_dmm.north west)  {(d) Generative Control Model};
  \node[anchor=south west, font=\bfseries\normalsize] at ([yshift=0.05cm]bg_fdm.north west)  {(e) Forward Dynamics Model (Top: $A$, Bottom: $B$)};
\end{tikzpicture}}
\caption{\textbf{Detailed \themethod components} as used on the WeatherBench dataset, corresponding to the experimental configuration ($d_{\bm{z}}=961$, $d_{\bm{u}}=16$, and $C=1$).}
\label{fig:warp_components}
\end{figure}

\Cref{fig:warp_components} provides a detailed view of key \themethod components at the configuration used in WeatherBench experiments (see  \cref{tab:model_parameters}):

\begin{enumerate}[(a)]
    \item \textbf{Implicit Neural Representation.} A 6-layer\footnote{Note that the output layer is included in the layer count; meaning that 6 layers corresponds to a depth of 5, following Equinox's convention \cite{kidger2021equinox}.} MLP of width 12 preceded by a Fourier coordinate encoding with 6 frequency bands. The weights and biases of all 5 layers are combined as $\bar{\bm{z}}+\bm{z}_t$ at render time. Non-batched inputs are $(\mathsf{x},\mathsf{y}) \in [-1,1]^2$; the output is the pixel value at the queried coordinate.
    \item \textbf{CNN Encoder.} A 5-layer strided convolutional backbone (stride 2, channels 64/128/256/512, ReLU activations) followed by a flatten and linear projection to $\mathbb{R}^{d_{\bm{z}}}$. We use $3\times 3$ kernels.
    \item \textbf{Inverse Dynamics Model.} A 4-layer MLP of width $d_{\bm{z}}$ taking concatenated $(\bm{z}_t, \bm{z}_{t+1})$ as input and outputting $\bm{u}_t \in \mathbb{R}^{d_{\bm{u}}}$.
    \item \textbf{Generative Control Model.} A causal Transformer with 4 blocks (8 attention heads, MLP ratio 4) that processes a sequence of concatenated $\{\bm{z}_\tau, \bm{u}_\tau\}_{1 \leq \tau \leq t}$ pairs, with learned positional embeddings. We set the values corresponding to $\bm u_t$ as $0$ in the input sequence.
    \item \textbf{Forward Dynamics Model.} Two independent 4-layer MLPs: $A$ of width $2d_{\bm{z}}$ taking $\bm{z}_t$ as input, and $B$ of width $2d_{\bm{z}}$ taking $\bm{u}_t$ as input. Their outputs are summed to produce $\bm{z}_{t+1}$.
\end{enumerate}

\paragraph{Architecture of the GCM.}
We employ Transformer architectures exclusively for the WeatherBench and MiniGrid datasets, whereas we utilise a GRU for Moving MNIST and an LSTM for PhyWorld. Because the latter two datasets serve as the basis for our intervention experiments, we prioritise attention-free recurrent architectures to prevent the model from inadvertently attending to historical, uncorrupted states. At each time step, the recurrent core (LSTM or GRU) receives the concatenated latent observation and previous action as input to update its internal hidden memory state. This updated memory state is then passed to an MLP decoder, which projects the concatenated hidden state and current latent observation into the target action space. Successfully accommodating this breadth of sequence-modelling backbones underscores the versatility of our framework, demonstrating that it is robust and agnostic to the choice of GCM architecture.

\paragraph{Structural inductive biases.}

It is important to note that while \themethod successfully disentangles background, object identity, and motion without any semantic supervision (i.e., no explicit labels nor ground-truth masks are provided), this emergence is facilitated by minimal structural bottlenecks rather than rigid inductive biases. Specifically, isolating $\bar{\bm{z}}$ from the temporal dynamics naturally encourages the absorption of static elements, while the additive structure of the FDM ($A(\bm{z}_t) + B(\bm{u}_t)$) provides a simple factorisation of state representations and temporal transitions. Furthermore, employing an Implicit Neural Representation supplies a basic coordinate-based spatial grounding without enforcing domain-specific priors.

\subsection{Context-Conditioned Video Generation}
\label{subsec_app:videogen}
To generate videos with up to $T_{\text{inf}}$ frames, a context ratio $\rho \in [0,1]$ controls the fraction of steps in which the reference ground-truth next frame is available. For $t/T \leq \rho$, the IDM provides the action; for $t/T > \rho$, the GCM takes over. Setting $\rho = 1$ gives oracle world modelling; $\rho = 0$ gives single-frame-conditioned video generation. \Cref{alg:nova_inference} details the full procedure.

\begin{algorithm}[h]
\caption{\themethod Context-Conditioned Video Generation}
\label{alg:nova_inference}
\begin{algorithmic}[1]
\Require Ref. video $\mathbf{o}_{1:T}$, inference steps $T_{\text{inf}}$, context ratio $\rho \in [0,1]$, coord. grid $(\mathsf{X},\mathsf{Y})$
\State $\bm{z}_1 \leftarrow \text{Enc}(\mathbf{o}_1)$
\State $\bm{m}_1 \leftarrow \text{GCM}_{\text{init}}()$ \Comment{Initialise memory buffer/state}
\For{$t = 1$ \textbf{to} $T_{\text{inf}}$}
    \State $ \bm{\theta}_t \leftarrow \texttt{unflat}(\bar{\bm{z}}+\bm{z}_t$)
    \State $\hat{\mathbf{o}}_t \leftarrow \text{MLP}_{\bm{\theta}_t}(\mathsf{X},\mathsf{Y})$ \Comment{Render current frame}
    \If{$t/T < \rho$ }
        \State $\bm{u}_t \leftarrow \text{IDM}(\bm{z}_t,\, \text{Enc}(\mathbf{o}_{t+1}))$ \Comment{Extract action from context}
    \Else
        \State $\bm{u}_t \leftarrow \text{GCM}_{\text{decode}}(\bm{m}_t,\, \bm{z}_{t})$ \Comment{Autoregressively predict action}
    \EndIf
    \State $\bm{m}_{t+1} \leftarrow \text{GCM}_{\text{encode}}(\bm{m}_t,\, \bm{z}_t,\, \bm{u}_t)$ \Comment{Update memory with chosen action}
    \State $\bm{z}_{t+1} \leftarrow A(\bm{z}_t) + B(\bm{u}_t)$ \Comment{Forward dynamics step}
\EndFor
\State \textbf{return} $\hat{\mathbf{o}}_{1:T_{\text{inf}}}$
\end{algorithmic}
\end{algorithm}

\paragraph{Managing memory during generation.} 
The internal memory state $\bm{m}_t$ presented in \cref{alg:nova_inference} takes different forms depending on the architecture: it is a fixed-size hidden vector in the recurrent architectures, but a sequence buffer (matrix) in the Transformer. We define three fundamental operations to abstract away internal GCM processing: \textit{init}, \textit{decode}, and \textit{encode}. 

For the \textbf{recurrent architectures} (GRU, LSTM), memory is initialised as a zero vector, $\bm{m}_1 = \mathbf{0} \in \mathbb{R}^{d_{\text{rnn}}}$. During \textit{encode}, the memory is recursively updated through the recurrent cell:
\begin{equation}
    \bm{m}_{t+1} = \text{RecurentCell}([\bm{z}_t, \bm{u}_t], \bm{m}_t).
\end{equation}
The \textit{decode} operation concatenates memory and latent state to predict the action via a MLP:
\begin{equation}
    \bm{u}_t = \text{MLP}_{\text{decode}}([\bm{m}_t, \bm{z}_t])
\end{equation}


For the \textbf{Transformer architecture}, memory is initialised as the zero matrix $\bm{m}_1 = \mathbf{0} \in \mathbb{R}^{T \times d_{\text{transformer}}}$. During \textit{encode}, the context token is computed using the current latent state and the newly predicted action. The memory is updated by permanently assigning this token to the $t$-th row of the buffer:
\begin{equation}
    \bm{m}_{t+1}[i, :] = 
    \begin{cases} 
        \bm{W}_{\text{in}}[\bm{z}_t, \bm{u}_t] + \bm{b}_{\text{in}} & \text{if } i = t \\
        \bm{m}_t[i, :] & \text{otherwise}.
    \end{cases}
\end{equation}
Conversely, during \textit{decode}, because $\bm{u}_t$ is not yet available, a query token is generated using a zero-padded action vector, $\bm{q}_t = \bm{W}_{\text{in}}[\bm{z}_t, \mathbf{0}] + \bm{b}_{\text{in}}$, and inserted into the $t$-th row of $\bm{m}_t$ to form a query buffer $\tilde{\bm{m}}_t$. This buffer is added to learned positional embeddings $\bm{E}_{\text{pos}}$ and processed through causally-masked Multi-Head Attention (MHA) and MLP blocks. The action is then linearly projected from the $t$-th output vector:
\begin{equation}
    \bm{u}_t = \bm{W}_{\text{out}} \, \text{Transformer}(\tilde{\bm{m}}_t + \bm{E}_{\text{pos}})_t + \bm{b}_{\text{out}}.
\end{equation}

In both architectures, the generation of the action $\bm{u}_t$ is causal. At any given timestep $t$, the GCM always sees the sequence of states up to the current step, $\bm{z}_{1:t}$, and the sequence of actions from the past, $\bm{u}_{1:t-1}$. This causal isolation justifies our simplified notations $\text{GCM}(\bm{z}_{t}, \bm{u}_{t-1})$ and $\text{GCM}(\bm{z}_{1:t}, \bm{u}_{1:t-1})$ utilised from \cref{fig:GCM_training} onwards.

\subsection{Zero-Shot Content and Motion Retargeting}

Because the context-conditioned video generation process is a causal system, we can perform zero-shot content and motion retargeting, either independently or simultaneously, through targeted inference-time interventions (see \Cref{alg:unified_retarget}). To perform \textbf{content retargeting} at specific time steps included in the intervention set $\mathcal{I}$,\footnote{One single intervention step (i.e., $|\mathcal{I}|=1$) can be enough for successful motion/content retargeting applicable to all subsequent steps, as we achieve on Moving MNIST.} we intervene just before the forward dynamics step by substituting the current state $\bm{z}_t$ with an external alien state $\bm{z}^{\text{alien}}_t$.\footnote{Note that all alien frame encodings $\bm{z}^{\text{alien}}_t$ could be identical, i.e., $\bm{z}^{\text{alien}}_t = \bm{z}^{\text{alien}}_{t+1}, \forall \, t = 1,\ldots,T_{\text{inf}}-1.$} Conversely, for \textbf{motion retargeting}, we intervene on the control variables by swapping the generated action $\bm{u}_t$ for an alien action $\bm{u}^{\text{alien}}_t$. Both interventions can be toggled as needed, operating cleanly within the existing generation pipeline without requiring any additional training or overhead.


\begin{algorithm}[h]
\caption{Content and/or Motion Retargeting}
\label{alg:unified_retarget}
\begin{algorithmic}[1]
\Require Reference video $\mathbf{o}_{1:T}$, inference steps $T_{\text{inf}}$, context ratio $\rho \in [0,1]$, coordinate grid $(\mathsf{X},\mathsf{Y})$, alien states $\bm{z}^{\text{alien}}$, alien actions $\bm{u}^{\text{alien}}$, intervention set $\mathcal{I}$
\State $\bm{z}_1 \leftarrow \text{Enc}(\mathbf{o}_1)$
\State $\bm{m}_1 \leftarrow \text{GCM}_{\text{init}}()$
\For{$t = 1$ \textbf{to} $T_{\text{inf}}$}
    \State $ \bm{\theta}_t \leftarrow \texttt{unflat}(\bar{\bm{z}}+\bm{z}_t$)
    \State $\hat{\mathbf{o}}_t \leftarrow \text{MLP}_{\bm{\theta}_t}(\mathsf{X},\mathsf{Y})$
    \If{$t/T < \rho$ }
        \State $\bm{u}_t \leftarrow \text{IDM}(\bm{z}_t,\, \text{Enc}(\mathbf{o}_{t+1}))$ 
    \Else
        \State $\bm{u}_t \leftarrow \text{GCM}_{\text{decode}}(\bm{m}_t,\, \bm{z}_{t})$
    \EndIf
    \State $\bm{m}_{t+1} \leftarrow \text{GCM}_{\text{encode}}(\bm{m}_t,\, \bm{z}_t,\, \bm{u}_t)$
    \If{$t \in \mathcal{I}$}
        \State \textcolor{red}{$\bm{z}_t \leftarrow \bm{z}^{\text{alien}}_t$} \Comment{Content intervention!}
        \State \textcolor{red}{$\bm{u}_t \leftarrow \bm{u}^{\text{alien}}_t$} \Comment{Motion intervention!}
    \EndIf
    \State $\bm{z}_{t+1} \leftarrow A(\bm{z}_t) + B(\bm{u}_t)$
\EndFor
\State \textbf{return} $\hat{\mathbf{o}}_{1:T_{\text{inf}}}$
\end{algorithmic}
\end{algorithm}

\section{Implementation and Experimental Details}

\subsection{Datasets Properties}
\label{subsec_app:datasets}





We evaluate our framework across diverse datasets, testing predictive capability, visual fidelity, and dynamic disentanglement across varied domains. Dataset properties are summarised in \cref{tab:datasets}.

\begin{enumerate}[(1)]
    \item \textbf{Moving MNIST} \cite{srivastava2015unsupervised} assesses deterministic spatial-temporal forecasting and boundary sharpness over time; sequences ($T=20$) show two digits bouncing on a $64 \times 64$ grayscale canvas. We use 8,000 training and 2,000 testing sequences.
    \item \textbf{PhyWorld 30K} \cite{kang2024far} evaluates modelling of complex, non-linear multi-body physical interactions. Rendered at $128 \times 128$ RGB resolution, we use exactly 26,066 training sequences and 1,635 testing sequences containing both in-distribution and out-of-distribution characteristics (ball radii and velocities).
    \item \textbf{WeatherBench} \cite{rasp2020weatherbench} tests continuous, global-scale real-world dynamics, popular as a benchmark for data-driven medium-range forecasting. The target temperature variable is standard-scaled using training statistics. We inherit the chronological split common in the literature: training (1979-2015), validation (2016), and testing (2017-2018).     
    \item \textbf{MiniGrid} \cite{MinigridMiniworld23} challenges the model to extract discrete control policies from visual data without explicit action labels. We generated offline goal-directed navigation sequences using a Breadth-First Search (BFS) optimal policy, rendered as $72 \times 72 \times 3$.
\end{enumerate}


\begin{table}[h]
    \centering
    \small
    \vspace{-1em}
    \caption{Dataset properties and evaluation splits. We abbreviate phase 1, phase 2, and phase 3 as p1, p2, and p3, respectively. NA is used when phase 1 was skipped, meaning world model-adjacent components (Enc, IDM, and FDM) were trained end-to-end. The star (*) indicated that a test set was not available; in which case the prescribed PhyWorld validation set was used for testing.}
    \label{tab:datasets}
    \begin{tabular}{@{}l c c c c c@{}}
        \toprule
        \textbf{Dataset} & $T$ & $H \times W \times C$ & \textbf{Train size} & \textbf{Test size} & \textbf{Batch sizes (p1, p2, p3)} \\
        \midrule
        Moving MNIST & 20 & $64 \times 64 \times 1$ & 8000           & 2000          & NA, 256, 256 \\
        PhyWorld 30K & 32 & $128 \times 128 \times 3$ & 26066 & 1635* & 16, 64, 64 \\
        WeatherBench & 24 & $32 \times 64 \times 1$ & 324313       &   17497         & 8, 128, 256 \\
        MiniGrid     & 10 & $72 \times 72 \times 3$ & 8000  & 2000 & 16, 24, 24 \\
        \bottomrule
    \end{tabular}
    \vspace{-1em}
\end{table}

\subsection{\themethod \& Baseline Implementations}

Hyperparameters were manually tuned based on empirical performance on the validation sets when available. Most batch sizes specified in \cref{tab:datasets}, along with a deep and narrow INR layers (which reduced the number of parameters $d_z = 961$ or $987$) meant the memory consumption was kept low across all datasets, never exceeding 4GB. This small memory usage was also made possible by the fact that we used the stop-gradient operator as described in \cref{fig:nova_method}, while instructing JAX to recompute all IDM- and FDM-related quantities during its backward pass.
We use a learning rate of $10^{-4}$ for MNIST, and $10^{-5}$ for the majority of other datasets and phases. Unless otherwise specified, we generate videos and do interventions using 10 context frames for Moving MNIST and 3 for PhyWorld, as suggested in their respective papers. In the discrete NOVA implementation described in \cref{subsec:discrete_navigation} for MiniGrid, we used a codebook of 200 vectors, each of dimension 2; the same codebook is reused by the GCM when performing behaviour cloning. Further \themethod training details are provided in \cref{tab:model_parameters}. 

We implement the Standard WM in similar conditions to \themethod, even matching its batch sizes. As for the encoder-decoder-free LAPO \cite{schmidt2023learning}, we adapt the code from its official repository, readily available and downloaded from \url{https://github.com/schmidtdominik/LAPO}.

\begin{table}[h]
    \centering
    \small
    \caption{Parameter count across datasets. ``Vector'' describes the dimension of one-dimension latent vectors, while ``Submodel'' describes submodel components within \themethod. ``Total Params'' counts all learnable submodel parameters, plus the length of the isolated $\bar{\bm z}$. To prevent confusion, we present configurations for the MiniGrid implementation with continuous actions. Across all datasets, total learnable parameters amount to roughly 40 million.}
    \label{tab:model_parameters}
    \setlength{\tabcolsep}{6pt}
    \renewcommand{\arraystretch}{1.15}
    \begin{tabular}{@{}l | cc | cccc | c@{}}
        \toprule
        \multicolumn{1}{@{}l}{\multirow{2}{*}{\textbf{Dataset}}} & \multicolumn{2}{c}{\textbf{Vector}} & \multicolumn{4}{c}{\textbf{Submodel}} & \multicolumn{1}{c@{}}{\multirow{2}{*}{\textbf{Total Params}}} \\
        \cmidrule(lr){2-3} \cmidrule(lr){4-7}
        \multicolumn{1}{@{}l}{} & \multicolumn{1}{c}{$d_{\bm u}$} & \multicolumn{1}{c}{$d_{\bm z}$} & \multicolumn{1}{c}{$d_{\text{Enc.}}$} & \multicolumn{1}{c}{$d_{\text{FDM}}$} & \multicolumn{1}{c}{$d_{\text{IDM}}$} & \multicolumn{1}{c}{$d_{\text{GCM}}$} & \multicolumn{1}{c@{}}{} \\
        \midrule
        Moving MNIST & 4 & 961 & 9423297  & 20338604 & 2776333 & 1251588        & 33790783 \\
        PhyWorld 30K & 4 & 987 & 16561398 & 21453432 & 3903589 & 1597444 & 43516850 \\
        WeatherBench &  16 &  961   &  14070658        &   20361668       &  3712359       &   3421712      &   41567358       \\
        MiniGrid     &  4 &   987  &    6707190      &    21453432      &   3903589      &    1597444     &    33662642      \\
        \bottomrule
    \end{tabular}
\end{table}

\begin{table}[h]
    \centering
    \small
    \caption{\themethod wallclock training times across datasets. NA indicates that phases 1 and 2 were performed jointly.}
    \label{tab:training_times}
    \setlength{\tabcolsep}{6pt}
    \renewcommand{\arraystretch}{1.15}
    \begin{tabular}{@{}l ccc@{}}
        \toprule
        \multirow{2}{*}{\textbf{Dataset}} & \multicolumn{3}{c}{\textbf{Train. Times (hours)}} \\
        \cmidrule(l){2-4}
        & \textbf{Phase 1} & \textbf{Phase 2} & \textbf{Phase 3} \\
        \midrule
        Moving MNIST & NA   & 7.0 & 1.2 \\
        PhyWorld 30K & 12.8 & 8.1 &  5.5   \\
        WeatherBench & 0.9   &  2.4   &  1.0   \\
        MiniGrid     &  3.3    &  16.4   &  1.4   \\
        \bottomrule
    \end{tabular}
\end{table}

\subsection{Evaluation Metrics}

\paragraph{Conventional metrics.}
We use standard pixel-level metrics such as Mean Squared Error (MSE) 
for direct comparison during training, validation, and testing. Mean Absolute Error (MAE) is used during the latter two stages.

To complement this on the \textbf{PhyWorld} physics datasets, we extract 
the centre-of-mass position $x_t^{(i)}$ of each ball $i \in \{1,2\}$ 
from each predicted frame and estimate the velocity via finite differences 
$v_t^{(i)} = (x_t^{(i)} - x_{t-1}^{(i)})/\Delta t$, with $\Delta t = 0.1$. 
Ball masses are set to $m_i = r_i^2$, proportional to the squared radius 
provided by the dataset's initial conditions. We then evaluate total linear 
momentum $\text{Mom}_t = m_1 v_t^{(1)} + m_2 v_t^{(2)}$ and total kinetic energy 
$\text{KE}_t = \tfrac{1}{2}m_1 (v_t^{(1)})^2 + \tfrac{1}{2}m_2(v_t^{(2)})^2$, 
averaging across time and reporting the mean absolute error against the same quantities computed from ground truth positions 
issued by \citet{kang2024far}.

Ball positions are extracted from predicted frames via colour thresholding. 
Each frame $\hat{\bm{o}}_t \in [0,1]^{H \times W \times 3}$ is segmented 
into two binary masks: a red-ball mask, selecting pixels where the R channel 
exceeds $\tau = 0.8$ while G and B remain below $\tau$, and a blue-ball mask, 
selecting pixels where B exceeds $\tau$ while R and G remain below $\tau$. 
The centre-of-mass position of each ball is then computed as the mean pixel 
coordinate over the corresponding mask and normalised to $[0,1]$ by dividing 
by the frame dimensions $W$ and $H$ respectively. If a mask is empty at 
time $t$, indicating the ball has left the field of view or is occluded,  the position is set to the last known position at $t-1$. 


\paragraph{Distributional and structural metrics.}
Beyond trajectory-level agreement, we further 
assess structural and distributional consistency at the frame level 
through the metrics below.

Let $\hat{\bm{o}}_t \in [0,1]^{H \times W\times C}$ denote the predicted frame at
time step $t$, and let $\hat{p}_t$ be its empirical pixel-intensity distribution,
estimated via a histogram over $[0,1]$.
We aim to quantify \emph{identity drift}, stemming from the fact that a faithful
world model should preserve the visual identity of each object even as it
undergoes rigid motion; pixel-distribution metrics are invariant to
such motion by construction.
To quantify structural and distributional consistency over time, all metrics compare
$\hat{\bm{o}}_t$ (or its histogram $\hat{p}_t$) against a ground truth $\bm{o}_t$ (or $\hat{p}$).

We remark that for long horizon forecasting tasks performed on Moving MNIST where ground truth beyond $T=20$ is not available, we use, in place of $p_t$, the aggregate empirical distribution $p_{\mathrm{emp}}$ pooled over all
available GT frames, i.e., $p_t = p_{\mathrm{emp}}, \forall \, t>1$. For pixel-based metrics such as SSIM and PSNR, we solely rely on the first frame
$\bm{o}_1$ because spatially averaging GT frames would introduce motion-blur artefacts that corrupt the structural reference.

All histogram-based metrics use $N{=}64$ equal-width bins;
Wasserstein-1, JSD distance, and Bhattacharyya are computed via
\texttt{scipy.stats.wasserstein\_distance} and
\texttt{scipy.spatial.distance.jensenshannon}, with bin centres as support
points for $W_1$;
SSIM and PSNR use \texttt{skimage.metrics} against $\bm{o}_t$;
the FFT metric is computed via \texttt{numpy.fft.fft2} with the reference
spectrum averaged over all GT frames for motion robustness.

\begin{itemize}

  \item \textbf{\emph{Wasserstein-1 distance}} ($W_1$\,, $\downarrow$).
  \[
    W_1(p_t,\hat{p}_t)
      = \inf_{\gamma\in\Gamma(p_t,\,\hat{p}_t)}
          \mathbb{E}_{(u,v)\sim\gamma}\!\bigl[|u-v|\bigr]
      = \int_0^1 \bigl|F_{t}(x)-\hat F_t(x)\bigr|\,\mathrm{d}x,
  \]
  where $F_{t}$ and $\hat F_t$ are the CDFs of $p_t$ and
  $\hat{p}_t$, respectively.
  The closed-form CDF reduction holds in one dimension \cite{vallender1974calculation}.

  \item \textbf{\emph{Jensen–Shannon Divergence}} ($\mathrm{JSD}$\,, $\downarrow$).
    \[
      \mathrm{JSD}(p_t, \hat{p}_t)
        = \sqrt{\tfrac{1}{2}D_{\mathrm{KL}}(p_t\|m)+\tfrac{1}{2}D_{\mathrm{KL}}(\hat{p}_t\|m)},
    \]
    where $m=\tfrac{1}{2}(p_t+\hat{p}_t)$ and
    $D_{\mathrm{KL}}(p\|q)=\sum_b p_b\ln(p_b/q_b)$.

\item \textbf{\emph{Bhattacharyya distance}} ($B$\,, $\downarrow$).
  \[
    B(p_t,\hat{p}_t)
      = -\ln\!\sum_{b=1}^{N}\sqrt{p_{t,b}\,\hat{p}_{t,b}},
  \]
  where the sum inside the logarithm is the Bhattacharyya coefficient, and $N$ the number of bins; 
  $B=0$ iff $p_t=\hat{p}_t$, and $B\to\infty$ as the distributions become disjoint.


    \item \textbf{\emph{Structural Similarity Index}} (SSIM, $\uparrow$).
    \[
      \mathrm{SSIM}(\bm{o}_t,\hat{\bm{o}}_t)
        = \frac{
            \bigl(2\mu_{o}\,\mu_{\hat{o}}+C_1\bigr)
            \bigl(2\sigma_{o\hat{o}}+C_2\bigr)
          }{
            \bigl(\mu_{o}^{2}+\mu_{\hat{o}}^{2}+C_1\bigr)
            \bigl(\sigma_{o}^{2}+\sigma_{\hat{o}}^{2}+C_2\bigr)
          },
    \]
    where all statistics are computed within a sliding $11{\times}11$
    Gaussian window ($\sigma_w=1.5$): $\mu_{o}$ and $\mu_{\hat{o}}$ denote
    the local means, $\sigma_{o}^{2}$ and $\sigma_{\hat{o}}^{2}$ the local
    variances, and $\sigma_{o\hat{o}}$ the local covariance of $\bm{o}_t$
    and $\hat{\bm{o}}_t$; the reported score is the mean of this map over
    all window positions.
    The stabilisation constants are $C_1=(K_1 L)^2$ and $C_2=(K_2 L)^2$,
    with $K_1=0.01$, $K_2=0.03$, and $L$ the pixel dynamic range
    (e.g.\ $L=1$ for images normalised to $[0,1]$).
    $\mathrm{SSIM}\in[-1,1]$, with $1$ indicating perfect structural
    agreement.

  \item \textbf{\emph{Peak Signal-to-Noise Ratio}} (PSNR, $\uparrow$).
  \[
    \mathrm{PSNR}(\bm{o}_t,\hat{\bm{o}}_t)
      = 10\log_{10}\!\left(\frac{1}
          {\mathrm{MSE}(\bm{o}_t,\hat{\bm{o}}_t)}\right)
        \;\mathrm{[dB]},
    \qquad
    \mathrm{MSE}(\bm{a},\bm{b})
      = \frac{1}{HWC}\|\bm{a}-\bm{b}\|_F^2,
  \]
  where the peak value $\mathrm{MAX}=1$ for the normalised range $[0,1]$
  has been substituted.
  Like SSIM, PSNR is not motion-invariant and is included as a complementary
  pixel-fidelity reference.

  \item \textbf{\emph{FFT magnitude distance}} ($d_{\mathcal{F}}$\,, $\downarrow$).
  \[
    d_{\mathcal{F}}(\mathsf{T},\hat{\bm{o}}_t)
      = \frac{1}{HW}\left\|
          \frac{1}{|\mathsf{T}|}\sum_{t'\in\mathsf{T}}
            \bigl|\mathcal{F}(\bm{o}_{t'})\bigr|
          - \bigl|\mathcal{F}(\hat{\bm{o}}_t)\bigr|
        \right\|_F,
  \]
  where $\mathcal{F}:\mathbb{R}^{H\times W}\!\to\mathbb{C}^{H\times W}$
  denotes the 2-D DFT, $|\cdot|$ the element-wise complex modulus,
  $\mathsf{T}$ the set of all GT frame indices, and $\|\cdot\|_F/(HW)$
  the size-normalised Frobenius norm.
  While global rigid translation perfectly preserves the magnitude spectrum, scenes with multiple independent entities may exhibit phase interference. Averaging over $\mathsf{T}$ mitigates these oscillating cross-terms, yielding a robust spectral reference of the underlying object structures. This metric is only computed for $C=1$.

\end{itemize}

\section{Additional Results}
\label{sec_app:results}

\subsection{Latent Structural Disentanglement}



\begin{figure}[h]
\centering
\includegraphics[width=.5\linewidth]{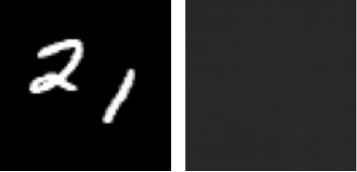}
\caption{Background isolation on Moving MNIST. Left: one frame from the dataset, rendered with $\bar{\bm z}+\bm z_t$; Right: a rendering of $\bar{\bm z}$ after training, indicating that it encodes the shared dark background}
\label{fig:background_zbar}
\end{figure}

\paragraph{Background absorption by $\bar{\bm{z}}$.}
In all Moving MNIST sequences, the background is uniformly dark. As seen in  \cref{fig:background_zbar}, we observe that $\bar{\bm{z}}$ naturally absorbs this, indicating that the per-frame offsets $\bm{z}_t$ have near-zero contribution to background pixels. 

The utility of $\bar{\bm z}$ extends beyond structural disentanglement. In particular, sequences with dynamic backgrounds or moving cameras, such as in-the-wild videos \cite{garrido2026learning}, could mean that a single dataset-wide $\bar{\bm z}$ is not able to perfectly isolate all non-dynamic components. In such settings, we hypothesise that $\bar{\bm z}$ still acts as an important weight-space anchor, as demonstrated by our ablation in \cref{fig:bar_z_ablation}. By framing the temporal predictions as small residuals $\bm{z}_t$ relative to $\bar{\bm z}$, analogous to residual learning \cite{he2016deep}, we stabilise optimisation and prevent the network from collapsing into local minima. Consequently, even in complex scenes where background disentanglement is challenging, the residual anchoring provided by $\bar{\bm z}$ may remain beneficial for efficient learning.

\begin{figure}[h]
\centering
\includegraphics[width=\linewidth]{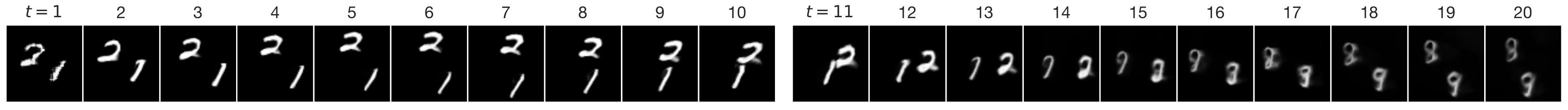}
\includegraphics[width=\linewidth]{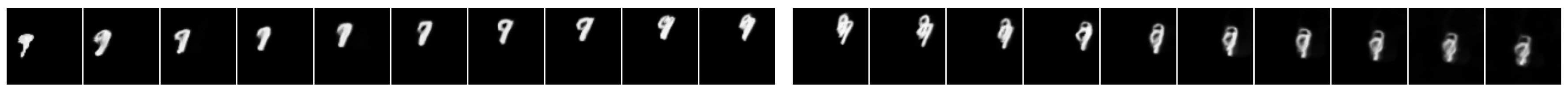}
\includegraphics[width=\linewidth]{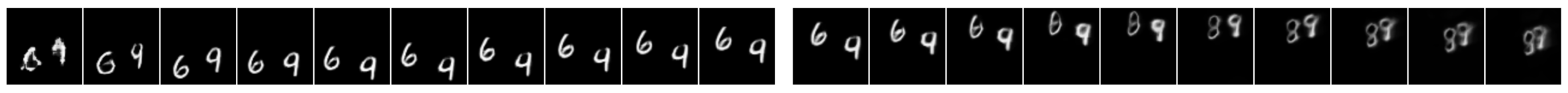}
\includegraphics[width=\linewidth]{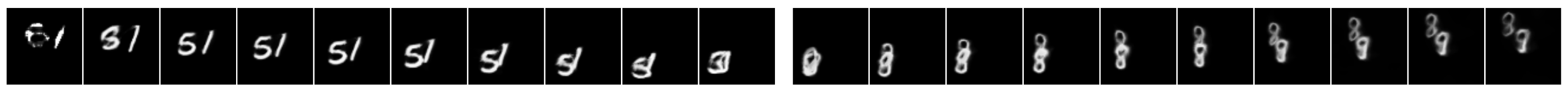}
\includegraphics[width=\linewidth]{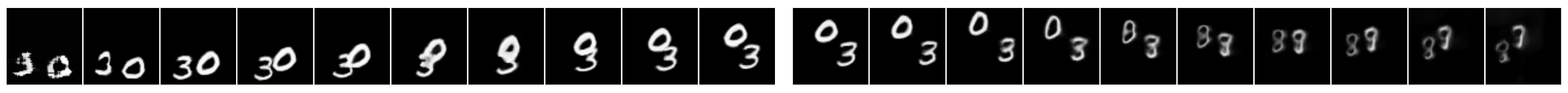}
\caption{\textbf{\themethod's content identity is encoded in $\bm{z}_t$.} After $t=10$ (context phase), we interpolate $\bm{z}_t \to \mathbf{0}$ while maintaining GCM-extracted actions $\bm{u}_t$. Digit identities gradually morph towards the base-network prototype (8s and 9s), while motion trajectories are preserved. No supervision on digit identity was provided.}
\label{fig:morphing_to_89}
\end{figure}

\paragraph{The $A$ component processes content identity.}
\Cref{fig:morphing_to_89} illustrates a controlled intervention on Moving MNIST. After a context phase ($t \leq 10$), we replace $\bm{z}_t$ with a convex combination towards $\mathbf{0}$ (the zero offset vector), while keeping $\bm{u}_t$ free from the IDM. The zero vector corresponds to the base network $\bar{\bm{z}}$ alone, which renders a ``prototype'' or dataset-average scene. Feeding this interpolated state through $A$ gradually morphs the digit identity towards lightly-shaded 8s and 9s, while preserving the motion trajectory computed by the GCM. Since perturbing $\bm{z}_t$ changes \emph{what} appears, not \emph{where} it moves, this demonstrates that $A(\bm{z}_t)$ processes the semantic identity of the objects.

\begin{figure}[h]
\centering
\includegraphics[width=\linewidth]{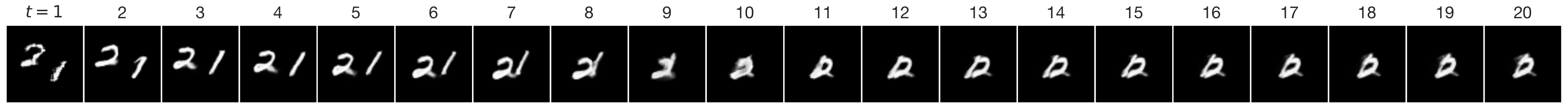}
\includegraphics[width=\linewidth]{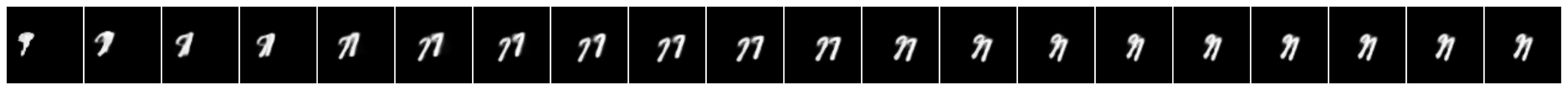}
\includegraphics[width=\linewidth]{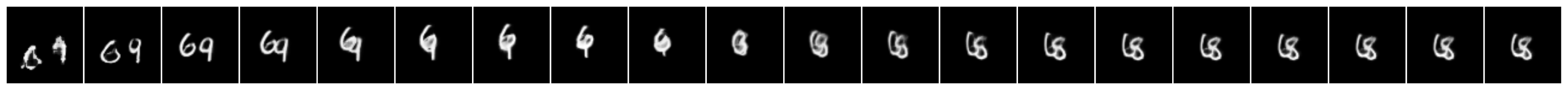}
\includegraphics[width=\linewidth]{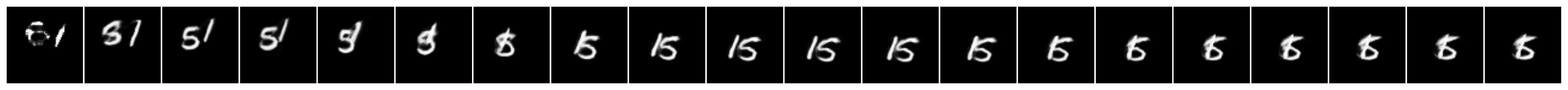}
\includegraphics[width=\linewidth]{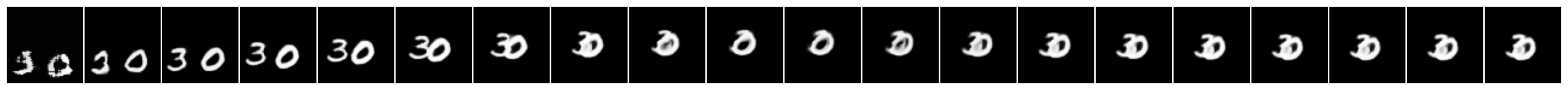}
\caption{\textbf{\themethod's spatial motion is encoded in $\bm{u}_t$.} From $t=0$, we interpolate $\bm{u}_t \to \mathbf{0}$ using GCM-generated actions, while keeping $\bm{z}_t$ free. Digits decelerate and converge towards the canvas centre, while their visual identities are fully preserved throughout.}
\label{fig:moving_to_00}
\end{figure}

\paragraph{The $B$ component processes spatial motion.}
\Cref{fig:moving_to_00} shows the complementary intervention: keeping $\bm{z}_t$ free while interpolating $\bm{u}_t$ towards $\mathbf{0}$. The digits progressively decelerate and converge towards the centre of the canvas. Their visual identity is fully preserved throughout. This confirms that $B(\bm{u}_t)$ processes motion instructions but carries no information about object identity.

\subsection{Context-Conditioned Video Generation}
\label{subsec_app:generation}

\subsubsection{Moving MNIST}

\begin{figure}[h]
\centering
\includegraphics[width=\linewidth]{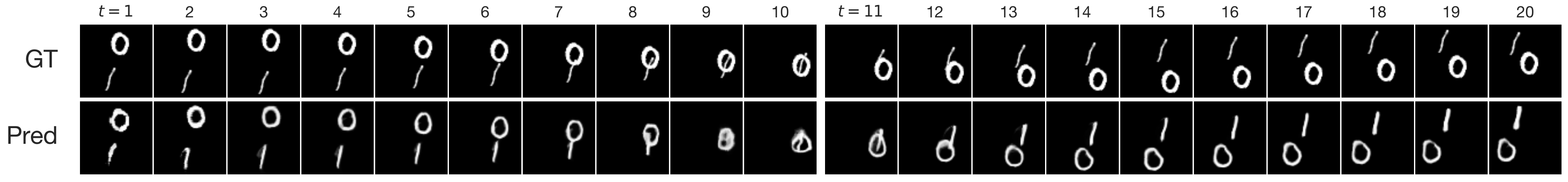}
\includegraphics[width=\linewidth]{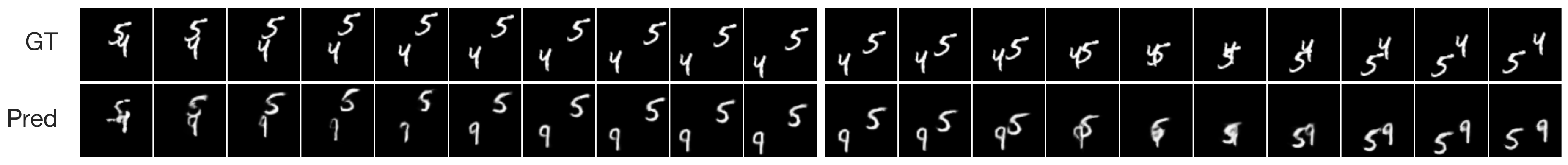}
\includegraphics[width=\linewidth]{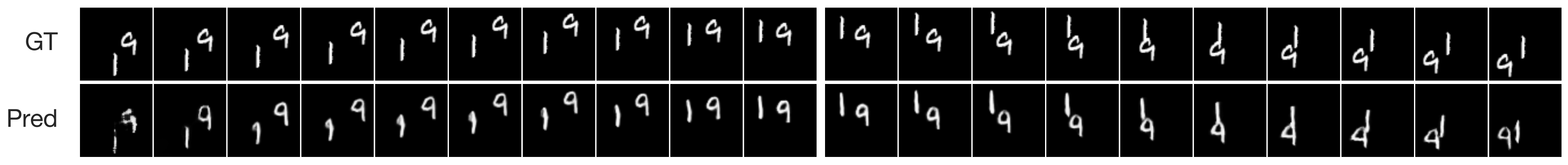}
\includegraphics[width=\linewidth]{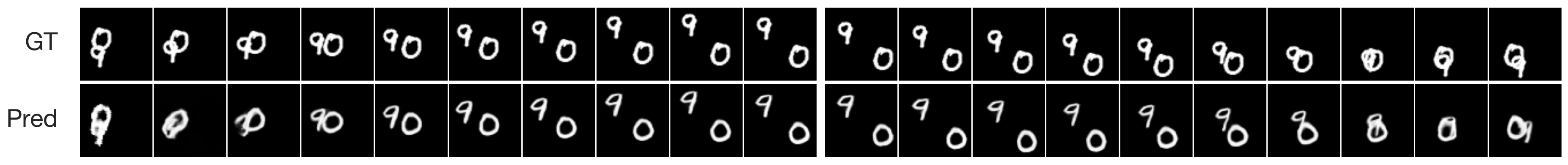}
\includegraphics[width=\linewidth]{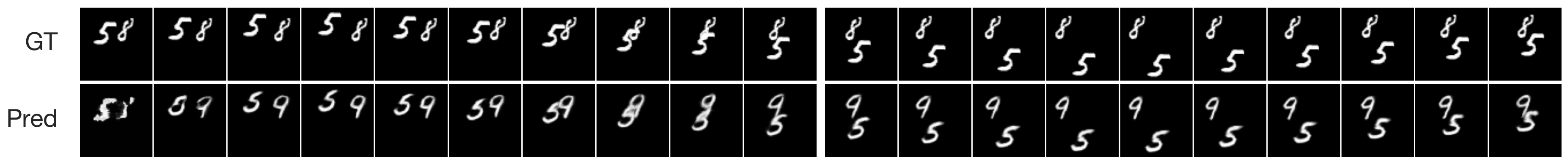}
\includegraphics[width=\linewidth]{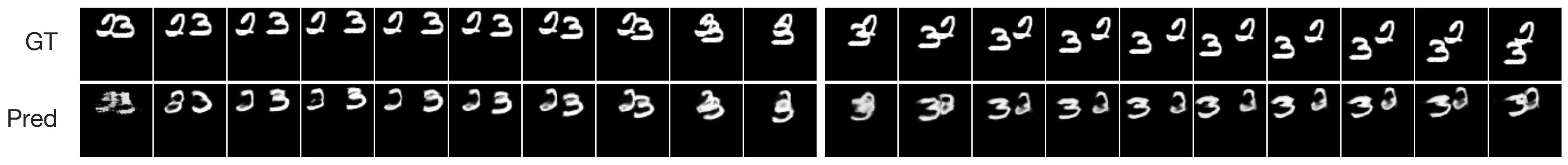}
\caption{\textbf{Context-conditioned video generation on Moving MNIST.} Frames 1--10 are context (ground truth); frames 11--20 are \themethod predictions. The model maintains digit identity and trajectory without artefacts.}
\label{fig:video_generation}
\end{figure}

\paragraph{Forecasting within the training horizon $T_{\text{inf}}=20$.} 
To complement results from  \cref{subsec:movingmnistgen}, we generate a number of frames using \cref{alg:nova_inference}. \Cref{fig:video_generation} shows qualitative video forecasting results on Moving MNIST, conditioned on 10 initial frames and rolling out for 10 predicted frames. \themethod produces sharp predictions without checkerboard artefacts, a common failure mode in transposed-CNN decoders. 

\paragraph{Forecasting over longer horizons $T_{\text{inf}}=1000$} Long-horizon generation is a potent test of identity consistency: a world model that merely interpolates over the temporal dimension will eventually collapse, while one with a structured representation should remain coherent far beyond its training horizon.
We use \cref{alg:nova_inference} to generate extremely long sequences. Given their storage and computational cost, we limit evaluation to two test sequences; we acknowledge this may not be fully representative of the test distribution. 

\begin{figure}
\centering
  \includegraphics[width=0.8\linewidth]{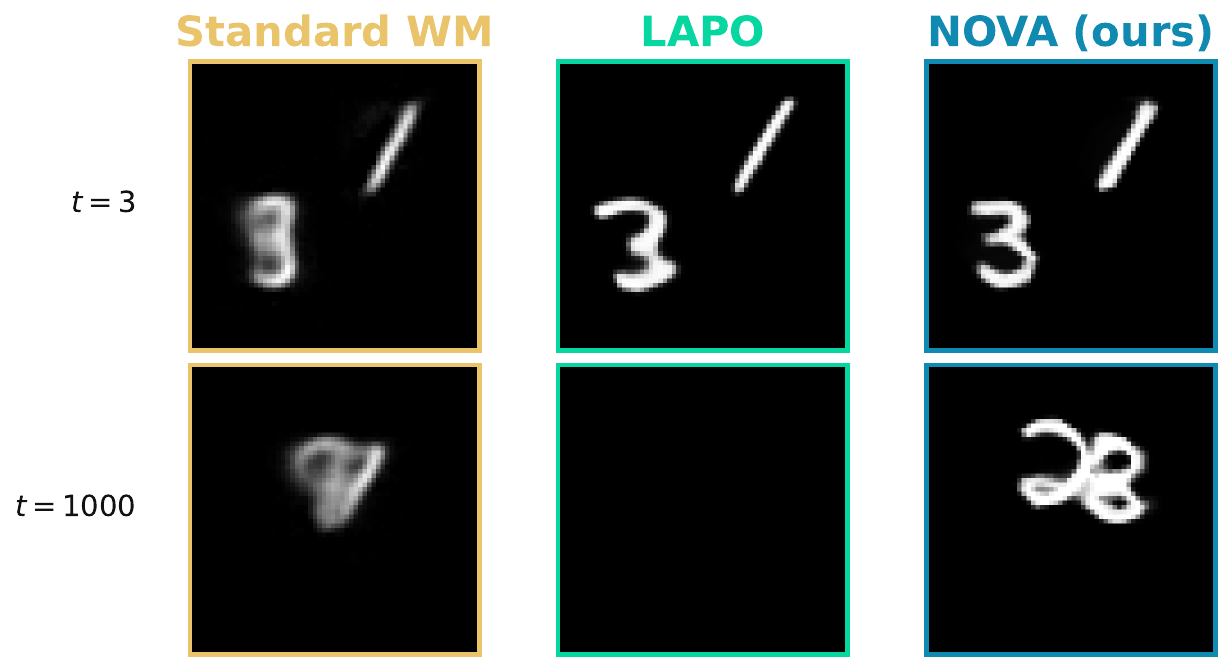}
  \caption{Comparison of long-horizon forecast ability. The models are only shown the initial two frames, and most autoregressively predict from frame $t=3$ to $t=1000$.}
  \label{fig:long_horizon}
\end{figure}

\begin{figure}[h]
\centering
\includegraphics[width=0.6\linewidth]{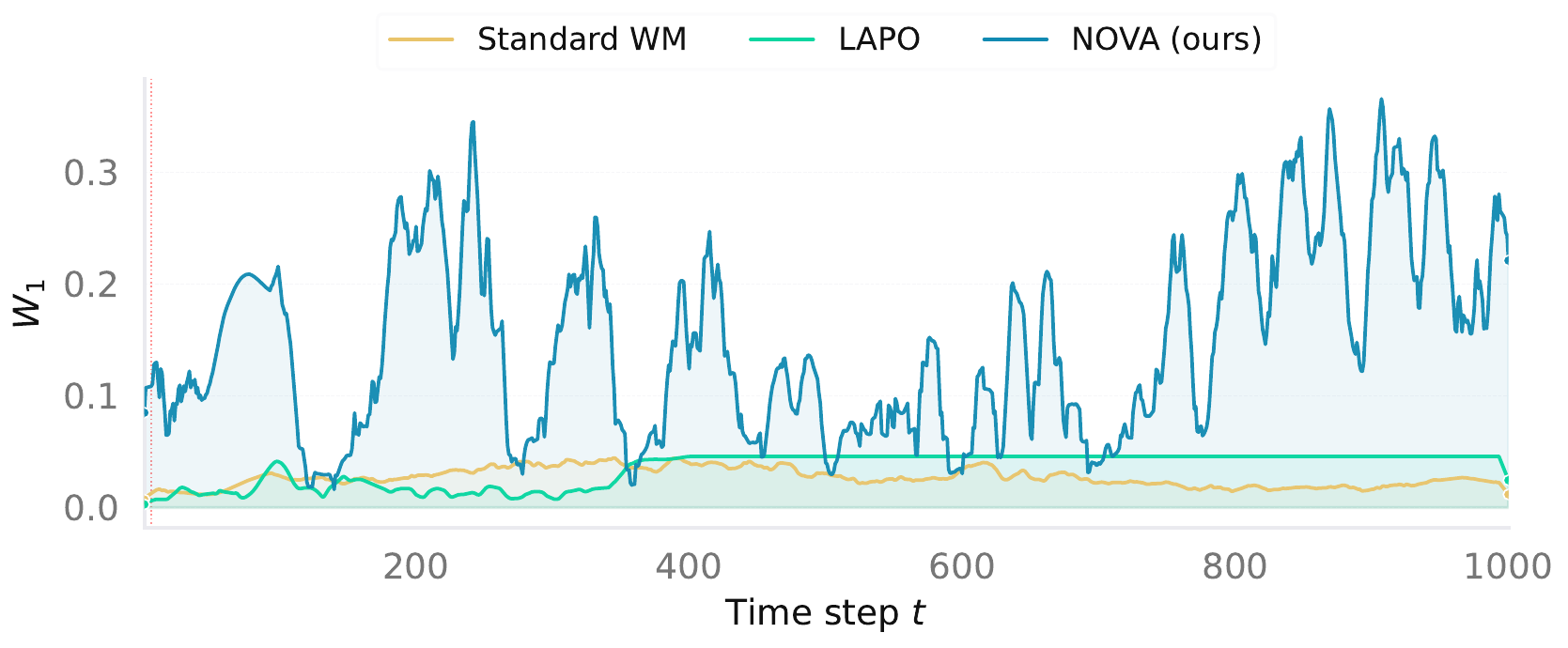}
\includegraphics[width=0.6\linewidth]{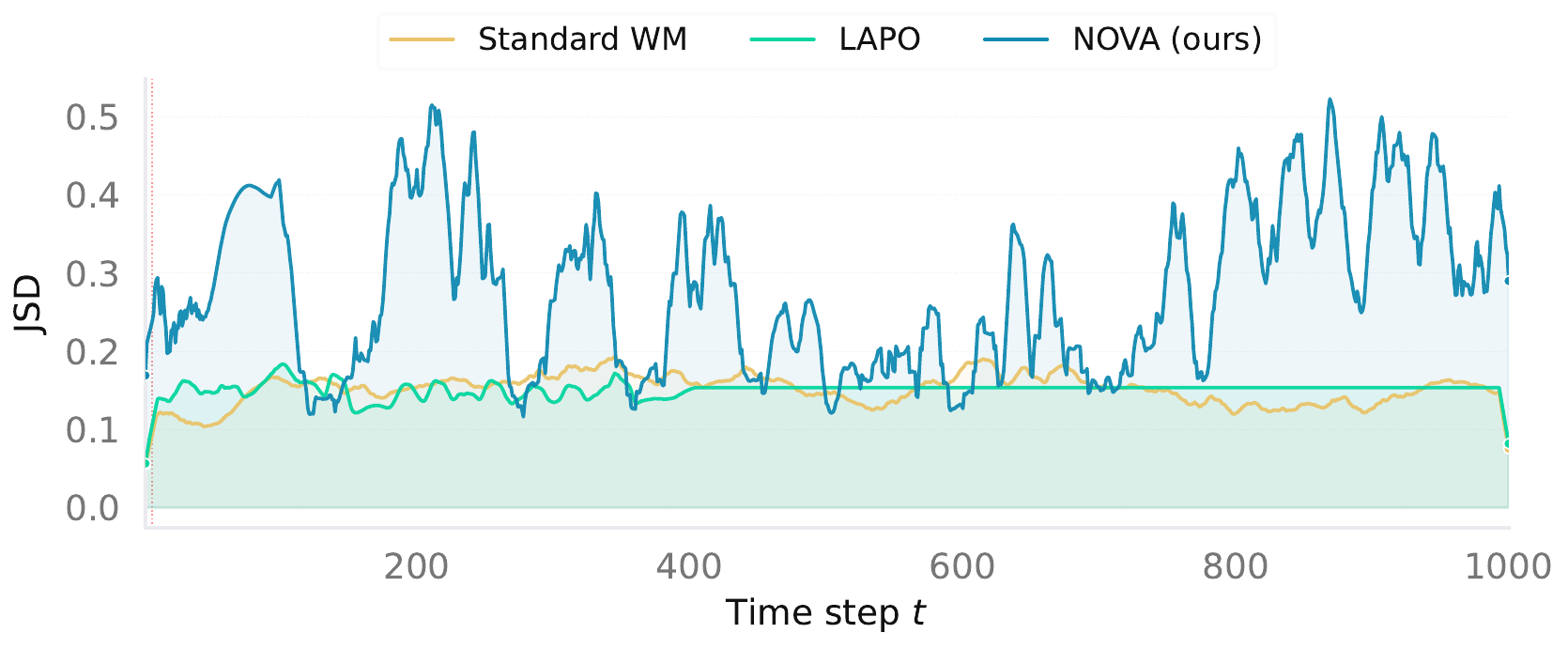}
\includegraphics[width=0.6\linewidth]{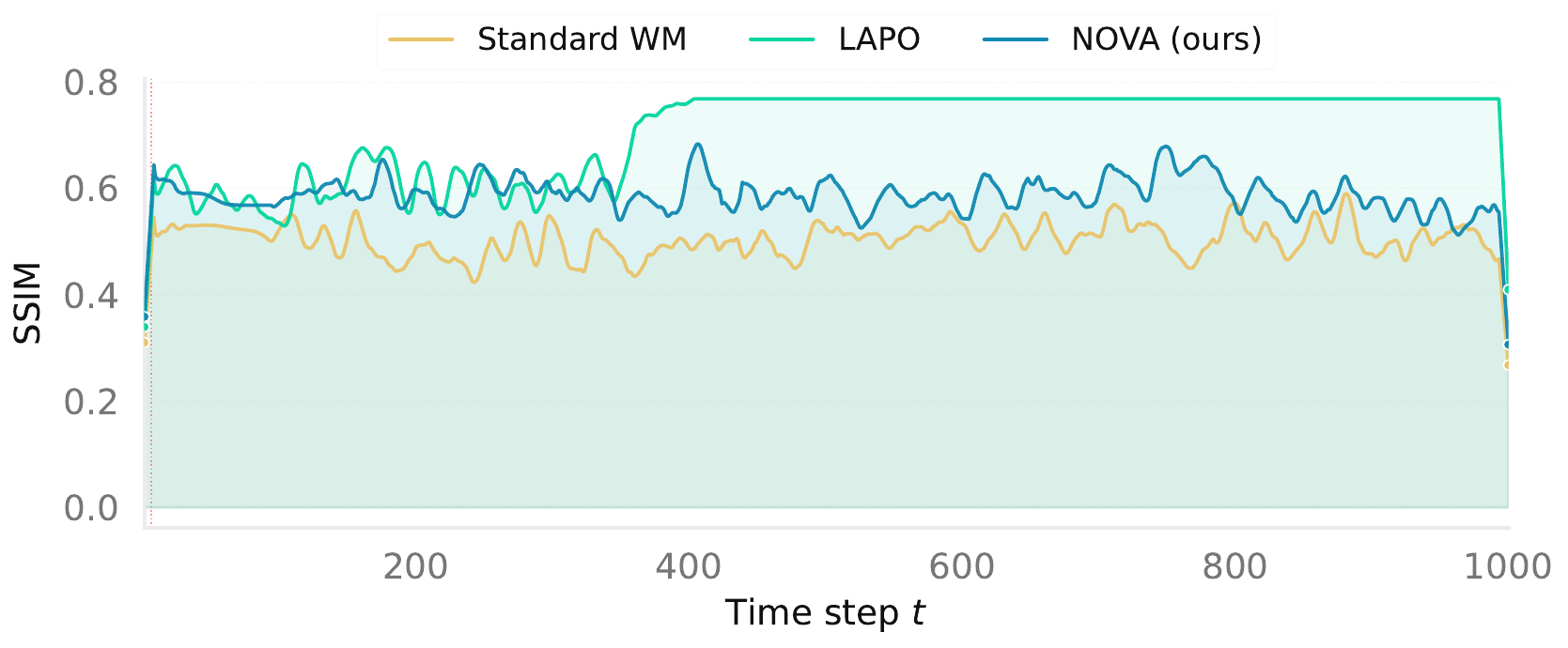}
\includegraphics[width=0.6\linewidth]{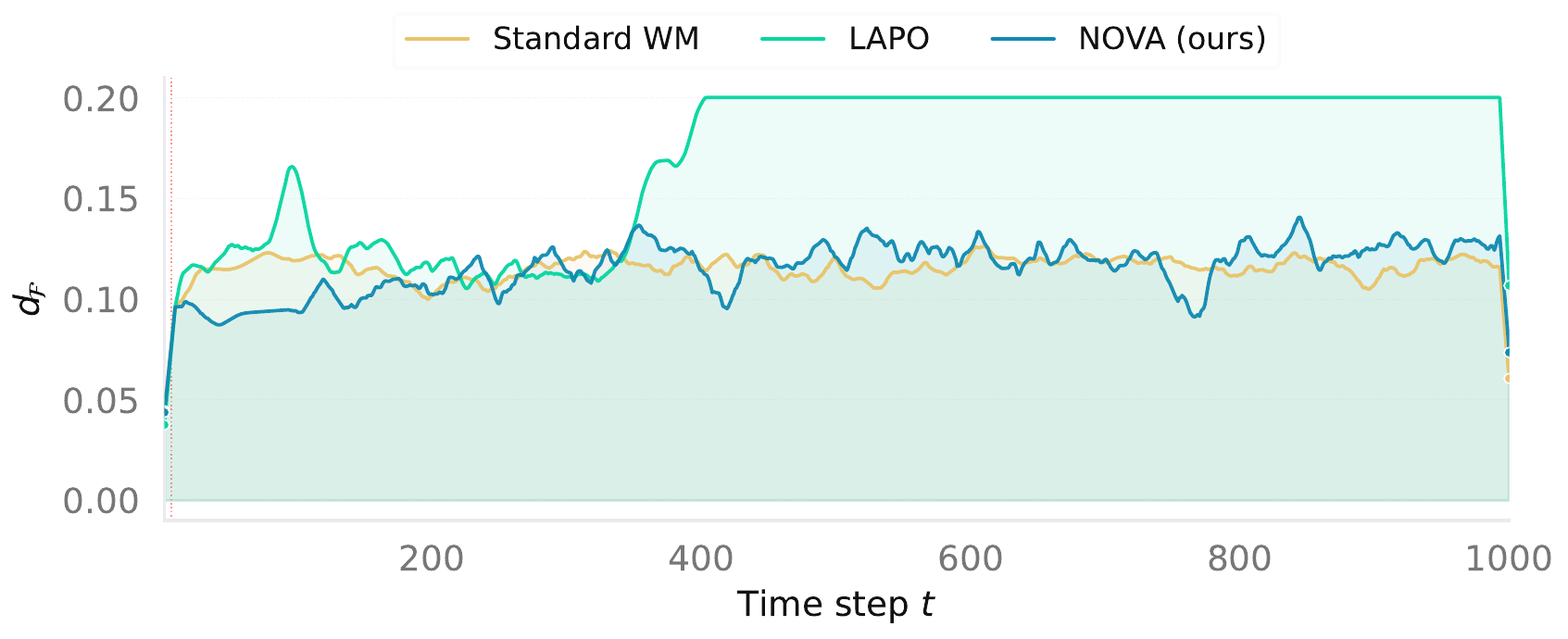}
\includegraphics[width=0.6\linewidth]{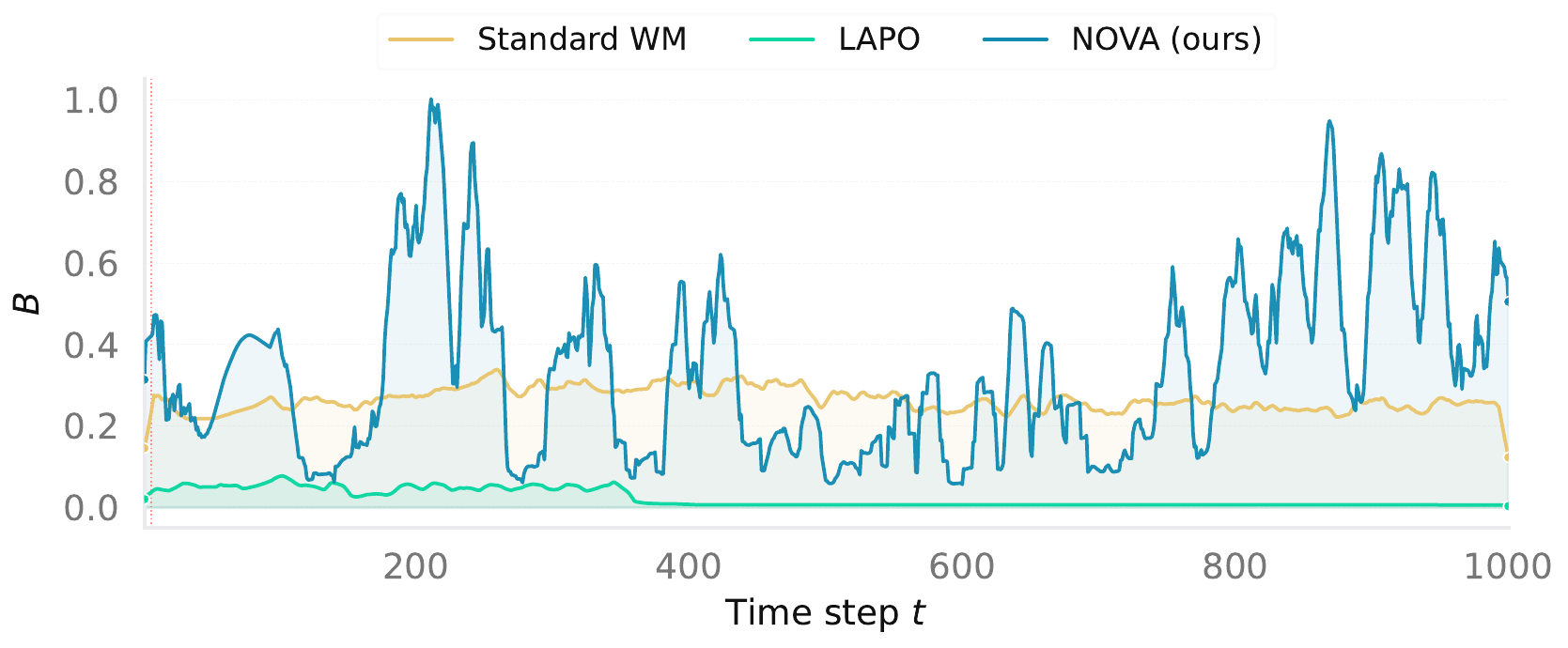}
\caption{\textbf{Long-horizon generation.} Comparison of long forecast ability using several metrics. The models are only shown the initial two frames, and most predict from frame $t=3$ to $t=1000$. This complements \cref{fig:long_horizon} by adding time-dependent quantitative values.}
\label{fig:long_horizon_app}
\end{figure}

Results in \cref{fig:long_horizon} show that LAPO collapses, as its pixel distribution fully darkens and degenerates over time. \cref{fig:long_horizon_app,tab:lh_metrics} further indicate that the standard WM is stable but for the wrong reason: it hedges toward a static, blurry distribution of pixels, hardly distinguishable from start to finish.
\themethod, by contrast, commits. It picks digits and generates them, which means it can be wrong at times (visible as spikes in \cref{fig:long_horizon_app}), but the median metrics in \cref{tab:lh_metrics} indicate that \themethod is more often right than wrong compared to the Standard WM.





\begin{table}[h]
\centering
\caption{%
  Long-horizon identity-consistency metrics. We report the median,  min, and max over
  \(T=1000\) frames and over two sequences \(\{54, 57\}\), limited as such for computational reasons. 
}
\label{tab:lh_metrics}
\setlength{\tabcolsep}{6pt}
\renewcommand{\arraystretch}{1.15}
\begin{tabular}{l l ccc}
\toprule
\textbf{Metric} & \textbf{Method}
  & \textbf{Median} & \textbf{Min} & \textbf{Max} \\
\midrule
$W_1$ \(\downarrow\) & Standard WM & 0.0780 & 0.0462 & 0.1308 \\
 & LAPO & 0.0110 & 0.0037 & 0.0847 \\
 & NOVA & 0.1222 & 0.0258 & 0.6138 \\
\midrule
$\mathrm{JSD}$ \(\downarrow\) & Standard WM & 0.4255 & 0.3313 & 0.4966 \\
 & LAPO & 0.0771 & 0.0703 & 0.2731 \\
 & NOVA & 0.3204 & 0.1418 & 0.7994 \\
\midrule
$\mathrm{SSIM}$ \(\uparrow\) & Standard WM & 0.5873 & 0.4337 & 0.8828 \\
 & LAPO & 0.7683 & 0.4664 & 1.0000 \\
 & NOVA & 0.5983 & 0.3552 & 0.9142 \\
\midrule
$d_{\mathcal{F}}$ \(\downarrow\) & Standard WM & 0.1224 & 0.0903 & 0.1982 \\
 & LAPO & 0.0409 & 0.0234 & 0.2343 \\
 & NOVA & 0.1953 & 0.1313 & 0.2517 \\
\midrule
$B$ \(\downarrow\) & Standard WM & 0.2698 & 0.1542 & 0.3785 \\
 & LAPO & 0.0086 & 0.0071 & 0.1025 \\
 & NOVA & 0.1442 & 0.0270 & 1.8013 \\
\bottomrule
\end{tabular}
\end{table}


\subsubsection{PhyWorld}
\label{subsec_app:phyworld_gen}
\begin{figure}[h]
\centering
\includegraphics[width=\linewidth]{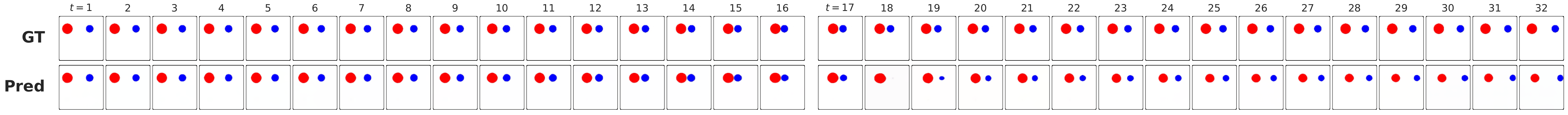}
\includegraphics[width=\linewidth]{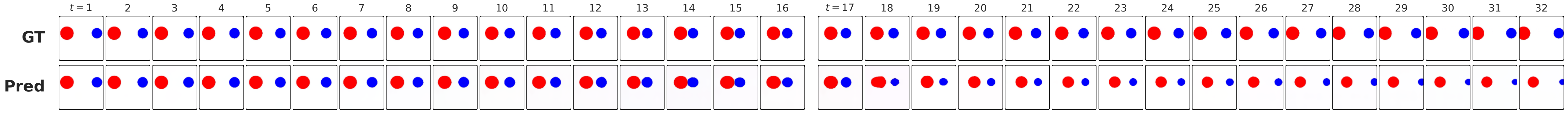}
\includegraphics[width=\linewidth]{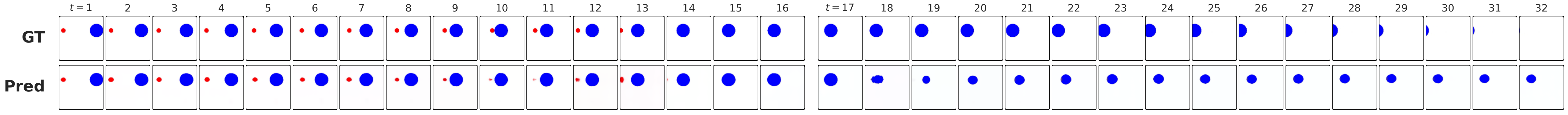}
\includegraphics[width=\linewidth]{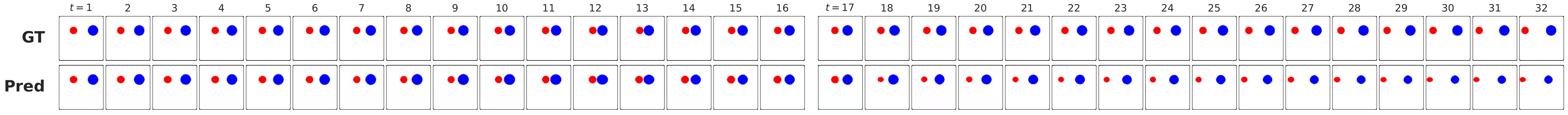}
\caption{\textbf{OOD generalisation on PhyWorld.} Ground truth (GT) versus \themethod predictions for ball radii and velocities outside the training distribution.}
\label{fig:ood_generalisation}
\end{figure}

\paragraph{Unperturbed context-conditioned video generation.} 
We also evaluate on PhyWorld under 32-step-long horizons, this time focusing on out-of-distribution (OOD) generalisation. Specifically, we test on ball radii and velocities that lie outside the training distribution, a regime where models without suitable inductive biases are expected to fail. \cref{fig:ood_generalisation} shows four such sequences: in each case, \themethod tracks the ground truth trajectory faithfully, correctly predicting both position and, implicitly, the physical properties of the balls. This suggests that the weight-space representation generalises beyond the training distribution, capturing the underlying collision dynamics rather than memorising seen configurations.

\paragraph{How far is video generation from world model \cite{kang2024far}?}

\begin{figure}[h]
\centering
\includegraphics[width=\linewidth]{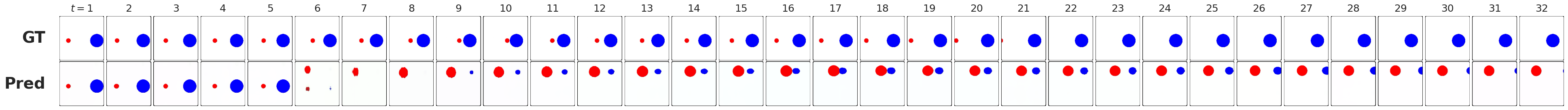}
\includegraphics[width=\linewidth]{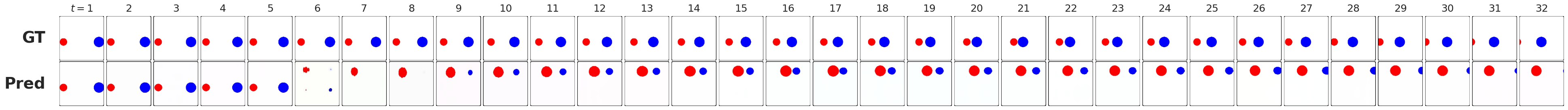}
\caption{PhyWorld \textbf{state} intervention after 5 steps, with forecast generated using 3 context frames. Despite manually editing their identities, we observe accurate collision dynamics that respect the balls' sizes and velocities in the injected states. The fact that the $y$-position of the alien balls is inherited (in addition to shape and colour) is addressed in \cref{sec_app:limitations}.}
\label{fig:phyworld_physics_state_intervention}


\vspace{10pt}


\includegraphics[width=\linewidth]{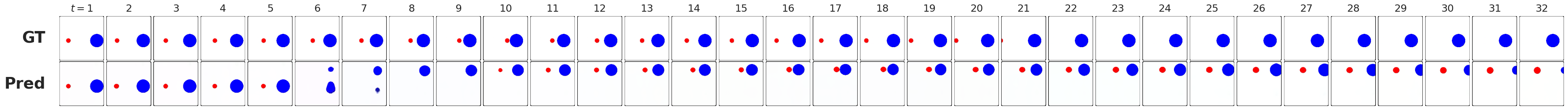}
\includegraphics[width=\linewidth]{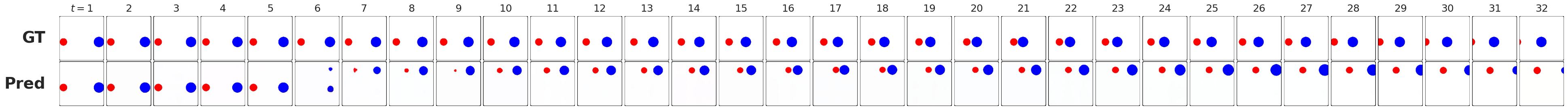}
\caption{PhyWorld \textbf{action} intervention after 5 steps, with forecast generated using 3 context frames. We use the same sequences as in \cref{fig:phyworld_physics_state_intervention}. Injected alien collision dynamics (actions) are the same, leading to unphysical behaviours.}
\label{fig:phyworld_physics_action_intervention}
\end{figure}

\cref{fig:phyworld_physics_state_intervention} shows three ground truth sequences and the resulting latent state interventions. We perform the intervention at step $t=5$, well before the collisions happen. We see that once the new identity is inherited, post-collision velocities are different from the ones we would have expected from the ground truth. These adjusted collision dynamics appear visually accurate, suggesting that \themethod has effectively learned the physics of the PhyWorld environment, placing our latent state intervention experiment as a potent test of world modelling accuracy \cite{kang2024far}.

Importantly, we take from \cref{fig:phyworld_physics_state_intervention} that the actions were recalculated, confirming that latent actions and states are completely dissociated. \cref{fig:phyworld_physics_state_intervention} shows that when actions are inherited, post-collision positions are the same (those of the alien sequence). These results indicate that physics are encoded in the action vector, and altering this vector would result in unphysical behaviour.

\subsubsection{WeatherBench}

\begin{figure}[h]
\centering
\includegraphics[width=\linewidth]{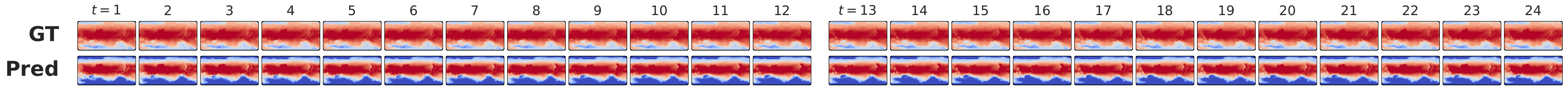}
\includegraphics[width=\linewidth]{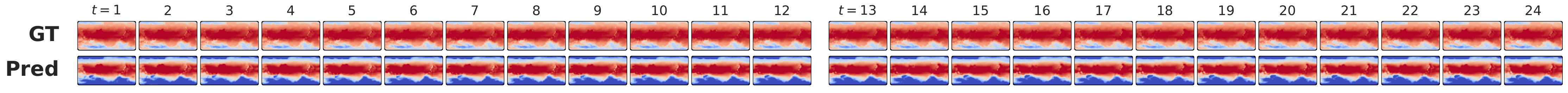}
\includegraphics[width=\linewidth]{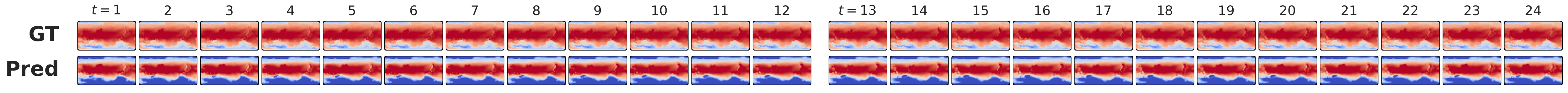}
\includegraphics[width=\linewidth]{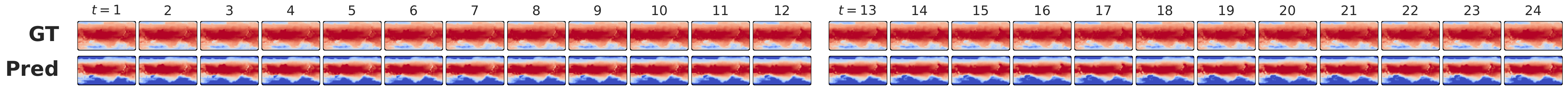}
\includegraphics[width=\linewidth]{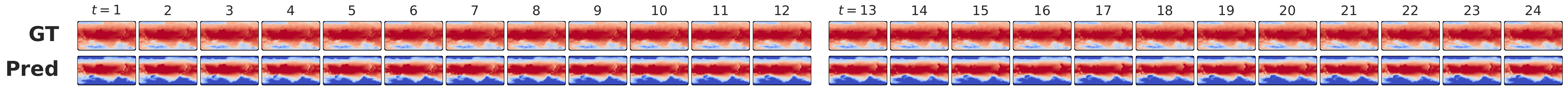}
\caption{\textbf{WeatherBench forecasting.} Ground truth (GT) versus \themethod predictions conditioned on 12 initial frames (frames 1--12) and rolling out for 12 steps (frames 13--24).}
\label{fig:weatherbench_generation}
\end{figure}

We evaluate on WeatherBench with 24-step rollouts: 12 context frames followed by 12 predicted steps. \cref{fig:weatherbench_generation} shows that \themethod accurately reproduces large-scale weather fronts and avoids the excessive smoothing characteristic of phase 2 models trained with reconstruction MSE over the pixel space. We avoid this dilemma by computing the loss in the latent weight space.

\subsection{Addressing Inference-Time Aliasing in INRs}
\label{app:aliasing}

\begin{figure}[h]
\centering
\begin{tikzpicture}[scale=1.05]
    \draw[->, thick] (-1.3, 0) -- (6.6, 0) node[right, font=\small] {Spatial coord.};
    \draw[->, thick] (0, -1.35) -- (0, 1.45) node[above, font=\small] {Amplitude};
    \draw[red!75!black, dashed, thick, domain=0:6, samples=400]
        plot (\x, {cos(324*\x)});
    \draw[blue!75!black, thick, domain=0:6, samples=120]
        plot (\x, {cos(36*\x)});
    \foreach \x in {0,1,2,3,4,5,6} {
        \filldraw[black] (\x, {cos(36*\x)}) circle (2.2pt);
        \draw[gray!60, dotted] (\x, 0) -- (\x, {cos(36*\x)});
    }
    \node[draw=gray!50, fill=white, rounded corners=3pt,
          inner sep=5pt, font=\footnotesize] at (5.85, 1.6) {%
        \begin{tabular}{@{}ll@{}}
            \textcolor{red!75!black}{\rule[0.5ex]{1.0em}{1.2pt}} &
                High frequency $\xi_{\mathrm{high}}$ (out-of-band) \\[2pt]
            \textcolor{blue!75!black}{\rule[0.5ex]{1.0em}{1.2pt}} &
                Low frequency $\xi_{\mathrm{low}}$ (in-band)
        \end{tabular}%
    };
\end{tikzpicture}
\caption{\textbf{Illustration of spatial aliasing.} The out-of-band signal (red, dashed) coincides with the in-band signal (blue) at every training sample point (black dots). During training, however, the loss cannot distinguish the two; weights are optimised against noise, producing artefacts when the INR is queried at a finer out-of-band grid.}
\label{fig:aliasing}
\end{figure}
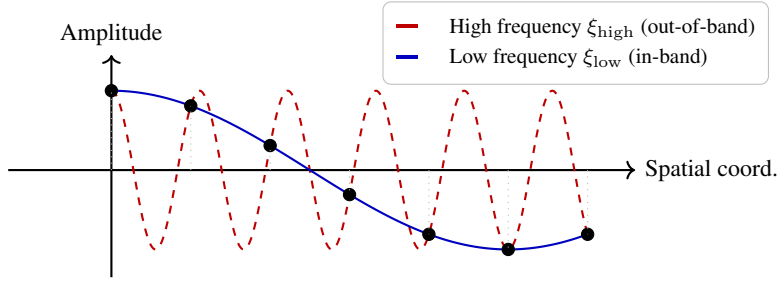

\begin{figure}[h]
\centering
\includegraphics[width=\linewidth]{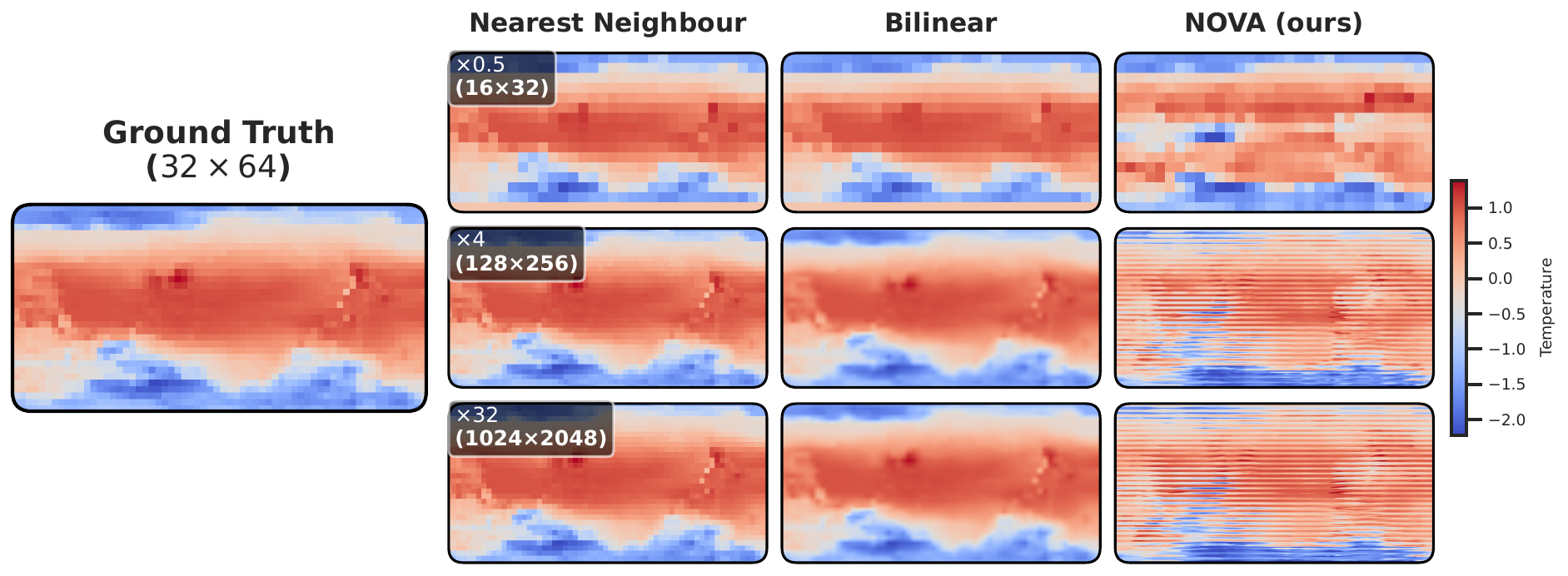}
\caption{\textbf{Zero-shot super-resolution on WeatherBench without frequency masking.} Compared to \cref{fig:superresolution}, horizontal bars are visible, indicating spectral aliasing along the vertical axis.}
\label{fig:superresolution_unmasked}
\end{figure}

Coordinate-based INRs map continuous coordinates to signals via high-frequency sinusoidal embeddings. When queried at a denser grid than the one used during training, as in super-resolution, frequency bands in the Fourier feature encoding may exceed the Nyquist limit of the training grid, producing aliasing artefacts \cite{barron2021mip,lindell2022bacon}. The effect is visible as horizontal bars in \cref{fig:superresolution_unmasked}, while \Cref{fig:aliasing} illustrates why out-of-band frequencies are indistinguishable from in-band ones at the training sample locations.\footnote{For our analysis, we consider aliasing along a single axis $y$ for clarity, as the two spatial dimensions are separable.} The loss cannot differentiate them, so the corresponding weights are optimised against noise and produce artefacts at super-resolution.

\begin{theorem}[Nyquist--Shannon Sampling Theorem (Informal)]
If a continuous function $f(y)$ is sampled at uniform intervals $\Delta y$, it can be unambiguously reconstructed if and only if $f$ contains no spatial frequencies exceeding $\xi_{\mathrm{Nyquist}} = \frac{1}{2\Delta y}$.
\label{theorem:nyquist}
\end{theorem}

For a grid of $H$ pixels over $y\in[-1,1]$, the sampling interval is $\Delta y = 2/H$, giving $\xi_{\mathrm{Nyquist}} = H/4$ periods per unit length. The standard Fourier feature encoding assigns frequency $\xi_k = 2^{k-1}$ to band $k$ as in \cite{tancik2020fourier}
\begin{equation}
    \gamma(y)_k = \bigl(\sin(2^k\pi y),\;\cos(2^k\pi y)\bigr), \qquad k \in \{0,1,\ldots,K-1\},
\end{equation}
or in full,
\begin{equation}
    \gamma(y) = \bigl(\sin(2^0\pi y),\,\cos(2^0\pi y),\;\ldots,\;
                       \sin(2^{K-1}\pi y),\,\cos(2^{K-1}\pi y)\bigr)^\top \quad \in\mathbb{R}^{2K}.
\end{equation}
The no-aliasing condition $\xi_k < \xi_{\mathrm{Nyquist}}$ requires:
\begin{equation}
    2^{k-1} < \frac{H}{4} \implies k_{\max} < \log_2(H) - 1.
    \label{eq:nyquist_limit}
\end{equation}
For MiniGrid ($H=72$), $k_{\max}< 5.17$; therefore, all $K=6$ bands remain within the safe manifold. However, for WeatherBench ($H=32$), $k_{\max}< 4$, meaning that the highest-frequency bands must be eliminated. 


To suppress these, we remark that entries $j=2k$ and $j=2k+1$ in $\gamma(y)$ correspond to band $k$.
We define a binary mask $\mathbf{m}\in\{0,1\}^{2K}$ with
\begin{equation}
    m_{2k} = m_{2k+1} = \mathbf{1}\!\left[k < \log_2(H) - 1\right],
    \qquad k=0,\ldots,K-1,
\end{equation}
and evaluate queries using the masked embedding
$\tilde{\gamma}(y) = \mathbf{m} \odot \gamma(y)$,
where $\odot$ denotes element-wise multiplication.
This zeros out every frequency band that violates \cref{eq:nyquist_limit},
projecting onto the aliasing-free subspace of $\mathbb{R}^{2K}$ induced by the training grid. A similar process is independently applied to the $x$ coordinate.

Bespoke architectures such as BACON \cite{lindell2022bacon} address this by enforcing band-limiting through specialised multiplicative filter networks at training time. Our principled approach, directly inspired by the Nyquist--Shannon Sampling Theorem (see \cref{theorem:nyquist}), requires only a single masking matrix applied to the Fourier embedding at inference time, eliminating aliasing zero-shot, with no architectural changes or retraining.

\subsection{Navigating Discrete Action Spaces within MiniGrid}
\label{subsec:discrete_navigation}


\begin{figure}[h]
\centering
\includegraphics[width=0.45\linewidth]{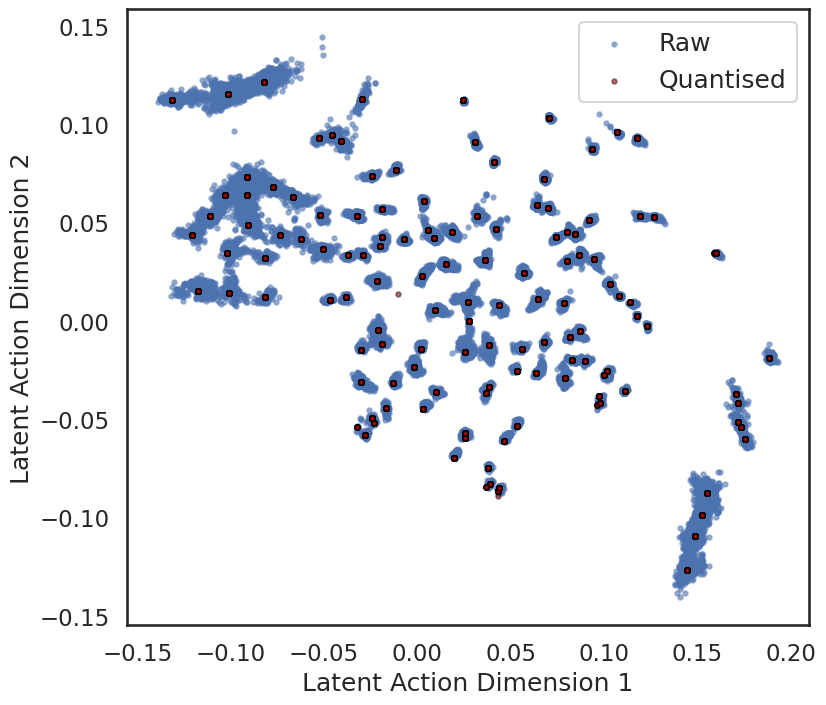}
\includegraphics[width=0.427\linewidth]{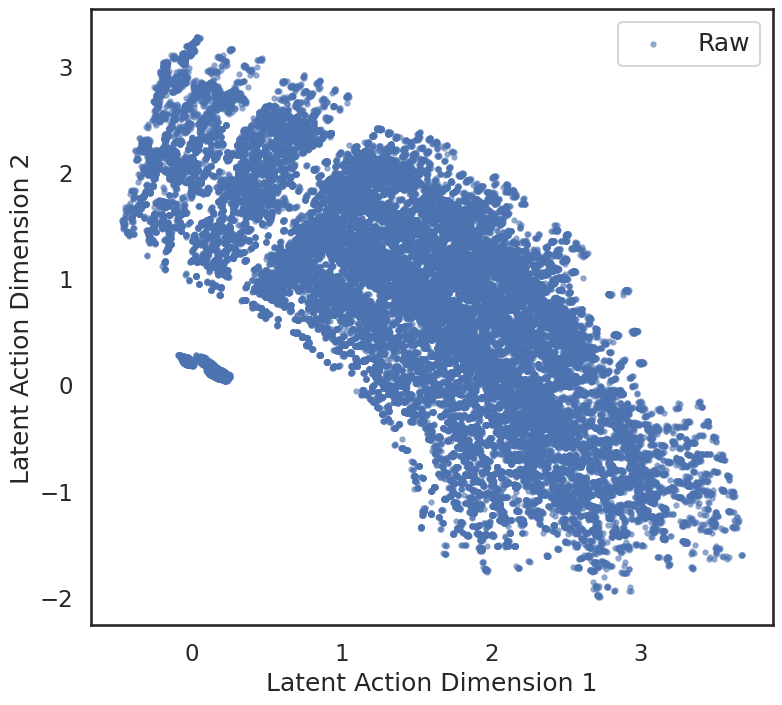}
\caption{\textbf{Latent action dimensions on MiniGrid.} (Left) \textbf{Discrete}: raw actions and their quantised versions. (Right) \textbf{Continuous}: raw actions, whose first two dimensions collectively encode something akin to the orientation of the agent, while the other two (not shown here) encode its location.}
\label{fig:discrete_action_scatter}
\end{figure}

For domains with discrete actions such as \textbf{MiniGrid} \cite{MinigridMiniworld23}, we design a discrete \themethod variant. We employ codebooks \cite{van2017neural}, and we quantise the continuous prediction $\bm{u}_t$ to its nearest codebook vector $\tilde{\bm{u}}_t$. To enable backpropagation through this non-differentiable operation, we apply the straight-through estimator (STE), redefining the action as $\tilde{\bm{u}}_t = \bm{u}_t + \texttt{sg}(\tilde{\bm{u}}_t - \bm{u}_t)$. The FDM is then evaluated on $\tilde{\bm{u}}_t$, and the objective \eqref{eq:phase2loss} is augmented with the standard codebook and commitment losses to align the embeddings.

\begin{figure}[htbp]
\centering
\includegraphics[width=\linewidth]{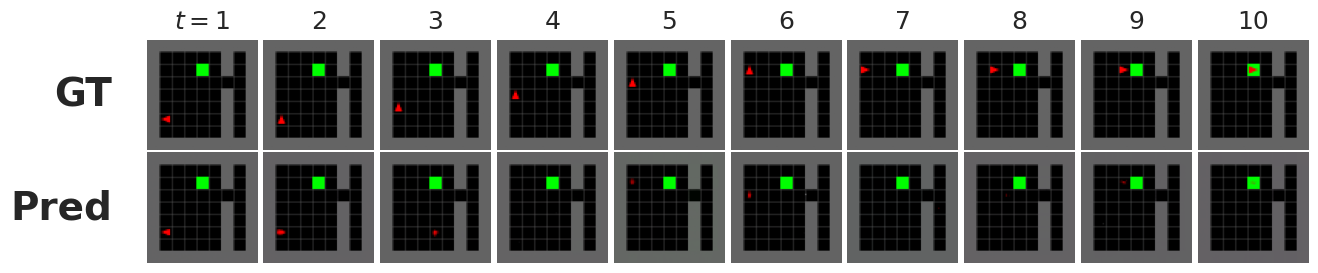}
\includegraphics[width=\linewidth]{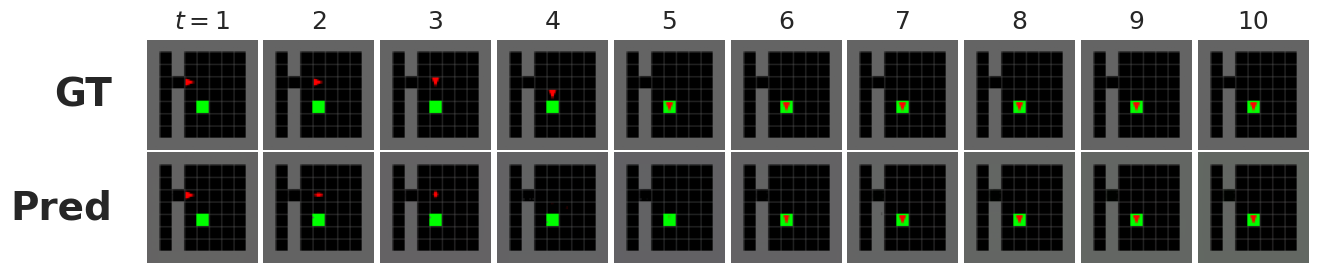}
\includegraphics[width=\linewidth]{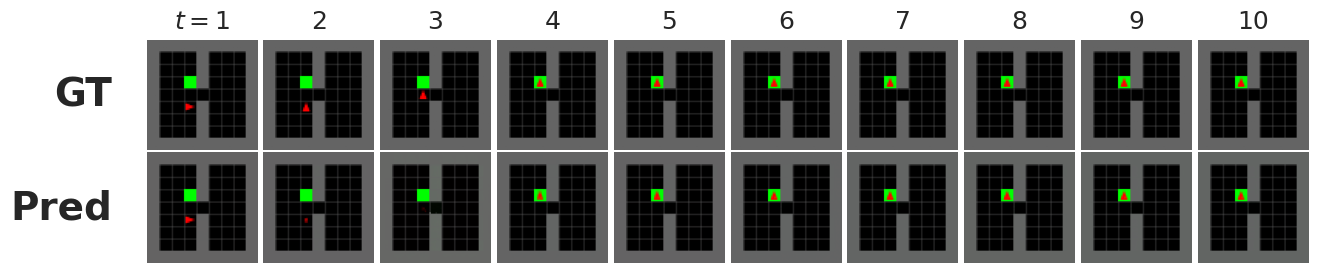}
\caption{Goal-directed navigation within a \textbf{\emph{discrete}} action space. The GCM, conditioned only on the initial frame, imitates the BFS policy to navigate towards the goal (green box). Ground truth (top row) vs. \themethod prediction (bottom row) per sequence.}
\label{fig:discrete_action}

\vspace{2em} 

\includegraphics[width=\linewidth]{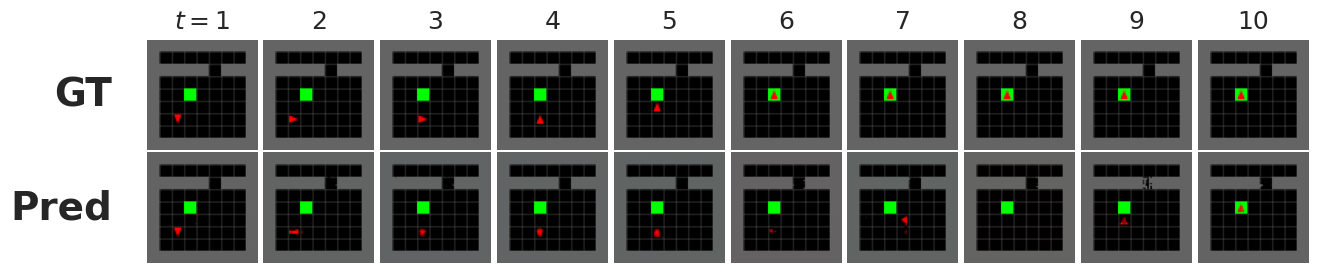}
\includegraphics[width=\linewidth]{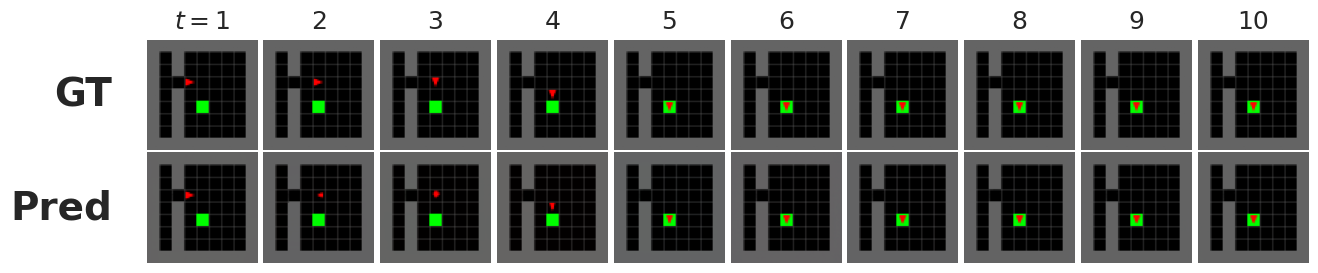}
\includegraphics[width=\linewidth]{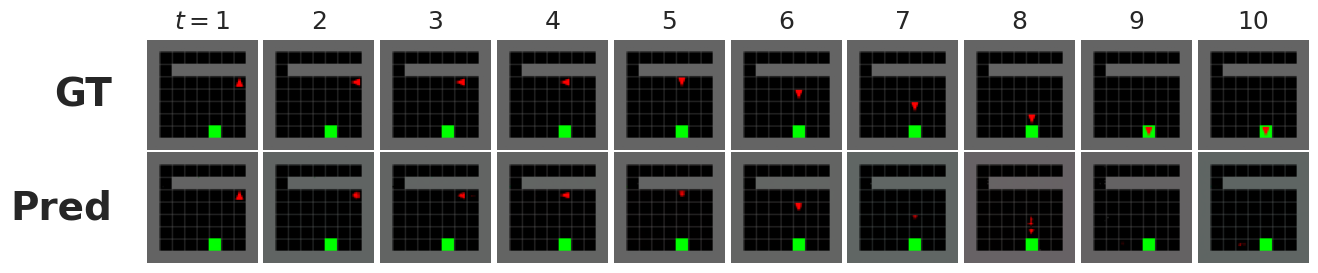}
\caption{Goal-directed navigation within a \textbf{\emph{continuous}} action space. Similar to \cref{fig:discrete_action}, the agent (red triangle) is driven by GCM-generated actions.}
\label{fig:continuous_action}
\end{figure}

\Cref{fig:discrete_action} shows \themethod's behaviour cloning GCM operating on MiniGrid. Although the environment has a small set of canonical discrete actions (turn left, turn right, move forward, toggle), we  employed 200 actions, hoping to form action clusters as done by \citet{schmidt2023learning}. \Cref{fig:discrete_action_scatter} (left) shows soft clusters around the quantised actions formed by the raw GCM-predicted actions. 




We also implement the continuous variant as defined in \cref{fig:nova_method} for direct comparison. Continuous actions train faster and drive the loss roughly two orders of magnitude lower, yielding better visual quality in the downstream GCM-based behaviour cloning rollouts (seen by comparing \cref{fig:continuous_action} to \cref{fig:discrete_action}). Furthermore, the continuous action geometry appears more interpretable, as \cref{fig:discrete_action_scatter} implies that latent action dimensions 1 and 2 encode the orientation of the agent (red triangle). Taken together, these results suggest that the \themethod framework is more naturally suited to continuous action representations. This is consistent with the finding of ~\citet{garrido2026learning} that discrete codebooks are insufficient for unlabelled videos in the wild.



\subsection{Comparison to Weight-Space Linear Recurrent Neural Networks}
\label{sec:comparison_wsl}

Within the emerging literature on Weight-Space Learning (WSL), our proposed framework (specifically, \themethod's GCM in phase 3) shares several similarities with contemporary architectures, most notably Weight-space Adaptive Recurrent Prediction (WARP) \cite{nzoyem2026weightspace}. Both frameworks view the latent state of the dynamic system not as a standard feature vector, but as the weights and biases of an INR. In WARP, this is governed by a continuous linear recurrence defined as $\bm{z}_t = A\bm{z}_{t-1} + B\Delta \bm o_t$, where the input difference $\Delta \bm o_t$ between consecutive frames acts as an implicit action driving the process. Despite these foundational similarities, WARP is a general sequence model, whereas our framework is designed as a comprehensive world model for modelling complex video dynamics.

\paragraph{Implementation} To conduct an empirical comparison, we implemented the WARP architecture utilising the JAX \cite{jax2018github} and Equinox \cite{kidger2021equinox} libraries. We meticulously controlled the experimental environment to ensure parity with our framework's training conditions, matching the batch size and total computational budget. The WARP baseline was trained using a fixed learning rate of $10^{-6}$. Given stability issues \cite[p. 23]{nzoyem2026weightspace}, we stabilise the linear recurrence and prevent the unbounded explosion by enforcing weight clipping $\bm{z}_t \in [-0.5, 0.5]$. Furthermore, the decoded outputs were subjected to pixel clipping between $0$ and $1$. We matched the capacity of \themethod by reusing an identical INR of dimension $d_{\bm{z}}=961$. Finally, the same \themethod CNN encoder architecture was utilised to act as WARP's initial hypernetwork $\phi$, mapping the initial visual observation into the high-dimensional weight space: $x_0 \mapsto \bm{z}_0$.

\begin{figure}[h]
    \includegraphics[width=\textwidth]{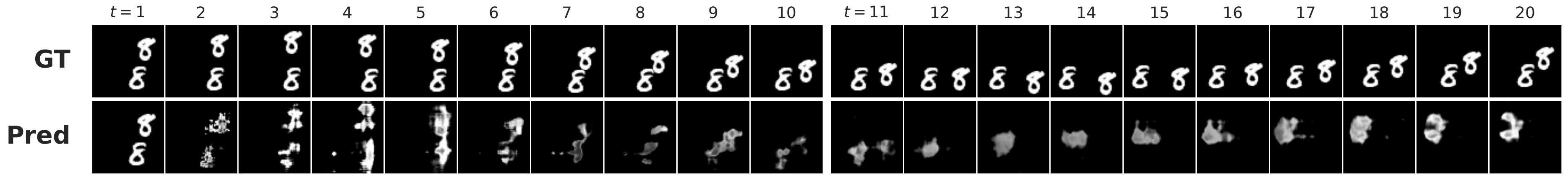}
    \includegraphics[width=\textwidth]{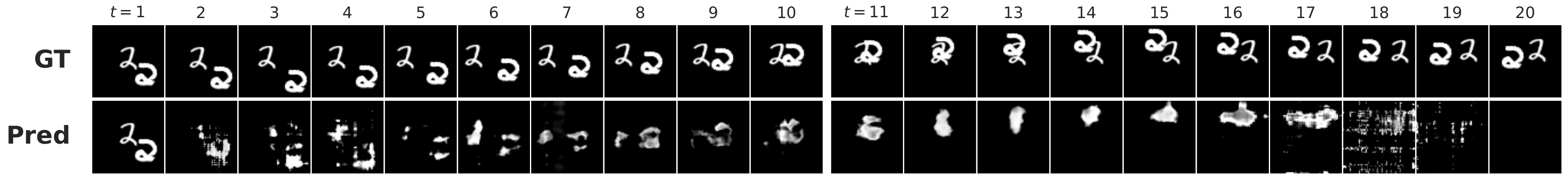}
    \includegraphics[width=\textwidth]{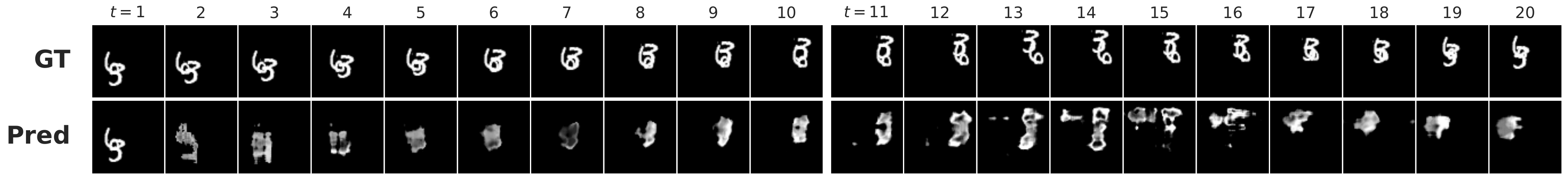}
    \includegraphics[width=\textwidth]{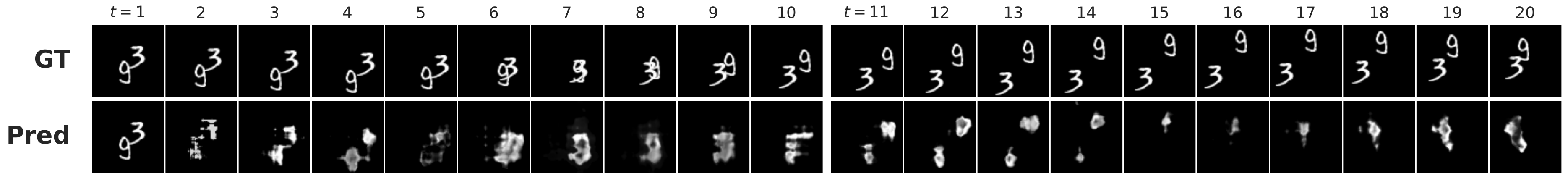}
    \includegraphics[width=\textwidth]{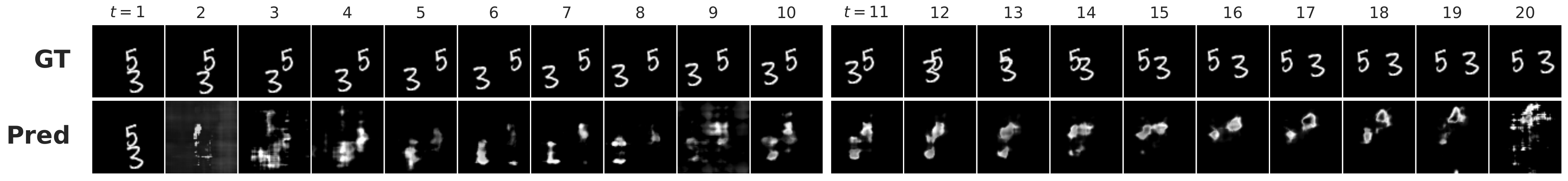}
    \includegraphics[width=\textwidth]{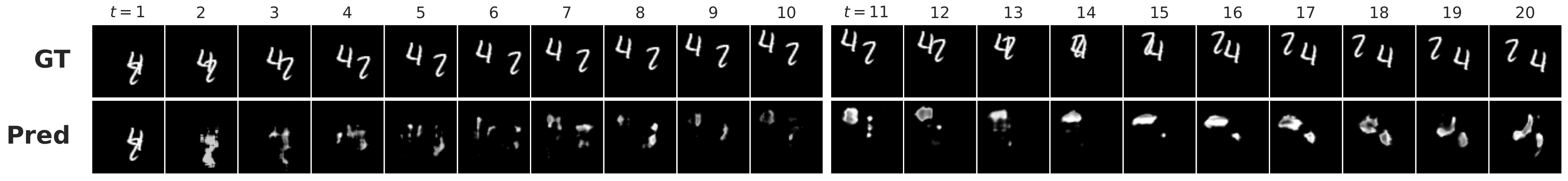}
    \vspace{-0.2cm}
    \caption{WARP forecasting results conditioned on $t=10$ context frames. While the model tracks general temporal dynamics, it fails to maintain sharp visual fidelity, yielding blurred and ghosted predictions.}
    \label{fig:warp_forecasting}
\end{figure}

\paragraph{Forecasting results.} Standard evaluation of the forecasted rollouts (Figure \ref{fig:warp_forecasting}) demonstrates that WARP is somewhat capable of capturing spatial-temporal dynamics when conditioned on 10 previous context frames. However, qualitative assessments of the generated sequences reveal a substantial degradation in visual fidelity. The predicted digits suffer from severe artifacting; they are noticeably blurry, ghosted, and lack crisp, well-defined boundaries. This visual degradation suggests that WARP's unconstrained linear recurrence struggles to balance two competing objectives: maintaining accurate physical trajectories and achieving photorealism.

\paragraph{Latent disentanglement analysis.} To probe the underlying topology of WARP's learned representations, we performed the latent state intervention experiment from Figure \ref{fig:fork_comparison}. Following the processing of the initial 3 frames and autoregressive generation after those, we intervened at $t=6$ by replacing the naturally evolved state $\bm{z}_t$ with the latent encoding of an alien frame. 

\begin{figure}[h]
    \centering
    \includegraphics[width=0.55\linewidth]{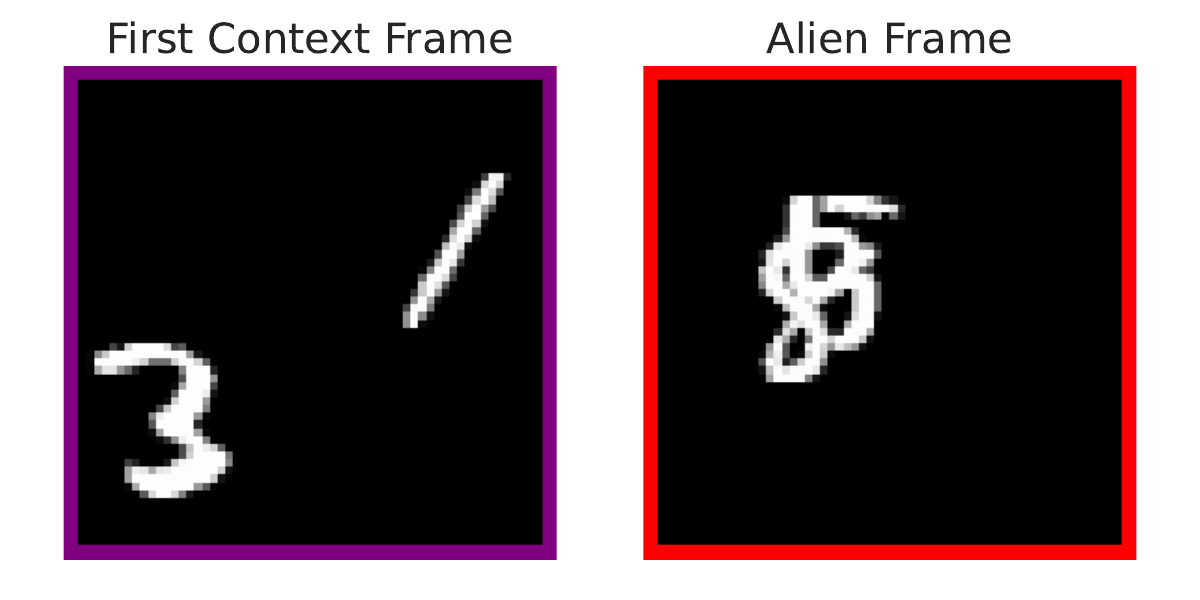}
    \vspace{0.1cm}
    \centerline{(a) Initial context frame and ground truth alien frame for latent intervention.}
    
    
    \includegraphics[width=0.95\linewidth]{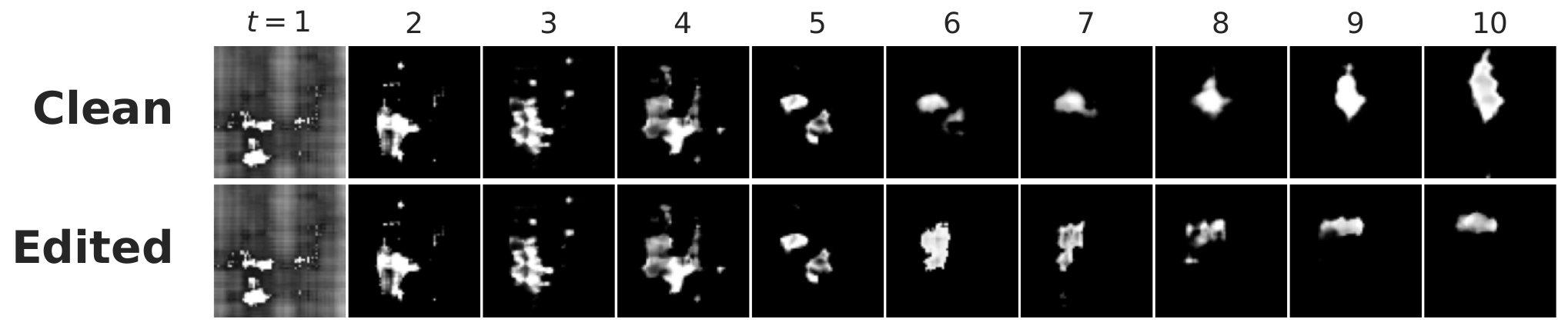}
    \vspace{0.1cm}
    \centerline{(b) Intervention sequence rollout}
    \caption{Latent State Intervention in WARP. (a) The original context frame alongside the isolated decoding of the alien latent state. Note the poor baseline reconstruction quality of the encoder. (b) When the alien state is substituted at $t=6$, the linear recurrence is fundamentally disrupted, proving that content and motion are highly entangled within $\bm{z}_t$.}
    \label{fig:warp_intervention}
\end{figure}

However, the intervention sequence (Figure \ref{fig:warp_intervention}b) demonstrates a catastrophic failure, as WARP allows the alien injection to fundamentally corrupt the forward dynamics (similar to LAPO). Furthermore, Figure \ref{fig:warp_intervention}a indicates that the CNN encoder appears unable to reconstruct the latents with high visual quality even in isolation. This is likely because the initial representation $\bm{z}_0$ it is forced to generate must be mathematically compatible with the subsequent linear recurrence. We believe this imposes an architectural compromise, yielding an initial state that neither perfectly suits the unrolled recurrence nor permits faithful visual reconstruction. 
This behaviour proves that the latent representations learned by WARP are thoroughly entangled, with spatial content, temporal coordinates, and motion confounded within $\bm{z}_t$.

\subsection{Additional Ablation Studies}
\label{sec:ablations}


Throughout our study, we performed several ablation studies, such as additive vs monolithic FDM \cref{subsec:motionretargeting}, and discrete vs continuous action spaces in \cref{subsec:discrete_navigation}. We use this section to present additional abstractions that justify our architectural choices.

\begin{figure}
\centering
  \includegraphics[width=0.65\linewidth]{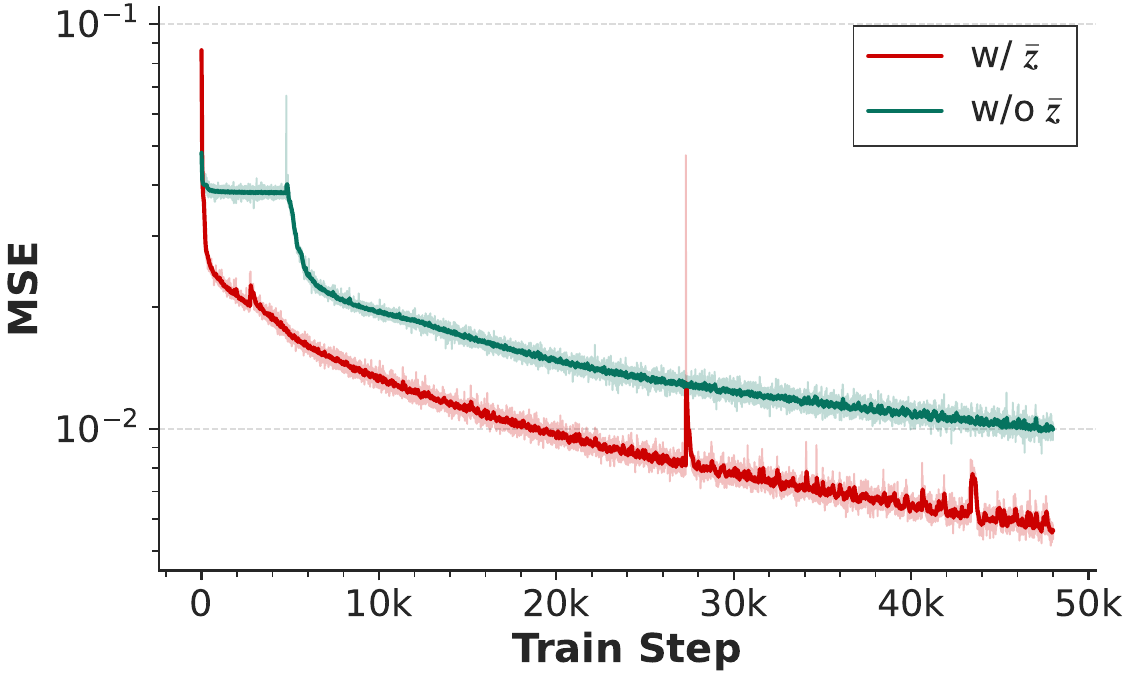}
  \caption{Phase 2 training loss with and without $\bar{\bm{z}}$ on Moving MNIST.}
  \label{fig:bar_z_ablation}
\end{figure}


\paragraph{Removing the $\bar{\bm{z}}$ anchor.}

Ablating the base parameter $\bar{\bm{z}}$ and predicting full INR weights $\bm{z}_t$ instead of relative offsets fundamentally destabilises the learning process, trapping the network in local minima for prolonged periods (see \cref{fig:bar_z_ablation}). This observation suggests that $\bar{\bm{z}}$ prevents severe bottlenecks and gradient instabilities that arise when the model is forced to independently regress the full high-dimensional weight space from scratch at every time step.

\paragraph{Replacing the recurrent GCM architectures.} In our study, we observed on PhyWorld 30K that latent interventions are most effective when paired with non-attention-based GCM architectures such as recurrent neural networks. We hypothesise that this is because standard Transformers compute global attention over the entire state-action history. As a result, injecting an alien action mid-sequence causes attention disruption and visual artifacting. 




\section{Limitations \& Failure Modes}

\label{sec_app:limitations}

\begin{figure}[h]
\centering
\includegraphics[width=0.9\linewidth]{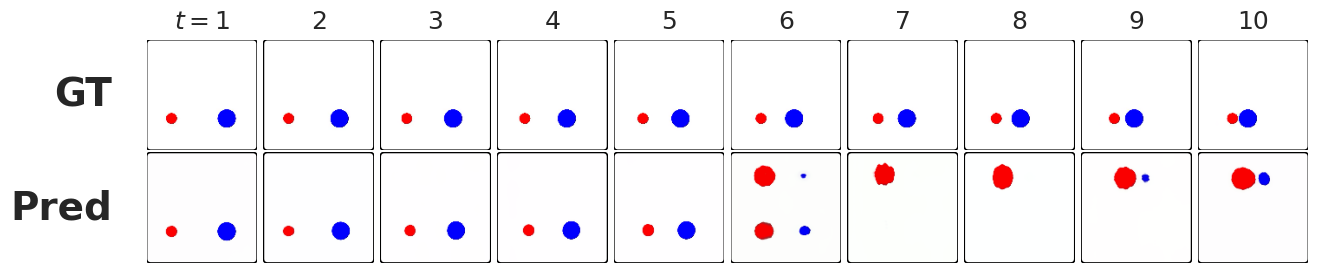}
\includegraphics[width=0.9\linewidth]{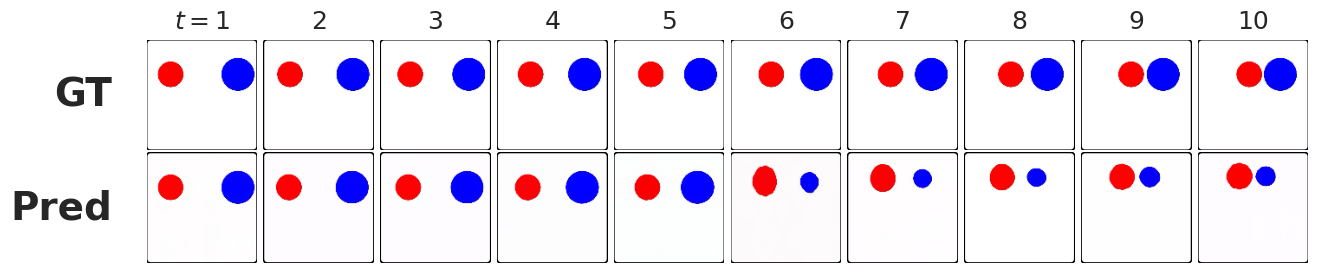}
\includegraphics[width=0.9\linewidth]{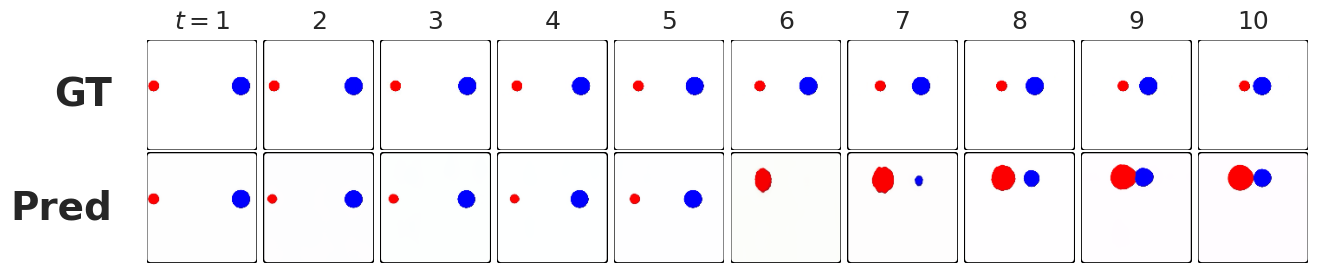}
\caption{Zero-shot content retargeting on PhyWorld. Intervention starts at $t=5$, with all subsequent states replaced with those corresponding to the same alien sequence. While all sequences inherit new ball colours and shapes, we find that in addition, the $y$-position is inherited, thus altering visual dynamics, which should be free. Future work will aim to make the model robust to such spurious correlations while preserving dynamics.}
\label{fig:phyworld_content_interv}
\end{figure}

\paragraph{Spurious information encoded within latent states and actions.}
On Moving MNIST, the latent action $\bm{u}_t$ encodes not only the next digit location but also brightness variations.
This is partly because the IDM is trained to explain \emph{all} inter-frame differences, including those due to noise \cite{zhang2025latent}. Furthermore, because the actions encode location as "go to $x,y$" instead of "increment by $\Delta x, \Delta y$", it becomes harder to map predicted actions to ground truths when available \cite{schmidt2023learning}. On PhyWorld, we find that intervening on the content also shifts the movement along the vertical (see \cref{fig:phyworld_content_interv}); we hypothesise that this is because during training, the model always sees each ball at the same height, and thus encodes the $y$-location as part of the identity.  Future work could explore regularisation mechanisms beyond discretisation ~\cite{garrido2026learning} to constrain the latent spaces. For example, applying SICReg \cite{maes2026leworldmodel} to bottleneck the action space and force it to encode only meaningful user-mappable motion information (for Moving MNIST and MiniGrid), and data augmentations such as video rotations for physics-abiding datasets such as PhyWorld.

\begin{figure}[H]
\centering
\includegraphics[width=0.45\linewidth]{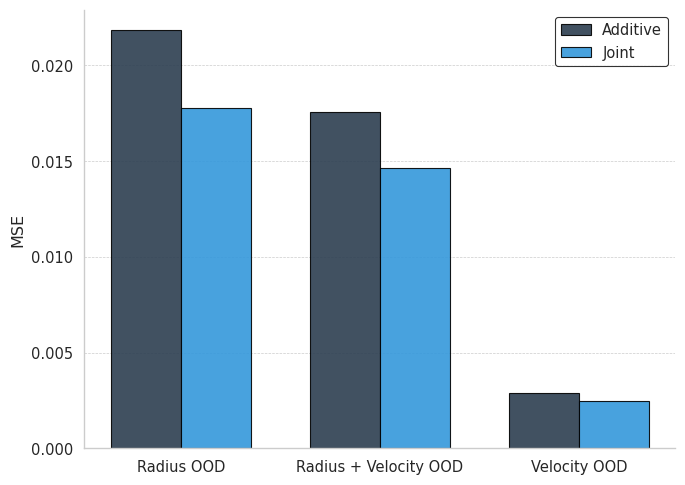}
\caption{Additive vs. monolithic (joint) forward dynamics on PhyWorld. Although detrimental for disentangled representations, a joint FDM performs better out-of-distribution.}
\label{fig:phyworld_ood_breakdown}
\end{figure}

\paragraph{Physical constraints and OOD generalisation.}
On PhyWorld, we observed diminished OOD performance after suppressing the $A/B$ seperation and using a single monolithic FDM (see \cref{fig:phyworld_ood_breakdown,tab:phyworld_phase2_ood}). We believe scientific machine learning techniques \cite{rackauckas2020universal} could help alleviate this issue, whereby the INR would be forced to satisfy laws of motion such as $\bm{z}_{t+1} = \bm{z}_t + \Delta t \cdot \bm{u}_t$ \cite{nzoyem2026weightspace}, or any other partial differential equation thanks to the implicit function theorem \cite{bai2019deep}.


\begin{table*}[h]
\centering
\caption{Comparison of additive and joint model performance on in-distribution (InD) and out-of-distribution (OOD) data after phase 2 training (i.e., using the IDM to generate actions). Values are reported as mean $\pm$ standard deviation.}
\label{tab:phyworld_phase2_ood}
\resizebox{0.95\textwidth}{!}{ 
\begin{tabular}{llcccccc}
\toprule
\multirow{2}{*}{\textbf{Distribution}} & \multirow{2}{*}{\textbf{Model}} & \multicolumn{3}{c}{\textbf{Image Metrics}} & \multicolumn{3}{c}{\textbf{Physical Metrics}} \\
\cmidrule(lr){3-5} \cmidrule(lr){6-8}
& & \textbf{MSE $\downarrow$} & \textbf{PSNR $\uparrow$} & \textbf{SSIM $\uparrow$} & \textbf{Pos. Error $\downarrow$} & \textbf{Mom. Error $\downarrow$} & \textbf{K.E. Error $\downarrow$} \\
\midrule
\multirow{2}{*}{InD}
& Additive & $0.0008 \pm 0.0004$ & $32.52 \pm 1.52$ & $0.9792 \pm 0.0072$ & $0.0065 \pm 0.0188$ & $0.0308 \pm 0.0455$ & $0.0404 \pm 0.2024$ \\
& Joint    & $0.0007 \pm 0.0003$ & $32.64 \pm 1.40$ & $0.9803 \pm 0.0071$ & $0.0031 \pm 0.0051$ & $0.0191 \pm 0.0099$ & $0.0041 \pm 0.0049$ \\
\midrule
\multirow{2}{*}{OOD}
& Additive & $0.0141 \pm 0.0194$ & $24.70 \pm 6.23$ & $0.9255 \pm 0.0701$ & $0.0739 \pm 0.0883$ & $0.4109 \pm 0.8788$ & $0.6531 \pm 1.8469$ \\
& Joint    & $0.0117 \pm 0.0160$ & $25.08 \pm 5.87$ & $0.9297 \pm 0.0649$ & $0.0753 \pm 0.0882$ & $0.3380 \pm 0.6255$ & $0.6463 \pm 1.7010$ \\
\bottomrule
\end{tabular}
}
\end{table*}

\paragraph{SOTA video generation.} While our framework enables controllable generation, we do not achieve state-of-the-art context-conditioned video forecasting. Supervised spatio-temporal models from the OpenSTL benchmark \cite{tan2023openstl} achieve highly optimised pixel-level metrics. This performance gap arises from an architectural constraint that forces \themethod to autoregressively rely on an inferred latent action signal to drive the generation process. Similar trade-offs where world models sacrifice raw pixel-level reconstruction performance to prioritise powerful latent representations have recently been proposed \cite{terver2026lightweight}. That said, bridging this gap remains an exciting frontier for future work, paving the way for architectures that seamlessly unite state-of-the-art visual fidelity with interpretable, action-driven control.



\end{document}